\documentclass[twoside,11pt]{article}

\usepackage{jmlr2e}
\usepackage{lastpage}
\jmlrheading{22}{2021}{1-\pageref{LastPage}}{9/20; Revised4/21}{7/21}{20-1061}{Yingfan Wang, Haiyang Huang, Cynthia Rudin, and Yaron Shaposhnik}
\ShortHeadings{Understanding How Dimension Reduction Tools Work}{Wang, Huang, Rudin, and Shaposhnik}

\usepackage{microtype}
\usepackage{graphicx}
\usepackage{amsfonts}
\usepackage{booktabs} 
\usepackage{epsfig}
\usepackage{amssymb}
\usepackage{natbib}
\usepackage{float}
\usepackage{subfig}
\usepackage{amsmath}
\usepackage{makecell}
\usepackage{xcolor}
\usepackage{algorithm} 
\usepackage{algorithmic} 

\usepackage{soul} 

\newcommand{\eqdef}{\stackrel{\rm{def}}{=}}

\newcommand{\iterations}{\textrm{iterations}}

\newcommand{\neighbors}{NB}
\newcommand{\nneighbors}{n_{\neighbors}}
\newcommand{\ratio}{{\textrm{ratio}}}

\usepackage{bbold} 

\newcommand{\eps}{\epsilon}

\newcommand\norm[1]{\left\lVert#1\right\rVert}

\usepackage[symbol]{footmisc}

\begin{document}

\title{Understanding How Dimension Reduction Tools Work: An Empirical Approach to Deciphering t-SNE, UMAP, TriMap, and PaCMAP for Data Visualization}

\author{\name Yingfan Wang$^*$ 
 \email yingfan.wang@duke.edu \\
       \addr Duke University, USA
       \AND
       \name Haiyang Huang$^*$
       \email haiyang.huang@duke.edu\\
        \addr Duke University, USA
        \AND
       \name Cynthia Rudin \email cynthia@cs.duke.edu\\
        \addr Duke University, USA
       \AND
       \name Yaron Shaposhnik \email{yaron@simon.rochester.edu}\\
       \addr University of Rochester, USA
     }

\editor{Tina Eliassi-Rad}

\maketitle

\begin{abstract}
Dimension reduction (DR) techniques such as t-SNE, UMAP, and TriMap have demonstrated impressive visualization performance on many real-world datasets. One tension that has always faced these methods is the trade-off between preservation of global structure and preservation of local structure: these methods can either handle one or the other, but not both. In this work, our main goal is to understand what aspects of DR methods are important for preserving both local and global structure: it is difficult to design a better method without a true understanding of the choices we make in our algorithms and their empirical impact on the low-dimensional embeddings they produce. Towards the goal of local structure preservation, we provide several useful design principles for DR loss functions based on our new understanding of the mechanisms behind successful DR methods. Towards the goal of global structure preservation, our analysis illuminates that \textit{the choice of which components to preserve} is important. We leverage these insights to design a new algorithm for DR, called Pairwise Controlled Manifold Approximation Projection (PaCMAP), which preserves both local and global structure. Our work provides several unexpected insights into what design choices both to make and avoid when constructing DR algorithms.\footnote[1]{denotes equal contribution} 
\end{abstract}


\section{Introduction}\label{submission}

Dimension reduction (DR) tools for data visualization can act as either a blessing or a curse in understanding the geometric and neighborhood structures of datasets. Being able to visualize the data can provide an understanding of cluster structure or provide an intuition of distributional characteristics. On the other hand, it is well-known that DR results can be misleading, displaying cluster structures that are simply not present in the original data, or showing observations to be far from each other in the projected space when they are actually close in the original space \citep[e.g., see][]{wattenberg2016use}. 
Thus, if we were to run several DR algorithms and receive different results, it is not clear how we would determine which of these results, if any, yield trustworthy representations of the original data distribution.

The goal of this work is to decipher these algorithms, and why they work or don't work. We study and compare several leading algorithms, in particular, 
t-SNE \citep{van_der_Maaten08}, UMAP \citep{UMAP}, and TriMap \citep{TriMap}.
Each of these algorithms is subject to different limitations. 
For instance, t-SNE can be very sensitive to the perplexity parameter and creates spurious clusters; both t-SNE and UMAP perform beautifully in preserving local structure but struggle to preserve global structure \citep{wattenberg2016use, UmapMammoth}. TriMap, which (to date) is the only triplet model to approach the performance levels of UMAP and t-SNE, handles global structure well; however, as we shall see, TriMap's success with global structure preservation is not due to reasons we expect, based on its derivation. TriMap also struggles with local structure sometimes. Interestingly, none of t-SNE, UMAP, or TriMap can be adjusted smoothly from local to global structure preservation through any obvious adjustment of parameters. Even basic comparisons of these algorithms can be tricky: each one has a different loss function with many parameters, and it is not clear which parameters matter, and how parameters correspond to each other across algorithms. For instance, even comparing the heavily related algorithms t-SNE and UMAP is non-trivial; their repulsive forces between points arise from two different mechanisms. In fact, several papers have been published that navigate the challenges of tuning parameters and applying these methods in practice \citep[see][]{wattenberg2016use,cao2017automatic,nguyen2019ten,Belkina19OptSNE}.

Without an understanding of the algorithms' loss functions and what aspects of them have an impact on the embedding, it is difficult to substantially improve upon them. Similarly, without an understanding of other choices made within these algorithms, it becomes hard to navigate adjustments of their parameters. Hence, we ask: What aspects of the various loss functions for different algorithms are important? Is there a not-very-complicated loss function that allows us to handle both local and global structure in a unified way? Can we tune an algorithm to migrate smoothly between local and global structure preservation in a way we can understand? Can we preserve both local and global structure within the same algorithm? Can we determine what components of the high-dimensional data to preserve when reducing to a low-dimensional space? 
We have found that addressing these questions requires looking at two primary design choices: the \textit{loss function}, and the choice of \textit{graph components} involved in the loss function. 

The loss function controls the attractive and repulsive forces between each pair of data points. 
In Section \ref{sec:principles_good_objective}, we focus on deciphering general principles of a good loss function. We show how UMAP and other algorithms obey these principles, and show empirically why deviating from these principles ruins the DR performance. We also introduce a simple loss function obeying our principles. This new loss function is used in an algorithm called Pairwise Controlled
Manifold Approximation Projection (PaCMAP) introduced in this work. Only by visualizing the loss function in a specific way (what we call a ``rainbow figure'') were we able to see why this new loss function works to preserve local structure.

The choice of graph components determines which subset of pairs are to be attracted and repulsed. This is the focus of Section \ref{sec:graph_component}.
Generally, we try to pull neighbors from the high-dimensional space closer together in the low-dimensional space, while pushing further points in the original space away in the low-dimensional space. We find that the choices of which points to attract and which points to repulse are important in the preservation of local versus global structure. We provide some understanding of how to navigate these choices in practice. Specifically, we illustrate the importance of \textit{having forces on non-neighbors}. One mechanism to do this is to introduce ``mid-near'' pairs to attract, which provide a contrast to the repulsion of further points. Mid-near pairs are leveraged in PaCMAP to preserve global structure.

Throughout, we also discuss other aspects of DR algorithms. For example, in Section \ref{sec:initialization}, we show that initialization and scaling can be important, in fact, we show that TriMap's ability to preserve global structure comes from an unexpected source, namely its initialization.

\begin{figure}[ht]
\begin{center}
\includegraphics[width=0.95\columnwidth]{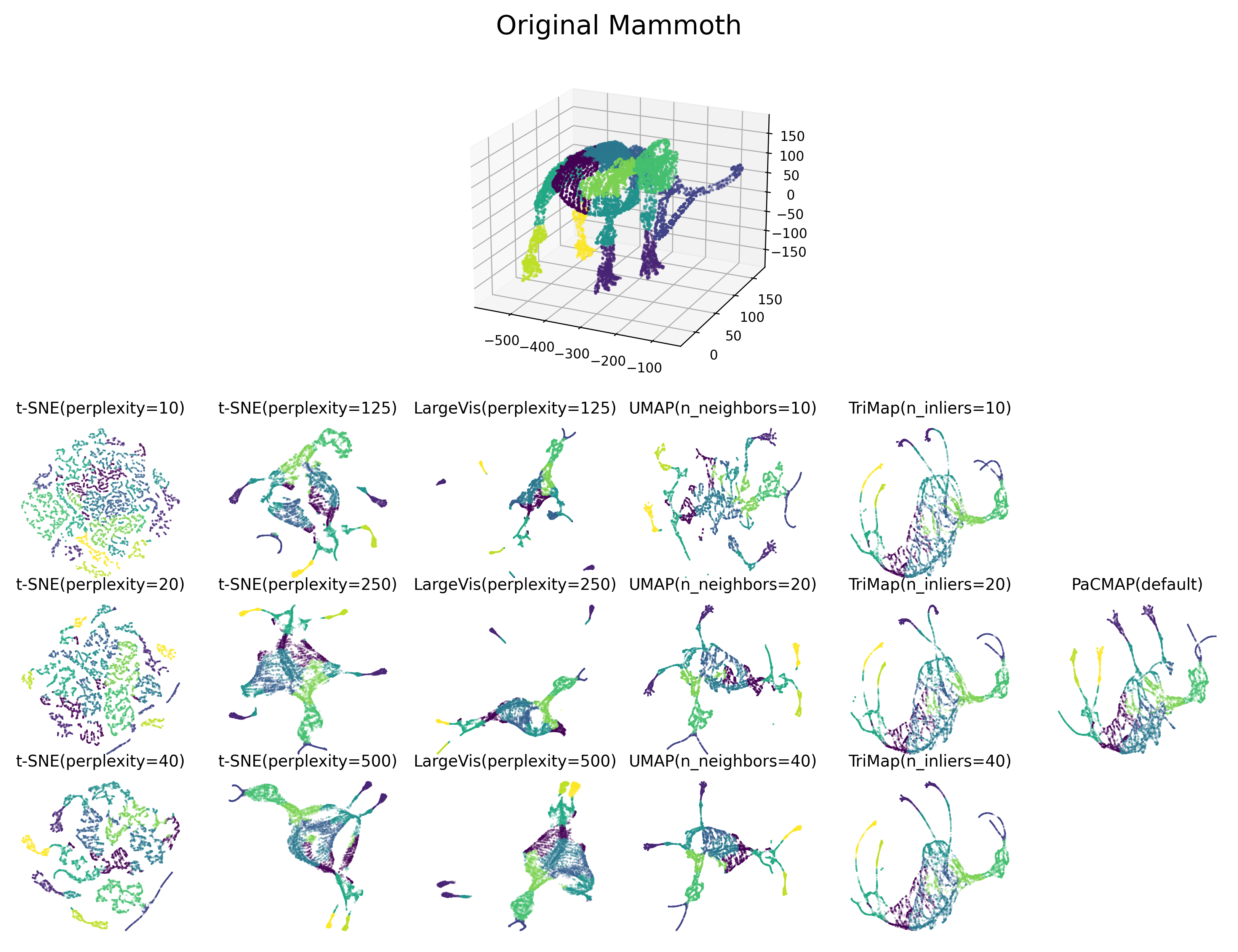}
   \caption{Application of t-SNE, LargeVis, UMAP, TriMap and PaCMAP to the Mammoth dataset. In judging these figures, one should consider preservation of both local structure (e.g., mammoth toes) and global structure (overall shape of the mammoth).
   }
\label{fig:Mammoth}
\end{center}
\end{figure}

\begin{figure}[ht]
\begin{center}
\centerline{\includegraphics[width=.95\columnwidth]{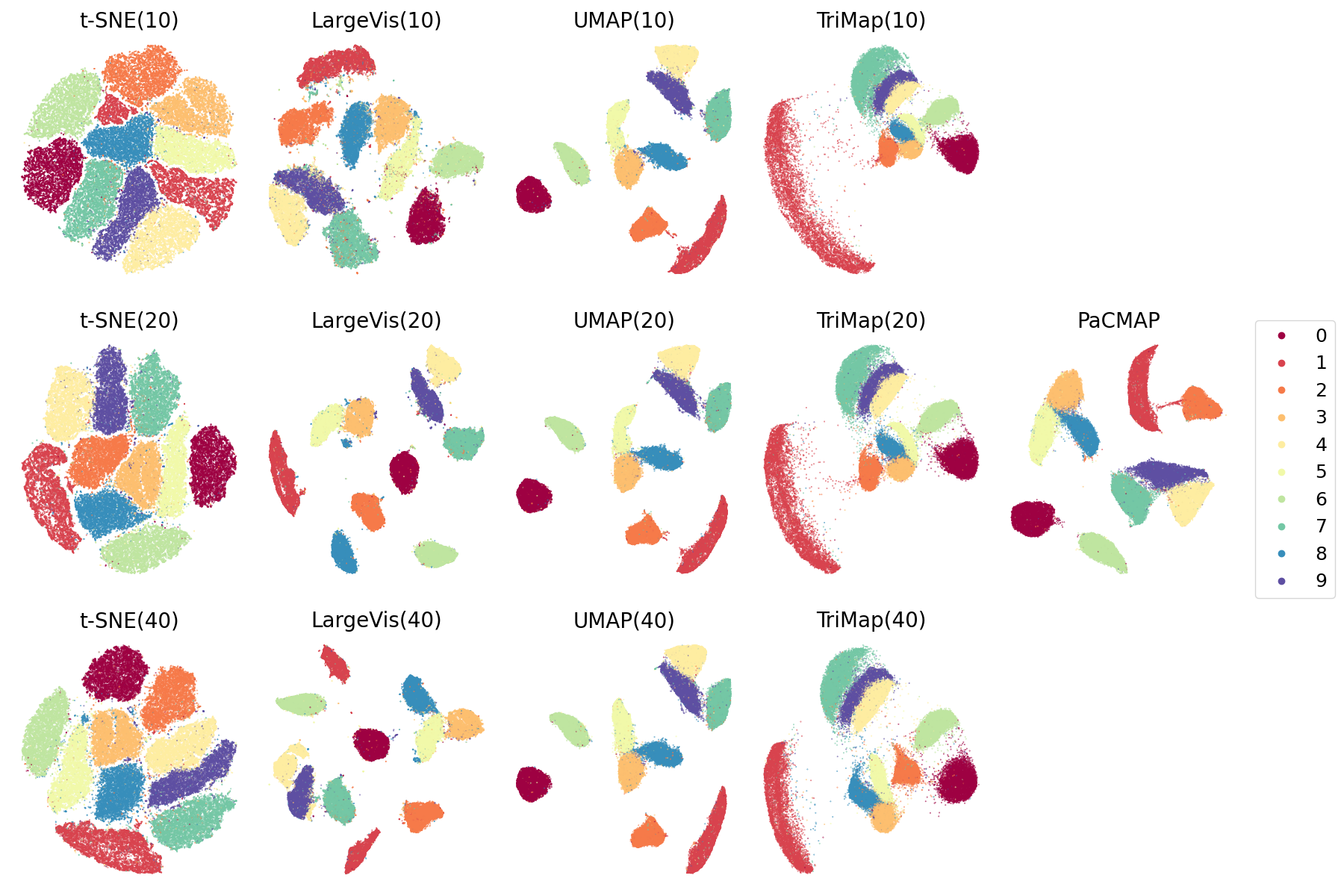}}
\caption{Application of t-SNE, UMAP, TriMap and PaCMAP to the MNIST dataset. Algorithms that preserve local structure tend to separate MNIST classes better. Similar to the experiment in Figure~\ref{fig:Mammoth}, we tuned the most important parameter controlling the visual effect for each algorithm, which is perplexity for t-SNE and LargeVis, $\nneighbors$ for UMAP, and $n_{inlier}$ for TriMap. The values we considered are in the parentheses.}
\label{fig:mnist}
\end{center}
\end{figure}

Figure~\ref{fig:Mammoth} illustrates the output of several algorithms applied to reduce the dimension of the Mammoth dataset \citep{UmapMammoth,SmithsonianMammoth}. By adjusting parameters within each of the algorithms (except PaCMAP, which dynamically adjusts its own parameters), we attempt to go from local structure preservation to global structure preservation. Here, each algorithm exhibits its typical characteristics. t-SNE, and its more modern sister LargeVis \citep{Tang16}, produces spurious clusters, making its results untrustworthy. UMAP is better but also has trouble with global structure preservation. TriMap performs very well on global structure, but cannot handle local structure preservation (without modifications to the algorithm), whereas PaCMAP seems to preserve both local and global structure. A counterpoint is provided by the MNIST dataset \citep{lecun2010mnist} in Figure~\ref{fig:mnist}, which has a natural cluster structure (unlike the Mammoth data), and global structure is less important. Here, the methods that preserve local structure tend to perform well (particularly UMAP) whereas TriMap (which tends to maintain global structure) struggles. PaCMAP rivals UMAP's local structure on MNIST, and rivals TriMap on maintaining the mammoth's global structure, using its default parameters. 

From these few figures, we already gain some intuition about the different behavior of these algorithms and where their weaknesses might be. One major benefit of working on dimension-reduction methods is that the results can be visualized and assessed qualitatively. As we show experimentally within this paper, the qualitative observations we make about the algorithms' performance correlate directly with quantitative local-and-global structure preservation metrics: in many cases, qualitative analysis, similar to what we show in Figure~\ref{fig:Mammoth}, is sufficient to assess performance qualities. In other words, with these algorithms, what you see is actually what you get. 





Our main contributions are: (1) An understanding of key elements of dimension reduction algorithms: the loss function, and the set of graph components. (2) A set of principles that good loss functions for successful algorithms have obeyed, unified in the rainbow figure, which visualizes the mechanisms behind different DR methods. (3) A new loss function (PaCMAP's loss) that obeys the principles and is easy to work with. (4) Guidelines for the choice of graph components (neighbors, mid-near pairs, and further points). (5) An insight that initialization has a surprising impact on several DR algorithms that influences their ability to preserve global structure. (6) A new algorithm, PaCMAP, that obeys the principles outlined above. PaCMAP preserves both local and global structure, owing to a dynamic choice of graph components over the course of the algorithm.

\section{Related Work}





There are two primary types of approaches to DR for visualization, commonly referred to as local and global methods. The key difference between the two lies in whether they aim to preserve the global or local structure in the high-dimensional data. While there is no single definition of what it means to preserve local or global structure, local structure methods aim mainly to preserve the set of high-dimensional \textit{neighbors} for each point when reducing to lower dimensions, as well as the relative distances or rank information among neighbors (namely, which points are closer than which other points). Global methods aim mainly to preserve relative distances or rank information between points, resulting in a stronger focus on the preservation of distances between points that are farther away from each other. As an example to explain the difference, we might consider a dataset of planets and moons collected from a planetary system. A local method might aim to preserve relative distances between the moons and the nearest planet, without considering relative distances between planets (it is satisfactory if all planets are far from each other); a global method might aim primarily to preserve the placement of planets, without regard to the exact placement of moons around each of the planets. Neither approach is perfect in this example but both convey useful information. In the global approach, even if some of each planets' moons were crowded together, the overall layout is better preserved. On the other hand, the local approach would fail to preserve the relative distances between planets, but would be more informative in displaying the composition of each cluster (a planet and its surrounding moons).

Prominent examples of global methods include PCA \citep{Pearson01} and MDS \citep{Torgerson52}. The types of transformations these methods use cannot capture complex nonlinear structure. More traditional methods that focus on local structure include LLE \citep{Roweis00}, Isomap \citep{Tenenbaum00}, and Laplacian Eigenmaps \citep{Belkin01}, and more modern methods that focus on preserving local structure are t-SNE \citep{van_der_Maaten08}, LargeVis \citep{Tang16}, and UMAP \citep{UMAP} \citep[see][for a survey of these methods and how to tune their parameters]{DRReview09,DRReview13,Belkina19OptSNE}. We will discuss these methods shortly.


More broadly, data embedding methods, which are often concerned with reducing the dimension of data to more than 3 dimensions, could also be used for visualization if restricted to 2 to 3 dimensions. 
These, however, are generally only employed to improve the representation of the data with the goal of improving predictions, and not for the purpose of visualization. 
For example, deep representation learning is one class of algorithms for embedding \citep[see, e.g., the Variational Autoencoder of][]{VAE} that can also be used for DR.
While embedding is conceptually similar to DR, embedding methods do not tend to produce accurate visualizations and struggle to simultaneously preserve global and local structure (and are not designed to do so). Embedding methods are rarely used for visualization tasks (unless they are explicitly designed for such tasks), so we will not discuss them further here.  
Nevertheless, techniques, such as those involving triplets, are used for both embedding \citep{Chechik, HDML} and DR \citep{DrLIM,STE,TriMap}.

\textbf{Local structure preservation methods before t-SNE:}
Since global methods such as PCA and MDS cannot capture local nonlinear structure, 
researchers first developed methods that preserve raw local \textit{distances} (i.e., the values of the distances between nearest neighbors). Isomap \citep{Tenenbaum00}, Local Linear Embedding (LLE) \citep{Roweis00}, Hessian Local Linear Embedding \citep{Donoho03}, and Laplacian Eigenmaps \citep{Belkin01} all try to preserve local Euclidean distances from the original space when creating embeddings. 
These approaches were largely unsuccessful because distances in high-dimensional spaces tend to concentrate, and tend to be almost identical, so that preservation of raw distances did not always lead to preservation of neighborhood structure (i.e., points that are neighbors of each other in high dimensions are not always neighbors in low dimensions).
As a result, the field mainly moved away from preserving raw distances, and aimed instead to preserve graph structure (that is, preserve neighbors) instead.


\citet{Hinton03} introduced the Stochastic Neighborhood Embedding (SNE), which converts Euclidean distances between samples into conditional probabilities that represent similarities. It creates a low-dimensional embedding by enforcing the conditional probabilities in lower dimension to be similar to the conditional probabilities in the higher dimension. SNE is hard to optimize due to its asymmetrical loss function. \citet{cook07} improved SNE by turning conditional probabilities into joint probability distributions, so that the loss could be symmetric. Using an additional novel repulsive force term, the low-dimensional embedding is guaranteed to converge to a local optimum in a smaller number of iterations. \citet{elasembed} pointed out that the attractive force used in Laplacian Eigenmaps is similar to that used in SNE. \citet{elasembed} designed Elastic Embedding, which further improved the robustness of SNE. SNE retained neighborhoods better than previous methods, but still suffered from \textit{the crowding problem}: many points in the low-dimensional space land almost on top of each other. In addition, SNE requires a careful choice of hyperparameters and initialization.



\textbf{t-SNE and its variants:}
Building on SNE, t-Distributed Stochastic Neighbor Embedding (t-SNE) \citep{van_der_Maaten08} successfully handled the crowding problem by substituting the Gaussian distribution, used in the low-dimensional space, with long-tailed t-distributions.  Researchers \citep[e.g.,][]{Kobak19} later showed that a long-tailed distribution in a low-dimensional space could improve the quality of the embeddings. Different implementations of t-SNE, such as BH-SNE  \citep{Maaten2014AcceleratingTU}, viSNE \citep{Amir13}, FIt-SNE \citep{Linderman17FItSNE, Linderman19FItSNE}, and opt-SNE \citep{Belkina19OptSNE}, have further improved t-SNE so it could converge faster and scale better to larger datasets, while keeping the same basic conceptual ideas.
We briefly note that some research \citep[see, e.g.,][]{ProvetSNE} looked into the dynamics of the optimization process applied by t-SNE and showed that finding a good loss function for the graph layout is similar to finding a good loss function for a discrete dynamical system.

The success of t-SNE has inspired numerous variants. LargeVis \citep{Tang16} and the current state-of-the art, UMAP \citep{UMAP}, are the most well-known of these variants. With loss functions building on that of t-SNE, both methods improved efficiency (running time), K-Nearest Neighbor accuracy (a common measure for preservation of clusters in the DR process) and preservation of global structure (assessed qualitatively by manually and visually examining plots). However, all these methods are still subject to some intrinsic weaknesses of t-SNE, which includes loss of global structure and the presence of false clusters, as we will show later.

\textbf{Triplet constraint methods:}
Triplet constraint methods learn a low-dimensional embedding from triplets of points, with comparisons between points in the form of ``\textit{a is more similar to b than to c}.'' Approaches using triplets include DrLIM \citep{DrLIM}, t-STE \citep{STE}, SNaCK \citep{SNaCK} and TriMap \citep{TriMap}. These triplets generally come from human similarity assessments given by crowd-sourcing (with the exception of TriMap).  
We discuss TriMap extensively in this paper, as it seems to be the first successful triplet method that performs well on general dimension reduction tasks. Instead of using triplets generated by humans, it randomly samples triplets for each point, including some of that point's nearest neighbors in most triplets; by this construction, the triplets encourage neighbors to stay close, and further points to stay far.

 The idea of using triplets to maintain structure is appealing: one would think that preserving triplets would be an excellent way to maintain global structure, which is potentially why there have been so many attempts to use triplets despite the fact that they had not been working before TriMap. While we honestly believe that triplets are a good idea and truly would be able to preserve both local and global structure, we also have found a way to do that without modifying TriMap, and without using triplets.

Despite the numerous approaches to DR that we have described above, there are huge gaps in the understanding of the basic principles that govern these methods. In this work, we aim--without complicated statistical or geometrical derivations--to understand what the real reasons are for the successes and failures of these methods.

\textbf{Other related work:} A work performed in parallel with ours is that of \citet{Unify}, which presents a framework for DR algorithms that makes different points from ours. Their main result is that stronger attraction tends to create and preserve continuous manifold structures, while stronger repulsion leads to discrete cluster structures. This makes sense: if all neighbors are preserved (corresponding to large attractive forces), continuous manifolds are preserved. Similarly, if repulsive forces are increased, clusters of points that can be separated from others will tend to be separated.
\citet{Unify} show that UMAP embeddings correspond to t-SNE embeddings with increased attractive forces, and that another algorithm, ForceAtlas2 \citep{ForceAtlas2}, has even larger attractive forces. Our work makes different points: we design a set of principles that govern the relationship of attractive and repulsive forces on neighbors and further points; when these principles are not obeyed, local structure tends not to be preserved. One of the algorithms studied by \citet{Unify} (namely ForceAtlas2) does not obey one of our principles, and we present a more formal analysis in the appendix. Further, while we agree with \citet{Unify} that the \textit{amount} of attraction and repulsion does matter, we point out that \textit{which points} are attracted and repulsed is more important for global structure preservation. Specifically, in PaCMAP's design, we use highly-weighted \textit{mid-near pairs} to help preserve global structure.

Another recent work, that of \citet{ZhangSt21}, presents the idea of forceful colorings, and a mean-field model for t-SNE; it would be interesting to apply these techniques to PaCMAP in future work.



\section{t-SNE, UMAP, TriMap, and PaCMAP}\label{sec:review}


For completeness, we review the algorithms t-SNE, UMAP, TriMap in common notation in Sections \ref{sec:tsne}, \ref{sec:umap}, and \ref{sec:trimap}; readers familiar with these methods can skip these subsections. We also place a short description of PaCMAP in Section \ref{sec:pacmapshort}, though the full explanation of the decisions we made in PaCMAP follows in later sections, with a full statement of the algorithm in Section \ref{sec:pacmap}.

We use $\mathbf{X} \in \mathcal{R}^{N \times P}$ to denote the original dataset, for which we try to find an embedding  $\mathbf{Y}\in \mathcal{R}^{N \times 2}$.
For each observation $i\in[N]$, we use $\mathbf{x}_i$ and $\mathbf{y}_i$ to respectively denote its corresponding representations in $\mathbf{X}$ and $\mathbf{Y}$, where $\mathbf{x}_i$ is given and $\mathbf{y}_i$ is a 2-dimensional decision variable. We denote by $\mathrm{Distance}_{i,j}$ some distance metric in $\mathcal{R}^P$ between observations $\mathbf{x}_i$ and $\mathbf{x}_j$. Note that the definition of distance varies across methods.

\subsection{t-SNE}\label{sec:tsne}

\begin{enumerate}
    \item Consider two points $i$ and $j$. Define the conditional probability $p_{j|i}=\frac{e^{-\|\mathbf{x}_i-\mathbf{x}_j\|^2/2\sigma_i^2}}{\sum_{k\neq i}e^{-\|\mathbf{x}_i-\mathbf{x}_k\|^2/2\sigma_i^2}}$, 
    where $\sigma_i$ is computed by applying a binary search to solve the equation: $$\mathrm{Perplexity}=2^{-\sum_j p_{j|i}\log_2p_{j|i}}.$$ 
    The user-defined parameter $\mathrm{Perplexity}$ is  monotonically increasing in $\sigma_i$. Intuitively, as $\mathrm{Perplexity}$ becomes larger, the probabilities become uniform. When $\mathrm{Perplexity}$ becomes smaller, the probabilities concentrate heavily on the nearest points.
    
    \item Define the symmetric probability $p_{ij}=\frac{p_{i|j}+p_{j|i}}{2n}$. Observe that this probability does not depend on the decision variables $\mathbf{Y}$ and is derived from $\mathbf{X}$. 
    
    \item Define the probability $q_{ij}=\frac{\left(1+\|\mathbf{y}_i-\mathbf{y}_j\|^2\right)^{-1}}{\sum_{k\neq l} \left(1+\|\mathbf{y}_k-\mathbf{y}_l\|^2\right)^{-1}}$. Observe that this probability is a function of the decision variables $\mathbf{Y}$.
    
    \item Define the loss function as $\sum_i \sum_j p_{ij}\log\frac{p_{ij}}{q_{ij}}$.
    
    \item Initialization: initialize $\mathbf{Y}$ using the multivariate Normal distribution $\mathcal{N}(0,10^{-4}I)$, with $I$ denoting the two-dimensional identity matrix.
    
    \item Optimization: apply gradient descent with momentum. The momentum term is set to 0.5 for the first 250 iterations, and to 0.8 for the following 750 iterations. The learning rate is initially set to 100 and updated adaptively. 
\end{enumerate}

Intuitively, for every observation $i$, t-SNE defines a relative distance metric $p_{ij}$ and tries to find a low-dimensional embedding in which the relative distances $q_{ij}$ in the low-dimensional space match those of the high-dimensional distances (according to Kullback–Leibler divergence).
See \citet{van_der_Maaten08} for additional details and modifications that were made to the basic algorithm to improve its performance.

\subsection{UMAP}\label{sec:umap}
The algorithm consists of two primary phases: graph construction for the high-dimensional space, and optimization of the low-dimensional graph layout. The algorithm was derived using simplicial geometry, but this derivation is not necessary to understand the algorithm's steps or general behavior.
\begin{enumerate}
    \item Graph construction
    \begin{enumerate}
        \item For each observation, find the $k$ nearest neighbors for a given distance metric $\mathrm{Distance}_{i,j}$ (typically Euclidean).
        \item For each observation, compute $\rho_i$, the minimal positive distance from observation $i$ to a neighbor. 
        \item Compute a scaling parameter $\sigma_i$ for each observation $i$ by solving the equation: 
        $$\log_2(k)=\sum_{j=1}^{k}\exp\left({\frac{-\max\{0,\mathrm{Distance}_{i,j}-\rho_i\}}{\sigma_i}}\right).$$
        Intuitively, $\sigma_i$ is used to normalize the distances between every observation and its neighbors to preserve the relative high-dimensional proximities.
        \item Define the weight function $w(\mathbf{x}_i,\mathbf{x}_j)\eqdef \exp\left({\frac{-\max\{0,\mathrm{Distance}_{i,j}-\rho_i\}}{\sigma_i}}\right).$
        \item Define a weighted graph $G$ whose vertices are observations and where for each edge $(i,j)$ it holds that $\mathbf{x}_i$ is a nearest neighbor of $\mathbf{x}_j$, or vice versa. The (symmetric) weight of an edge $(i,j)$ is equal to 
        \begin{equation*}
        \bar{w}_{i,j} = w(\mathbf{x}_i,\mathbf{x}_j)+w(\mathbf{x}_j,\mathbf{x}_i)-w(\mathbf{x}_i,\mathbf{x}_j)\cdot w(\mathbf{x}_j,\mathbf{x}_i).
        \end{equation*}


    \end{enumerate}


    \item Graph layout optimization
    \begin{enumerate}
            \item Initialization: initialize $Y$ using spectral embedding. 
            
            \item Iterate over the graph edges in $G$, applying gradient descent steps on observation $i$ on every edge $(i,j)$ as follows:
        \begin{enumerate}

            \item Apply an attractive force (in the low-dimensional space) on observation $i$ towards observation $j$ (the force is an approximation of the gradient):
            $$
            \mathbf{y}_i = \mathbf{y}_i + \alpha\cdot
            \frac{-2ab\|\mathbf{y}_i-\mathbf{y}_j\|_2^{2(b-1)}}{1+a\left(\|\mathbf{y}_i-\mathbf{y}_j\|_2^{2}\right)^b}
             \bar{w}_{i,j}(\mathbf{y}_i-\mathbf{y}_j).
            $$
            The hyperparameters $a$ and $b$ are         tuned using the data by fitting the function $\left(1+a\left(\|\mathbf{y}_i-\mathbf{y}_j\|_2^{2}\right)^b\right)^{-1}$ to the non-normalized weight function $\exp\left({-\max\{0,\mathrm{Distance}_{i,j}-\rho_i\}}\right)$ with the goal of creating a smooth approximation. The hyperparameter $\alpha$ indicates the learning rate.
            
            \item Apply a repulsive force to  observation $i$, pushing it away from observation $k$. Here, $k$ is chosen by randomly sampling non-neighboring vertices, that is, an edge $(i,k)$ that is not in $G$. This repulsive force is as follows:
            $$
            \mathbf{y}_i = \mathbf{y}_i + \alpha\cdot
            \frac{b}{\left(\eps+\|\mathbf{y}_i-\mathbf{y}_k\|_2^2\right)\left(1+a\left(\|\mathbf{y}_i-\mathbf{y}_k\|_2^{2}\right)^b\right)}
            \left(1- \bar{w}_{i,k}\right)(\mathbf{y}_i-\mathbf{y}_k),
            $$
            where $\eps$ is a small constant.
        \end{enumerate}

    \end{enumerate}
\end{enumerate}

Intuitively, the algorithm first constructs a weighted graph of nearest neighbors, where weights represent probability distributions.
The optimization phase can be viewed as a stochastic gradient descent on individual observations, or as a force-directed graph layout algorithm \citep[e.g., see][]{gibson2013survey}. The latter refers a family of algorithms for visualizing graphs by iteratively moving points in the low-dimensional space closer that are close to each other in the high-dimensional space, and pushing apart points in the low-dimensional space that are further away from each other in the high-dimensional space.

In the UMAP paper \citep{UMAP}, the authors discuss several variants of the algorithm, as well as finer details and additional hyperparameters. The description provided here mostly adheres to Section~3 of \citet{UMAP} with minor modifications to the denominators based on the implementation of UMAP \citep{umapcode}. We also note that in the actual implementation, the edges are sampled sequentially with probability $\bar{w}(\mathbf{x}_i,\mathbf{x}_j)$ rather than selected uniformly, and the repulsive force sets $\bar{w}(\mathbf{x}_i,\mathbf{x}_k)=0$.

\subsection{TriMap}\label{sec:trimap}

\begin{enumerate}
    \item Find a set of triplets $\mathcal{T}:=\{(i,j,k)\}$ such that $\mathrm{Distance}_{i,j}<\mathrm{Distance}_{i,k}$:
    \begin{enumerate}
        \item For each observation $i$, find its 10 nearest neighbors (``n\_inliers'') according to the distance metric $\mathrm{Distance}$.  
        \item For every observation $i$ and one of its neighbors $j$, randomly sample 5 observations $k$ that are not neighbors (``n\_outliers'') to create 50 triplets per observation. 
        \item For each observation $i$ randomly sample 2 additional observations to create a triplet. Repeat this step 5 times (``n\_random''). 
        \item The set $\mathcal{T}$ consists of $|\mathcal{T}|=55\cdot N$ triplets.
    \end{enumerate}
    
    \item Define the loss function as $\sum_{(i,j,k)\in\mathcal{T}}l_{i,j,k},$ where $l_{i,j,k}$ is the loss associated with the triplet $(i,j,k)$:
    \begin{enumerate}
        \item $l_{i,j,k}=\omega_{i,j,k}\frac{s(\mathbf{y}_i,\mathbf{y}_k)}{s(\mathbf{y}_i,\mathbf{y}_j)+s(\mathbf{y}_i,\mathbf{y}_k)}$ where $\omega$ is statically assigned based on the distances within the triplet (mostly by $\mathrm{Distance}_{i,j}$ and $\mathrm{Distance}_{i,k}$), and the fraction is computed based on the low-dimensional representations $\mathbf{y}_i$, $\mathbf{y}_j$, and $\mathbf{y}_k$. Here, $s(\mathbf{y}_i,\mathbf{y}_j)$ is defined below in (c).
        
        \item Specifically, $\omega_{i,j,k}$ is defined as follows:
        $$
        \omega_{i,j,k} = \log\left(1+500\left(\frac{e^{d^2_{i,k}-d^2_{i,j}}}{\max_{(i',j',k')\in\mathcal{T}} e^{d^2_{i',k'}-d^2_{i',j'}}}+10^{-4}\right)\right),
        $$
        where $d^2_{i,j}=\mathrm{Distance}_{i,j}^2/{\sigma_i \sigma_j}$, and $\sigma_i$ is the average distance between $\mathbf{x}_i$ and the set of its 4--6 nearest neighbors. With a slight abuse of notation, later in the work we define $d^2_{i,j}$ in a close but slightly different way. 
        
        Intuitively, $\omega_{i,j,k}$ is larger when the distances in the tuple are more significant, suggesting that it is more important to preserve this relation in the low-dimensional space.
        \item Finally, $s(\mathbf{y}_i,\mathbf{y}_j)$ is defined as:
        $$
        s(\mathbf{y}_i,\mathbf{y}_j)=\left(1+\|\mathbf{y}_i-\mathbf{y}_j\|^2\right)^{-1}.
        $$
        This construction, which is similar to expressions used in t-SNE and UMAP, approaches 1 when observations $i$ and $j$ are placed closer, and approaches 0 when the observations are placed further away. This implies that the fraction in the triplet loss function would approach the maximal value of 1 when $i$ is placed close to $k$ and far away from $j$ (which contradicts the definition of a triplet $(i,j,k)$ where $i$ should be closer to $j$ than to $k$). Otherwise, as $k$ moves further away, the loss approaches 0. 
    \end{enumerate}
    
    \item Initialization: initialize $\mathbf{Y}$ using PCA.
    \item Optimization method: apply full batch gradient descent with momentum using the delta-bar-delta method (400 iterations with the value of momentum parameter equal to 0.5 during the first 250 iterations and 0.8 thereafter).
\end{enumerate}

Intuitively, the algorithm tries to find an embedding that preserves the ordering of distances within a subset of triplets that mostly consist of 
a k-nearest neighbor and a further observation
with respect to a focal observation, and a few additional randomized triplets which consist of two non-neighbors. As can be seen in the description above, the algorithm contains many design choices that work empirically, but it is not evident which of these choices are essential to creating good visualizations. 


\subsection{PaCMAP}\label{sec:pacmapshort}
PaCMAP will be officially introduced in Section \ref{sec:pacmap}. In this section we will provide a short description, leaving the derivation for later on.

First, define the scaled distances between pairs of observations $i$ and $j$: $ d^{2,\textrm{select}}_{ij}=\frac{\|\mathbf{x}_{i}-\mathbf{x}_{j}\|^2}{\sigma_{ij}} \text{ and }  \sigma_{ij}=\sigma_{i}\sigma_{j}$,
    where $\sigma_i$ is the average distance between $i$ and its Euclidean nearest fourth to sixth neighbors. Do not precompute these distances, as we will select only a small subset of them and can compute the distances after selecting.
\begin{enumerate}
    \item Find a set of pairs as follows:
    \begin{itemize}
        \item Near pairs: Pair $i$ with its nearest $n_{NB}$ neighbors defined by the scaled distance $d^{2,\textrm{select}}_{ij}$. To take advantage of existing implementations of k-NN algorithms, for each sample we first select the $\min(n_{NB}+50,N)$ nearest neighbors according to the Euclidean distance and from this subset we calculate scaled distances and pick the $n_{NB}$ nearest neighbors according to the scaled distance $d^{2,\textrm{select}}_{ij}$ (recall that $N$ is the total number of observations).

        \item Mid-near pairs: For each $i$, sample 6 observations, and choose the second closest of the 6, and pair it with $i$. The number of mid-near pairs to compute is $n_{MN} = \lfloor \nneighbors \times MN\_{\ratio} \rfloor$. Default $MN\_{\ratio}$ is 0.5. 
        \item Further pairs: Sample non-neighbors. The number of such pairs is $n_{FP}=\lfloor \nneighbors \times FP\_{\ratio} \rfloor$, where default $FP\_{\ratio}=2$.
    \end{itemize} 
    \item Initialize $\mathbf{Y}$ using PCA.
    \item Minimize the loss below in three stages, where $\tilde{d}_{ab} = \|\mathbf{y}_a-\mathbf{y}_b\|^2+1$:
     \begin{eqnarray*}
    \textrm{Loss}^{\textrm{PaCMAP}}&=&
    w_{\neighbors}\cdot\sum_{i,j \text{ are neighbors}}\frac{\tilde{d}_{ij}}{10 + \tilde{d}_{ij}} + w_{MN}\cdot\sum_{i,k \text{ are mid-near pairs}}\frac{\tilde{d}_{ik}}{10000 + \tilde{d}_{ik}} \\
    &&  + w_{FP}\cdot\sum_{i,l \text{ are further points}}\frac{1}{1 + \tilde{d}_{il}}.    \end{eqnarray*}
    In the first stage (i.e., the first 100 iterations of an optimizer such as Adam), set $w_{\neighbors}=2, w_{MN}(t)=1000\cdot\left( 1-\frac{t-1}{100} \right) + 3\cdot \frac{t-1}{100}, w_{FP}=1$. (Here, $w_{MN}$ decreases linearly from 1000 to 3.)\\
    In the second stage (iterations 101 to 200), set $w_{\neighbors}=3, w_{MN}=3, w_{FP}=1$. \\
    In the third stage (iteration 201 to 450), $w_{\neighbors}=1, w_{MN}=0, w_{FP}=1$.
    \item Return $\mathbf{Y}$.
\end{enumerate}



\section{Principles of a Good Objective Function for Dimension Reduction, Part I: The Loss Function}\label{sec:principles_good_objective}

In this section, we discuss the principles of good objective functions for DR algorithms based on two criteria: (1) local structure: the low-dimensional representation preserves each observation's neighborhood from the high-dimensional space; and (2) global structure: the low-dimensional representation preserves the relative positions of neighborhoods from the high-dimensional space. Note that a high-quality local structure is sometimes necessary for high-quality global structure, for instance to avoid overlapping non-compact clusters of different classes. 

\begin{figure}[ht]
\begin{center}
\centerline{\includegraphics[width=0.95\columnwidth]{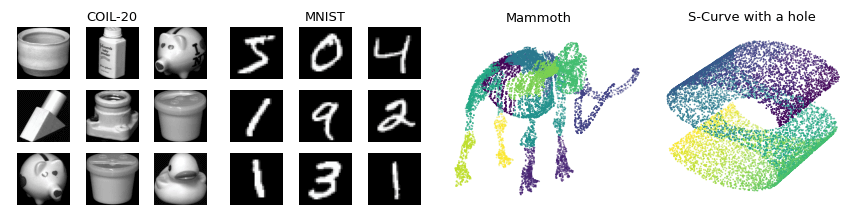}}
\caption{Datasets for local/global criteria. Samples for MNIST and COIL-20 dataset are provided on the left, while the full 3D datasets for S-curve-with-a-hole and Mammoth are directly shown.}
\label{fig:four_datasets}
\end{center}
\end{figure}

We use different datasets (shown in Figure~\ref{fig:four_datasets}) to demonstrate performance on these two criteria.
For the local structure criterion, we use the MNIST \citep{lecun2010mnist} and COIL-20 \citep{coil20} datasets, using the labels as representing the ground truth about cluster formation (the labels are not given as input to the DR algorithms). 
The qualitative evaluation criteria we use for assessing the preservation of local structure is whether clusters are preserved despite the fact DR takes place without labels. (Later in Section~\ref{sec:numericals} we use quantitative metrics; the qualitative measures directly reflect quantitative outcomes.)
For global structure, we use the 3D datasets S-curve and Mammoth in which the relative positions of different neighborhoods contain important information 
(additional information about these datasets can be found in Section~\ref{sec:numericals}). 

In Section~\ref{sec:unifiedobj}, we provide a generic unified objective for DR. In Section~\ref{sec:rainbow}, we will introduce the rainbow figure, which can be used to directly compare loss functions from different algorithms to understand whether they are likely to preserve local structure. Section~\ref{sec:principles_good_loss} provides the principles of loss functions for local structure preservation. Section~\ref{sec:sufficient} provides sufficient conditions for obeying the principles, namely when the loss is separable into a sum of additive and repulsive forces obeying specific technical conditions. Finally, in Section \ref{sec:apply_principle_to_pacmap_loss} we present PaCMAP's loss.

\subsection{A unified DR objective}\label{sec:unifiedobj}

All of the algorithms for DR that we discuss can be viewed as graph-based algorithms where a weighted graph is initially constructed based on the high-dimensional data (in particular, the distances in the high-dimension). 
In such graphs, nodes correspond to observations and edges signify a certain similarity between nodes which is further detailed by edge weights.
Thereafter, a loss function based on the graph structure is defined and optimized to find the low-dimensional representation. 
If we respectively denote by $\mathcal{C}^{H}_i$ and $\mathcal{C}^{L}_i$ two corresponding graph components (e.g., a graph component could be an edge, or a triplet of three nodes) in the graphs corresponding to the high (indicated by $H$) and low (indicated by $L$) dimensional spaces, these algorithms minimize an objective function of the following form: 


\[
\sum_{\textrm{Subset of graph components } \{i\}} \textrm{Loss}_{i} 
=
\sum_{\textrm{Subset of graph components }  \{i\}}  
\textrm{Weight}^{\mathbf{X}}(\mathcal{C}^{H}_i)\cdot \textrm{Loss}^{\mathbf{Y}}(\mathcal{C}^{L}_i),
\]
where $\textrm{Loss}_{i}$ refers to the overall loss associated with component $i$, decomposed into the high- and low-dimensional terms $\textrm{Weight}^{\mathbf{X}}(\mathcal{C}^{H}_i)$ and $\textrm{Loss}^{\mathbf{Y}}(\mathcal{C}^{L}_i)$, respectively.

Table~\ref{tbl:graphs} lists several DR algorithms, and breaks their objectives down in terms of which graph components they use and what their loss functions are. 
Note that the definition of the loss functions in the considered algorithms is intertwined with the graph weights (e.g., weights of graph components can be thought of as either a property of the loss or the graph components).

\begin{table}[htb]
\begin{center}
\begin{tabular}{ |c|l| } 
 \hline
 Algorithm    & Graph components and Loss function \\ 
 \hline
 t-SNE  &    
 \makecell[l]{
 Graph components: Edges $(i,j)$\\
 $\textrm{Loss}^\textrm{t-SNE}_{i,j} =  p_{ij}\log\frac{p_{ij}}{q_{ij}}$, where 
    $q_{ij}=\frac{\left(1+\|\mathbf{y}_i-\mathbf{y}_j\|^2\right)^{-1}}{\sum_{k\neq l} \left(1+\|\mathbf{y}_k-\mathbf{y}_l\|^2\right)^{-1}}$\\
    where $p_{ij}$ is a function of $\mathbf{x}_i$, $\mathbf{x}_j$ and other $\mathbf{x}_{\ell}$'s.}\\ 
 \hline
 UMAP   & 
    \makecell[l]{
    Graph components: Edges $(i,j)$\\
    $\textrm{Loss}^\textrm{UMAP}_{i,j} =$
    $\begin{cases} 
     \bar{w}_{i,j}\log \left( 1+a\left(\|\mathbf{y}_i-\mathbf{y}_j\|_2^{2}\right)^b \right)^{-1}       & i,j \mbox{ neighbors}  \\
    \left(1-\bar{w}_{i,j}\right)\log \left(1 - \left( 1+a\left(\|\mathbf{y}_i-\mathbf{y}_j\|_2^{2}\right)^b \right)^{-1}\right)  & \mbox{otherwise, }
    \end{cases}$ \\ 
    where $\bar{w}_{i,j}$ is a function of $\mathbf{x}_i$, $\mathbf{x}_j$ and nearby $\mathbf{x}_{\ell}$'s.}\\
 \hline
    TriMap &
    \makecell[l]{
    Graph components: Triplets $(i,j,k)$  where $\textrm{Distance}_{i,j}\leq \textrm{Distance}_{i,k}$\\
    $\textrm{Loss}^\textrm{TM}_{i,j,k}=\omega_{i,j,k}\frac{s(\mathbf{y}_i,\mathbf{y}_k)}{s(\mathbf{y}_i,\mathbf{y}_j)+s(\mathbf{y}_i,\mathbf{y}_k)}$, where      $s(\mathbf{y}_i,\mathbf{y}_j)=\left(1+\|\mathbf{y}_i-\mathbf{y}_j\|^2\right)^{-1}$ \\ and $\omega_{i,j,k}$ is a function of $\mathbf{x}_i$, $\mathbf{x}_j$,  $\mathbf{x}_k$ and nearby points.}\\ 
 \hline
\end{tabular}
\caption{Comparison of graph components and loss functions. Constant terms (that are independent of the low-dimensional variables) are given in Section~\ref{sec:review}. Note that the actual implementation of UMAP approximates the derivative of this loss. These losses are difficult to compare by considering their functional forms; instead, the rainbow plot provides a useful way to compare loss functions.}
\label{tbl:graphs}
\end{center}
\end{table}

When gradient methods are used on the objective functions, the loss function induces forces (attraction or repulsion) on the positions of the observations in the low-dimensional space. It is therefore insightful to consider how the algorithms define these forces (direction and magnitude) and apply them to different observations, which we do in more depth in Section~\ref{sec:graph_component}. 
The objectives for the DR methods we consider (e.g., t-SNE, UMAP and TriMap), define for each point a set of other points to attract, and a set of further points to repulse. To illustrate the differences between the various methods,  we consider triplets of points $(i,j,k)$, where $i$ is attracting $j$ and repulsing $k$. Notation $d_{ij}\eqdef\norm{\mathbf{y}_i-\mathbf{y}_j}_2$ indicates the distance between two points in the low-dimensional space. The specific choices of which observations are attracted and repulsed are complicated, but at its core, the loss governs the balance between attraction and repulsion; each algorithm aims to preserve the structure of the high-dimensional graph and reduce the crowding problem in different ways. 


We find these losses difficult to compare directly: they each have a different functional form. The key instrument that we will use to compare the algorithms is a plot that we refer to as the ``rainbow figure,'' discussed next. 

\subsection{The rainbow figure for loss functions}\label{sec:rainbow}

The rainbow figure is a visualization of the \textit{loss and its gradients} for $(i,j,k)$ triplets, where $i$ and $j$ should be attracted to each other (typically they are neighbors in the high-dimension), and $i$ and $k$ are to be repulsed (typically further points in the high-dimension). While the edges $(i,j)$ and $(i,k)$ need not be neighbors and further points, to build intuition and for ease of presentation, we refer to them as such.
Note that this plot applies to local structure preservation, because it handles mainly a group of nearest neighbors and further points. (We note that t-SNE considers a group of nearest neighbors to keep close, although whether to attract or repulse them also depends on the comparison of distributions in low- and high- dimensional space; UMAP explicitly attracts only a group of nearest neighbors; TriMap mainly considers triplets that contain a k-nearest neighbor, and we will show in Figure~\ref{fig:trimap_with_or_without_random_triplets} that the rest of the triplets actually have little effect on the final results; PaCMAP has both a group of nearest neighbors and a group of mid-near points to attract, but here in the rainbow figures we only consider attraction on the group of nearest neighbors, just as what PaCMAP does in the last phase of optimization--discussed below--where it focuses on refining local structure, in order to be consistent with other algorithms we discuss here.)

Since plotting the rainbow figures requires loss functions for $(i,j,k)$ triplets, we include such losses for each of t-SNE, UMAP,  TriMap, and PaCMAP (which we call ``good'' losses as the first 3 methods are widely considered to be very effective methods) in Table~\ref{tbl:graphs}. We ignore weights in t-SNE, UMAP, TriMap for simplicity. PaCMAP uses uniform weights.
We also note that while UMAP is motivated by the loss shown in Table~\ref{tbl:graphs}, in practice it applies an approximation to the gradient of that loss (see Section~\ref{sec:umap}). In what follows, we numerically compute and present the actual loss that UMAP implicitly uses.

Note that even though the rainbow figure visualizes triplets, it is not necessarily the case that the loss is computed using triplets. In fact, TriMap is the only algorithm we consider that uses a loss explicitly based on triplets. Also note that t-SNE does not fix pairs of points on which to apply attractive and repulsive forces in advance. It determines these over the course of the algorithm. The other methods choose points in advance.

In addition to the ``good'' losses, we show rainbow figures for four ``bad'' losses, that we (admittedly) had constructed in aiming to construct good losses. These bad losses correspond to four DR algorithms that fail badly in experiments; these bad losses will allow us to see what properties are valuable when considering the losses of triplets for DR algorithms. The bad losses are as follows:
\begin{itemize}
    \item BadLoss 1:\;\;\; $\textrm{Loss}_{i,j,k}^{1} = \log \left(1+e^{\frac{\left(d_{ij}^2 - d_{ik}^2\right)}{10}}\right)$;  (the constant 10 was used for numerical stability).
    \item BadLoss 2:\;\;\; $\textrm{Loss}_{i,j,k}^{2} = \frac{d_{ij}^2+1}{d_{ik}^2+1}$.
    \item BadLoss 3:\;\;\; $\textrm{Loss}_{i,j,k}^{3} = -\frac{d_{ik}^2+1}{d_{ij}^2+1}$.
    \item BadLoss 4:\;\;\; $\textrm{Loss}_{i,j,k}^{4} = \log \left(1+e^{d_{ij}^2} -e^{d_{ik}^2}\right)$.
\end{itemize}
Each of the bad losses fits some of the principles of good loss functions (considered in the next subsection), but not all of them.
The choice of the particular loss functions was guided by exploring additive and multiplicative relations between the attractive and repulsive forces, as well as applying the logarithm and exponential functions to moderate the impact of large penalties.

\begin{figure}[t]
\begin{center}
\centerline{\includegraphics[width=0.9\columnwidth]{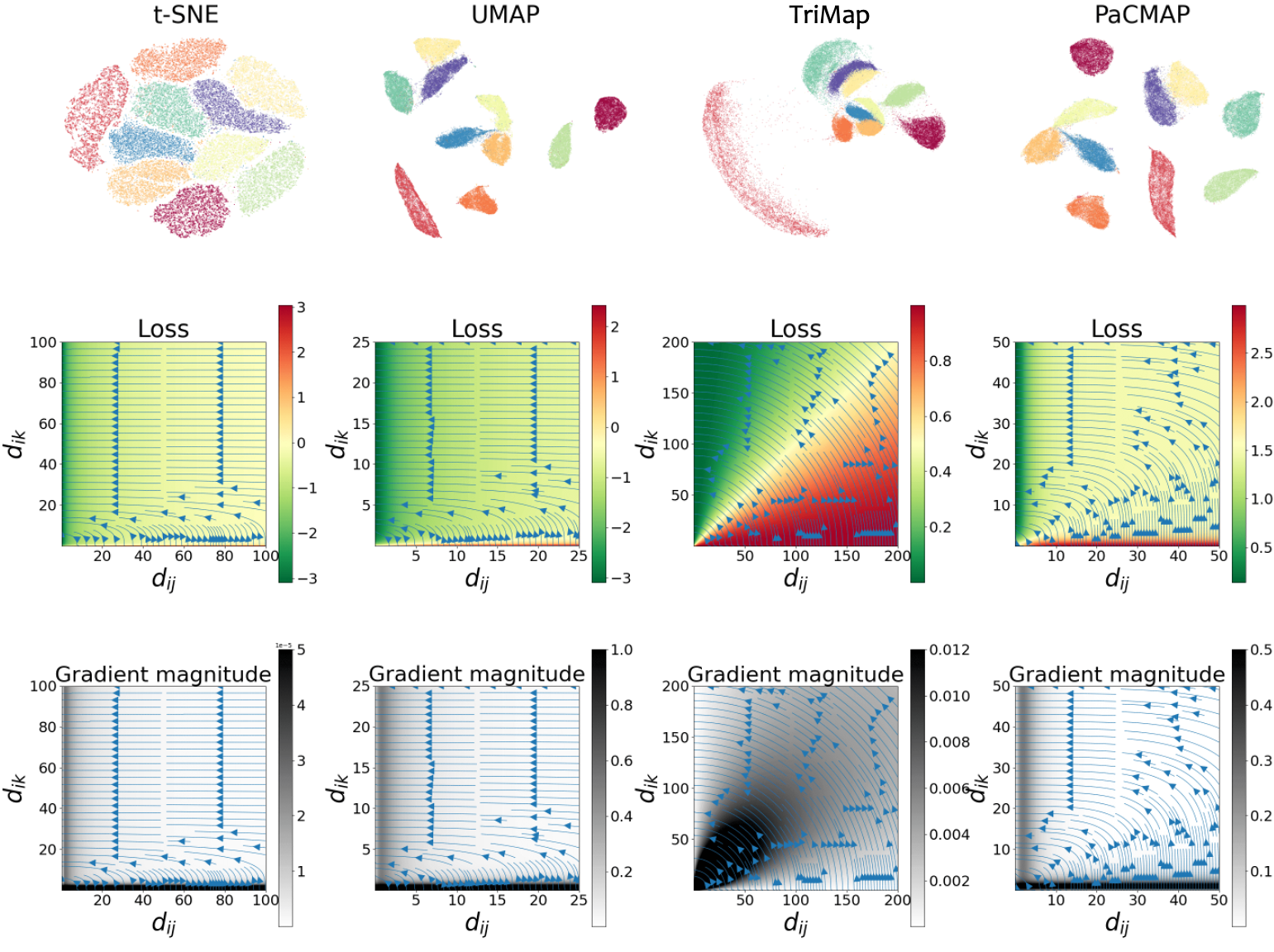}}
\caption{Visualizations of several good losses 
(comparing to other methods t-SNE uses many further points, each of which has a small impact on the total loss; to present its loss in a more salient way, we scaled t-SNE's loss so that the relative magnitude of attractive and repulsive forces are comparable).
}
\label{fig:good_loss_with_gradient}
\end{center}
\end{figure}
\begin{figure}[t]
\begin{center}
\centerline{\includegraphics[width=0.9\columnwidth]{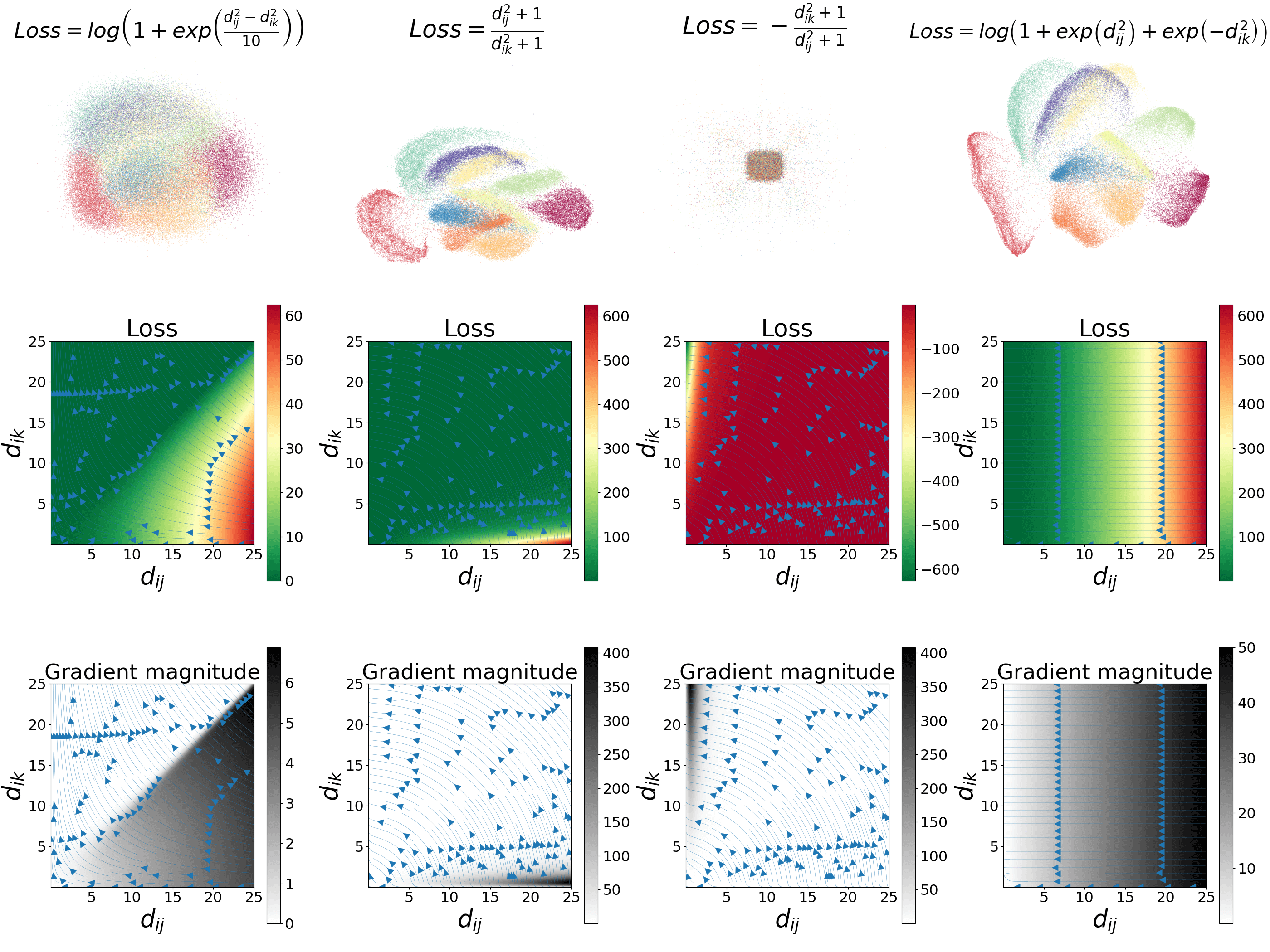}}
\caption{Visualizations of several bad losses. MNIST \citep{lecun2010mnist} is a useful dataset for testing DR methods: any loss function that cannot accommodate a good low-dimensional representation of MNIST would not achieve the minimum requirements for a good loss function.
}
\label{fig:bad_loss}
\end{center}
\end{figure}

The upper row of figures within Figures~\ref{fig:good_loss_with_gradient} and \ref{fig:bad_loss} show the outcome of each DR algorithm on the MNIST dataset. The middle figures of Figures~\ref{fig:good_loss_with_gradient} and \ref{fig:bad_loss} show rainbow figures for the loss functions. The magnitudes of the gradients are shown in the lower figures. Directions of forces are shown in arrows (which are the same in both middle and lower figures). The attractive pair $(i,j)$ of the triplet is shown on the horizontal axis, whereas the repulsive pair $(i,k)$ is shown on the vertical axis. Each point on the plot thus represents a triplet, with forces acting on it, depending on how far the points are placed away from each other: if $i$'s neighbor $j$ is far away, there is a large loss contribution for this pair (more 
yellow or red on the right). Similarly, if repulsion point $k$ is too close to $i$, the loss is large (more red near the bottom of the plot, sometimes the red stripe is too thin to be clearly seen). For each triplet, $d_{ij}$ and $d_{ik}$ are (stochastically) optimized following the gradient curves on the rainbow figure; $j$ is pulled closer to $i$, while $k$ is pushed away.

The rainbow figures are useful for helping us understand the tradeoffs made during optimization between attraction and repulsion among different neighbors or triplets. In particular, the rainbow figures of the good losses share many similarities, and differ from those of the bad losses. Several properties that induce good performance, which we have found to be common to the good losses, have been formalized as principles in the next section.


\subsection{Principles of good loss functions}\label{sec:principles_good_loss}

The rainbow figures provide the following insights about what constitutes a good loss function. These principles are correlated with good preservation of local structure: 

\begin{enumerate}
    \item Monotonicity. As expected, the loss functions are monotonically decreasing in $d_{ik}$ and are monotonically increasing in $d_{ij}$. That is, as a far point (to which repulsive forces are typically applied) becomes further away, the loss should become lower, and as a neighbor (to which attractive forces are typically applied) becomes further, the loss should become higher. This principle can be formulated as follows: 
    $$\forall d_{ij}:\frac{\partial Loss}{\partial d_{ik}}\leq 0 \text{ and } \forall d_{ik}:\frac{\partial Loss}{\partial d_{ij}}\geq 0.$$
\end{enumerate}

There should be asymmetry in handling the preservation of near and far edges. We have different concerns and requirements for neighbors and further points. The following three principles deal with asymmetry.

\begin{enumerate}
    \setcounter{enumi}{1} 
    
    
    \item Except in the bottom area of the rainbow figures (where $d_{ik}$ is small), the gradient of the loss should go mainly to the \textit{left}. If $d_{ik}$ is not too small (assuming that the further point $k$ is sufficiently far from $i$), we no longer encourage it to be even larger, and focus on minimizing $d_{ij}$. That is, the main concern for further points is to separate them from neighbors, which avoids the crowding problem and enables clusters to be separated (if clusters exist). There should be only small gains possible even if we push further points very far away. After $d_{ik}$ is sufficiently large, we focus our attention on minimizing $d_{ij}$ (i.e., pulling nearer points closer).
    Formally, 
    $$\forall d_{ij}, \forall \epsilon>0,\; \exists \theta^{\epsilon}_{ik} : \forall d_{ik}>\theta^{\epsilon}_{ik} \textrm{ we have }
    \left|
    \frac{\partial Loss}{\partial d_{ik}}/
    \frac{\partial Loss}{\partial d_{ij}}
    \right|
    <\eps.$$
      \\
    The advantage of this formulation of the principle (as well as the following ones) is that they are scale-invariant, and thus do not require quantification of ``large'' or ``small'' gradients. This means that they do not capture absolute magnitudes. For example, we observe empirically that when projecting the gradients on each axis, the mode of the repulsive force is smaller and has a higher peak, compared to the analogous plot of the attractive force (see, for example, Figures~\ref{fig:tsne_force} and \ref{fig:umap_loss_force_viz}).

    \item In the bottom area of rainbow figures, the gradient of the loss should mainly go \textit{up}, which increases $d_{ik}$. 
    (When $d_{ik}$ is very small, the further point is too close -- it should have been further away.)  One of possibly many ways to quantify this principle is as follows:
    $$\forall d_{ik}>0, \; \forall \epsilon>0,\;\exists \theta^{\epsilon}_{ij}: 
    \forall d_{ij}>\theta^{\epsilon}_{ij} 
     \textrm{ we have }
    \left|
    \frac{\partial Loss}{\partial d_{ij}}/
    \frac{\partial Loss}{\partial d_{ik}}
    \right| < \epsilon.$$

    
    \item Along the vertical axis (when $d_{ij}$ is very small), the \textit{magnitude of the gradient should be small}. That is, when $d_{ij}$ is small enough (when the near point is sufficiently close), there is little gain in further decreasing it, thus we should turn our attention to optimizing other points. This is not easy to see on the magnitude plot in Figure \ref{fig:good_loss_with_gradient}; it appears as a very thin stripe to the left of a thicker vertical stripe.
    Formally, $$\forall \epsilon>0, \;\exists \theta_{ik}^{\epsilon} : \forall d_{ik}\geq \theta_{ik}^{\epsilon},\;
    \lim_{d_{ij}\rightarrow 0} \left|\frac{\partial Loss}{\partial d_{ik}}\right|<\epsilon, \textrm{ and }
    \lim_{d_{ij}\rightarrow 0}\left|\frac{\partial Loss}{\partial d_{ij}}\right| <\epsilon.$$

    \item Along the horizontal axis (observed in 3 out of 4 methods when $d_{ik}$ is small), the \textit{magnitude of the gradient should be large}.  When $d_{ik}$ is too small (the further point is too close), it is pushed hard to become further away.
    Formally, $$\forall d_{ij}, \; \exists \theta_{ik}\; \forall d_{ik}>\theta_{ik}: \left|\frac{\partial Loss}{\partial d_{ik}}\right|^2 + 
    \left|\frac{\partial Loss}{\partial d_{ij}}\right|^2 \textrm{ is non-increasing in }d_{ik}.$$
 
   \item Gradients should be \textit{larger when $d_{ij}$ is moderately small, and become smaller when $d_{ij}$ is large.} 
   This reflects the fact that not all nearest neighbors can be preserved in the lower dimension and therefore the penalty for the respective violation should be limited. In a sense, we are essentially giving up on neighbors that we cannot preserve, suffering a fixed penalty for not preserving them.
    Formally, $$\forall d_{ik}:
    \left(\frac{\partial Loss}{\partial d_{ik}}\right)^2+\left(\frac{\partial Loss}{\partial d_{ij}}\right)^2 \text{ is unimodal in } d_{ij}.$$
    Also,
    $$\forall \epsilon>0,\; \exists \theta_{ik} : \forall d_{ik}\geq \theta_{ik},\;
    \lim_{d_{ij}\rightarrow \infty} \left|\frac{\partial Loss}{\partial d_{ik}}\right|
    <\epsilon, \textrm{ and }
    \lim_{d_{ij}\rightarrow \infty}\left|\frac{\partial Loss}{\partial d_{ij}}\right| <\epsilon.$$


\end{enumerate}

\begin{figure}[ht]
\begin{center}
\centerline{\includegraphics[width=0.3\columnwidth]{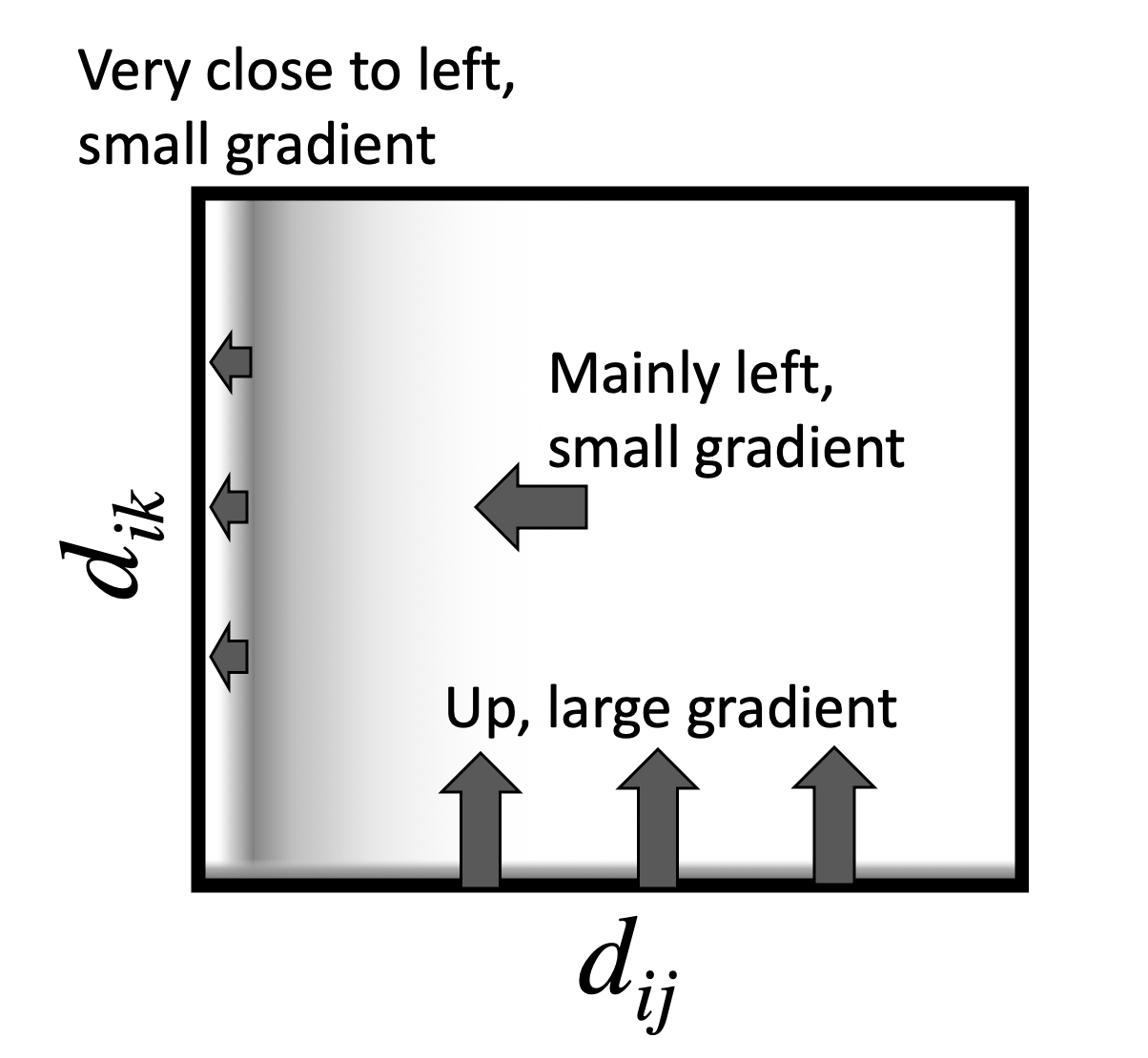}}
\caption{An illustration for the principles of a good loss function.}
\label{fig:cartoon_principles}
\end{center}
\end{figure}

An illustration of the principles is given in Figure~\ref{fig:cartoon_principles}. As we can see from the figure (and as we can conclude from the principles), the only large gradients allowed can be along the bottom axis, also in a region slightly to the right of the vertical axis.

Principles 2-6 arise from the inherently limited capacity of any low-dimensional representation. We cannot simply always pull neighbors closer and push further points away; not all of this information can be preserved. The algorithms instead prioritize what information should be preserved, in terms of which points to consider, and how to preserve their relative distances (and which relative distances to focus on).  We will consider the choice of points in Section \ref{sec:graph_component}.




Note that the rainbow figure of TriMap seems to violate one of our principles, namely Principle 5, that the gradient along the horizontal axis should be large. This means that two points that should be far from each other may not actually be pushed away from each other when constructing the low-dimensional representation. Interestingly, the algorithm's initialization may prevent this from causing a problem; if there are no triplets that include further points that are initialized close to each other, and neighbors that are initialized to be far away, then the forces on such triplets are irrelevant. We further discuss how TriMap obeys the principles in Section~\ref{sec:NNpair}.

Let us look at the bad loss functions, and find out what happens when the principles are violated. Recall that these bad loss functions were unsuccessful attempts to create good losses, so they each (inadvertently) obey some of the principles. All loss functions aim to minimize $d_{ij}$ and maximize $d_{ik}$ (thus obeying the first principle about monotonicity). One might think that monotonicity is enough, since it encourages attraction of neighbors and repulsion of far points, but it is not: the tradeoffs made by these loss functions are important.
\begin{itemize}
    \item BadLoss 1: This loss cannot separate the clusters that exist in the data. This could potentially stem, first, from a violation of Principle 3, where if further points are close (when they should not be), the loss effectively ignores them and focuses instead on pulling neighbors closer (see Figure~\ref{fig:bad_loss}). Second, in a violation of Principle 2, the gradient is essentially zero in the upper left of the rainbow figure, rather than going toward the left. This means there are almost no forces on pairs for which $d_{ij}$ is relatively small and $d_{ik}$ is fairly large; the loss essentially ignores many of the triplets that could help fine-tune the structure.
    
    \item BadLoss 2: This loss preserves more of the cluster structure than BadLoss 1, but still allows different clusters to overlap. This loss works better than BadLoss 1 because it does abide by Principles 3, 4, and 5, so that far points in the high-dimensional space are well separated and neighbors are pulled closer. However, due to the violation of Principle 6, the gradient does not sufficiently preserve local structure. In fact, we can see in Figure~\ref{fig:bad_loss} that the gradient increases with $d_{ij}$, which is visible particularly along the bottom axis. This means it tends to focus on fixing large $d_{ij}$ (global structure) at the expense of small $d_{ij}$ (local structure), where the gradient is small. 
    
    \item BadLoss 3: While this loss is seemingly similar to BadLoss 2 (it is equal to the inverse and negation of that loss), the performance of this loss is significantly worse, exhibiting a severe crowding problem. We believe this failure is the result of a violation of Principle 5: This loss has larger gradient for larger $d_{ik}$ values than for smaller $d_{ik}$ values. 
    That is, the algorithm makes little effort to repulse further points when $d_{ik}$ is small, so the points stick together. There is simply not enough force to separate points that should be distant from each other. One can see this from looking at the gradient magnitude plot of BadLoss 3 in Figure \ref{fig:bad_loss}, showing small gradient magnitudes along the whole bottom of the plot.
    Consequently, although the direction of the gradient is almost the same as that of BadLoss 2, the quality of the visualization result drops significantly.
    
    \item BadLoss 4: This loss is the most successful one in the four bad losses we presented here. Most of the clusters are well separated with a few exceptions of overlapping. The successful preservation of local structure could be credited to the obedience to Principles 2 and 4. These principles pull the neighbors together, but not too close. However, the violation of Principle 6 causes a bad trade-off among different neighbors. This loss makes great efforts to pull neighbors closer that are far away (large gradient when $d_{ij}$ is large), at the expense of ignoring the neighbors that are closer; if these nearer neighbors were optimized, they would better define the fine-grained structure of the clusters. Less attention to these closer neighbors yields non-compact clusters that sometimes overlap. Additional problems are caused by violations of Principles 3 and 5, implying that further points are not encouraged to be well-separated, which is another reason that clusters are overlapping.

\end{itemize}

\subsection{Sufficient conditions for obeying the principles}\label{sec:sufficient}
There are simpler ways to check whether our principles hold than to check them each separately. In particular:


\begin{proposition}
\label{prop1}
Consider loss functions of the form:
\[
Loss = 
\sum_{ij} Loss_{\textrm{attractive}}(d_{ij}) + 
\sum_{ik} Loss_{\textrm{repulsive}}(d_{ik}),
\]
where its derivatives are:
\[
f(d_{ij}):=\frac{\partial Loss_{\textrm{attractive}}(d_{ij})}{\partial d_{ij}}, \quad
g(d_{ik}):=-\frac{\partial Loss_{\textrm{repulsive}}(d_{ik})}{\partial d_{ik}}.
\]
Then, each of the six principles is obeyed under the following conditions on the derivatives:
\begin{enumerate}
    \item The functions $f(d_{ij})$ and $g(d_{ik})$ are non-negative and unimodal.
    \item $\lim_{d_{ij}\rightarrow 0}f(d_{ij})=\lim_{d_{ij}\rightarrow \infty}f(d_{ij})=0$, 
    $\lim_{d_{ik}\rightarrow 0}g(d_{ik})=\lim_{d_{ik}\rightarrow \infty}g(d_{ik})=0$.  
\end{enumerate}
\end{proposition}

\textbf{Proof.}
By construction, the loss associated with every triplet $(i,j,k)$, in which attractive force is applied to edge $(i,j)$ and where the repulsive force is applied to edge $(i,k)$ is simply $f(d_{ij})-g(d_{ik})$. We observe the following.
\begin{itemize}
    \item Principle 1: the monotonicity of the loss functions follows directly from the non-negativity of $f$ and $g$.

    \item Principle 2: for a fixed value of $d_{ij}$, the term $f(d_{ij})$ is constant. Moreover, since $\lim_{d_{ik}\rightarrow\infty}g(d_{ik})=0$, it follows that $g(d_{ik})/f(d_{ij})$ approaches zero as $d_{ik}$ increases.
    
    \item Principle 3: the principle follows from the assumption that $\lim_{d_{ij}\rightarrow \infty}f(d_{ij})=0$.
    
    \item Principle 4: the principle holds since  $\lim_{d_{ij}\rightarrow 0}f(d_{ij})=0$ and $\lim_{d_{ik}\rightarrow \infty}g(d_{ik})=0$.
    
    \item Principle 5: the principle follows from the unimodality of $g(d_{ik})$ and the additive loss function. 
    
    \item Principle 6: the first part follows from the unimodality of $f(d_{ij})$ and the additive loss function. The second part follows from the assumption that $\lim_{d_{ij}\rightarrow \infty}f(d_{ij})=0$ and $\lim_{d_{ik}\rightarrow \infty}g(d_{ik})=0$.
\end{itemize}
$\blacksquare$

\begin{proposition}
\label{prop2}
The UMAP algorithm with the hyperparameter $b>0.5$ satisfies the sufficient conditions of Proposition \ref{prop1}.
\end{proposition}
\textbf{Proof.}
For UMAP, the functions $f$ and $g$ can be written as (Section~\ref{sec:umap}): 
\[
f(d_{ij}) = const\cdot\frac{d_{ij}^{2b-1}}{1+ad_{ij}^{2b}} \;\;\textrm{ and }\;\;
g(d_{ik}) = const\cdot\frac{d_{ik}}{(\eps+d_{ik}^{2})(1+ad_{ik}^{2b})}.
\]
Note that here we have simplified the expression for $f$ in Section \ref{sec:umap}, so the vector ($\mathbf{y}_i-\mathbf{y}_j$) has become $\mathbf{y}_i-\mathbf{y}_j = \|\mathbf{y}_i-\mathbf{y}_j\|\cdot \vec{e}_{\mathbf{y}_i-\mathbf{y}_j}$ and the expression $\|\mathbf{y}_i-\mathbf{y}_j\|=d_{ij}$ was also included in the numerator of $f$ and $g$ just above.
It is easy to see that 
$\lim_{d_{ij}\rightarrow 0}f(d_{ij})=\lim_{d_{ij}\rightarrow \infty}f(d_{ij})=\lim_{d_{ik}\rightarrow 0}g(d_{ik})=\lim_{d_{ik}\rightarrow \infty}g(d_{ik})=0$, and that the functions $f$ and $g$ are non-negative. The functions are differentiable and have unique extremum points in $d_{ij}>0$ and $d_{ik}>0$, respectively, and are therefore unimodal. 
To see why the functions have single extremum points,  we compute their derivatives: 
$$f'(d_{ij})=-\dfrac{d_{ij}^{2b-2}\left(ad_{ij}^{2b}-2b+1\right)}{\left(1+ad_{ij}^{2b}\right)^2},$$ 
and 
$$g'(d_{ik})=-\dfrac{a{d_{ik}}^{2b}\left(\left(2b+1\right){d_{ik}}^2+
\epsilon(2b-1)
\right)+{d_{ik}}^2-\epsilon}{\left({d_{ik}}^2+\epsilon\right)^2\left(a{d_{ik}}^{2b}+1\right)^2}.$$
%
We already established that the functions $f$ and $g$ are non-negative, and approach zero when their argument goes to zero and infinity. To show unimodality, we show that the derivatives of these functions are positive up to some threshold, after which they turn negative. 

First, we observe that both denominators are positive. 
Second, the term $-\left(ad_{ij}^{2b}-2b+1\right)$ determines the sign of $f'(d_{ij})$, and it is monotonically decreasing in $d_{ij}$ and positive for small values of $d_{ij}$ when $b > 0.5$.
Third, the numerator of $g'(d_{ij})$ is monotonically decreasing in $d_{ik}$ and positive for small values of $d_{ik}$, both when $b>0.5$.
$\blacksquare$

We note that the hyperparameter $b$ of UMAP is derived from the hyperparameter min\_dist. Changing the value of min\_dist within its range [0,1] results in values of $b$ ranging from 0.79 to 1.92 (which satisfy Proposition~\ref{prop2}).



Later in Section~\ref{sec:NNpair} (see Equation~\eqref{eq:tsne_decomposition}) we discuss how t-SNE's loss can be decomposed into attractive and repulsive losses between every pair of observations. However, the attractive and repulsive losses contain a normalization term that ties together all observations, and thus cannot be computed by single $d_{ij}$ and $d_{ik}$, and therefore t-SNE's loss does not satisfy the conditions of Proposition~\ref{prop1}. 
Also note that TriMap cannot be decomposed and thus would not satisfy Proposition~\ref{prop1}. PaCMAP's loss (introduced below) satisfies Proposition~\ref{prop1} by construction.

\subsection{Applying the principles of a good loss function and introducing PaCMAP's loss}\label{sec:apply_principle_to_pacmap_loss}

PaCMAP uses a simple objective consisting of three types of pairwise loss terms, each corresponding to a different kind of graph component: nearest neighbor edges (denoted neighbor edges), mid-near edges (MN edges) and repulsion edges with further points (FP edges). We synonymously use the terms ``edges'' and ``pairs.''
\[\textrm{Loss}^{\textrm{PaCMAP}}=w_{\neighbors}\cdot\textrm{Loss}_{\neighbors} + w_{MN}\cdot\textrm{Loss}_{MN} +w_{FP}\cdot \textrm{Loss}_{FP},\]
where
$$
\textrm{Loss}_{\neighbors} = \sum_{i,j} \frac{\tilde{d}_{ij}}{10 + \tilde{d}_{ij}},\quad
\textrm{Loss}_{MN} = \sum_{i,k}  \frac{\tilde{d}_{ik}}{10000 + \tilde{d}_{ik}},\quad
\textrm{Loss}_{FP} = \sum_{i,l} \frac{1}{1 + \tilde{d}_{il}},$$ 
and $\tilde{d}_{ab}=d_{ab}^2+1=\|\mathbf{y}_{a}-\mathbf{y}_{b}\|^2 + 1$. 
The term $\tilde{d}$ was first used in the t-SNE algorithm, and inspired similar choices for UMAP and TriMap, which we also adopt here. The weights $w_{\neighbors}$, $w_{MN}$, and $w_{FP}$ are weights on the different terms that we discuss in depth later, as they are important.
The first two loss functions (for neighbor and mid-near edges) induce attractive forces while the latter induces a repulsive force. While the definition and role of mid-near edges are discussed in detail in Section~\ref{sec:graph_component}, we briefly note that mid-near pairs are used to improve global structure optimization in the early part of training. 

The exact functions we have used in the losses (simple fractions) could potentially be replaced with many other functions with similar characteristics: the important elements are that they are easy to work with, the attractive and repulsive forces obey the six principles, and that there is an attractive force on mid-near pairs that is separate from the other two forces.

\begin{figure}[ht]
\begin{center}
\centerline{\includegraphics[width=0.95\columnwidth]{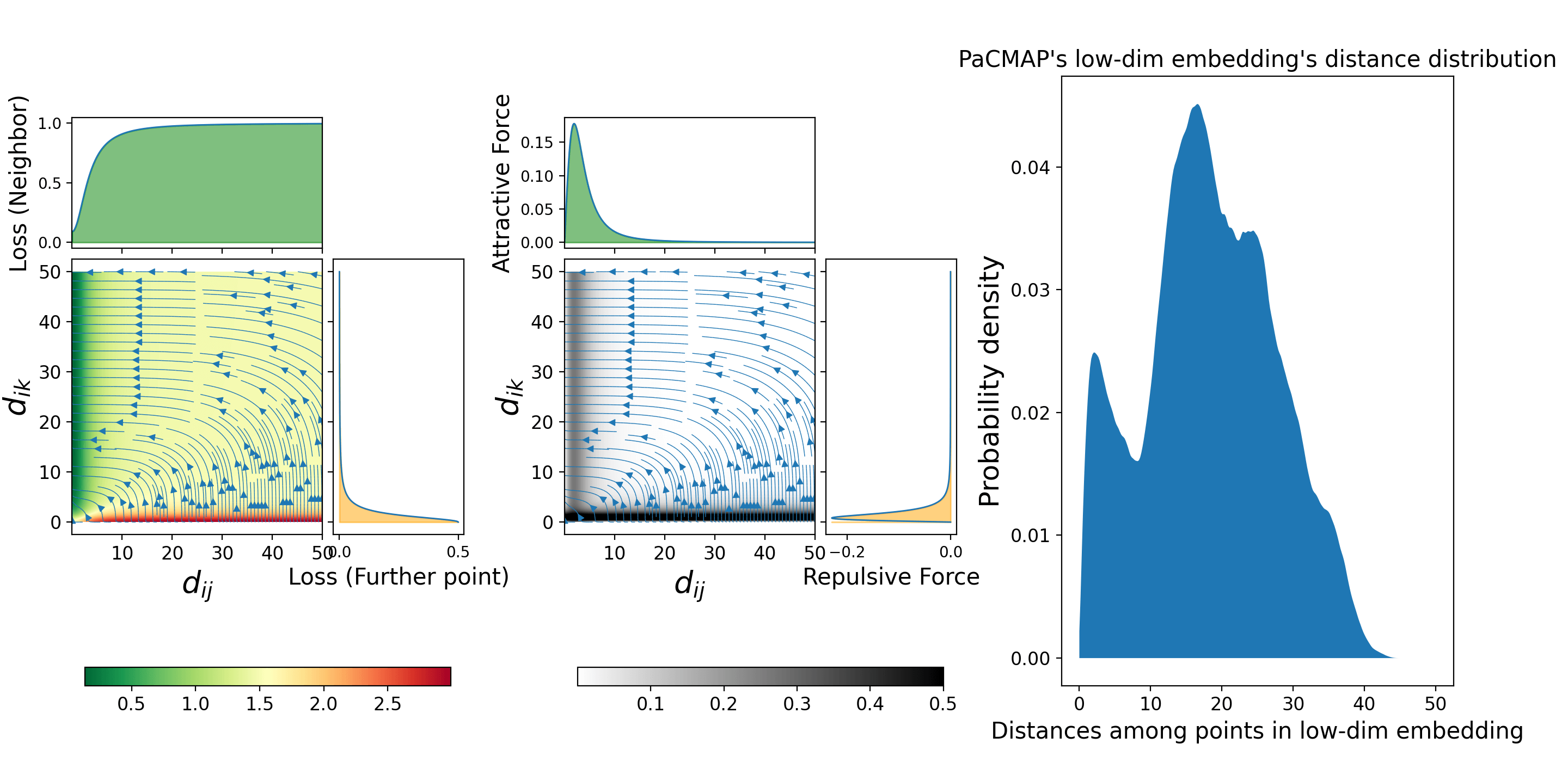}}
\caption{Similar to previous rainbow figures: (a) PaCMAP's losses ($\textrm{Loss}_{\neighbors}$ and $\textrm{Loss}_{FP}$); (b) PaCMAP's gradient magnitude (forces); (c) A histogram of the distances in the low-dimensional embedding attained by applying PaCMAP to the MNIST dataset.}
\label{fig:pacmap_loss_force_distri_viz}
\end{center}
\end{figure}

In what follows, we discuss how the above loss functions relate to the principles of good loss functions. For better illustration,  in Figure~\ref{fig:pacmap_loss_force_distri_viz} we visualize PaCMAP's losses on neighbors and further points (subplot a) and their gradients (subplot b) which can be seen as forces when optimizing the low-dim embedding. 
We see in Figure~\ref{fig:pacmap_loss_force_distri_viz}a that monotonicity holds for PaCMAP's loss, so Principle 1 is met. For a neighbor $j$, $\textrm{Loss}_{\neighbors}$ changes rapidly (large gradient/forces) when $d_{ij}$ is moderately small (Principle 6), and the magnitude of the gradient is small for extreme values of $d_{ij}$ (Principle 4). This is a result of the slow growth of the square function  $\tilde{d}_{ab}=d_{ab}^2+1$ that is used in the loss of PaCMAP, when $d_{ab}$ is very small (which encourages small gradients for very small $d_{ij}$ values), and the fact that $\textrm{Loss}_{\neighbors}$ saturates for large $d_{ij}$ (encourages slow growth for large $d_{ij}$ values). This can be seen in Figure~\ref{fig:pacmap_loss_force_distri_viz}b. These nice properties of $\textrm{Loss}_{\neighbors}$ enable PaCMAP to meet Principles 4 and 6. For further points, $\textrm{Loss}_{FP}$ induces strong repulsive force for very small $d_{ik}$, but the force dramatically decreases to almost zero when $d_{ik}$ increases. In this way, PaCMAP's loss meets the requirements of Principles 3 and 5. Compared to $\textrm{Loss}_{FP}$ where the gradient (force) is almost zero for $d_{ik}$ values that are not too small, $\textrm{Loss}_{\neighbors}$ induces an effective gradient (force) for a much wider range of $d_{ij}$. As a result of this trade-off, $\textrm{Loss}_{\neighbors}$ is dominant when $d_{ik}$ is not too small, implying that Principle 2 is obeyed.

While the principles of good loss functions are important, they are not the only key ingredient. We will discuss the other key ingredient below: the construction of the graph components that we sum over in the loss function.

\section{Principles of a Good Objective Function for Dimension Reduction, Part II: Graph Construction and Attractive and Repulsive Forces}\label{sec:graph_component}

Given that we cannot preserve all graph components, which ones should we preserve, and does it matter? In this section, we point out that \textit{the choice} of graph components is critical to preservation of global and local structure, and the balance between them. \textit{Which points} we choose to attract and repulse, and how we attract and repulse them, matter. If we choose only to attract neighbors, and if we repulse far points only when they get too close, we will lose global structure.

Interestingly, most of the DR algorithms in the literature have made choices that force them to be ``near-sighted.'' In particular, as we show in Section \ref{sec:NNpair}, most DR algorithms leave out graph components that preserve global structure. In Section \ref{sec:global_structure_and_midnear_edges}, we provide mechanisms to better balance between local and global structure through the choice of graph components.

\subsection{Graph component selection and loss terms in most methods mainly favors preservation of local structure}\label{sec:NNpair}





\begin{figure}[ht]
\begin{center}
\includegraphics[width=0.16\columnwidth,trim={1cm 0 0 1.5cm}]{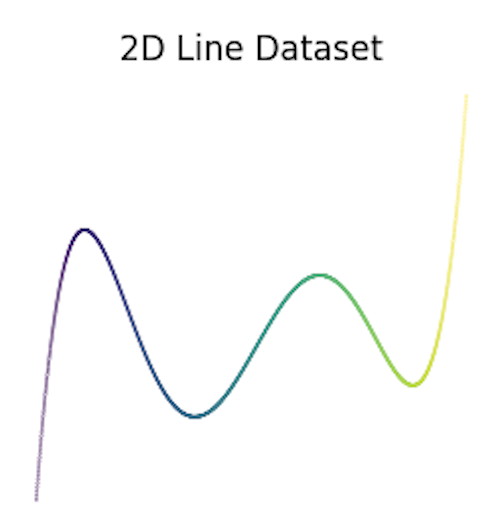}\;\;
\includegraphics[width=0.16\columnwidth,trim={1cm 0 0 1cm},clip]{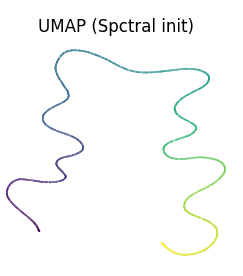}
\caption{
2D line dataset (left), and a projection also onto 2D that respects local structure but not global structure (right). The right curve (created using UMAP) would achieve perfect (zero) 0-1 triplet loss, as long as all triplets contain a nearby unit.
}
\label{fig:2D_curve_line}
\end{center}
\end{figure}

Let us try a thought experiment. Consider a triplet loss, as used in TriMap: 
\[  
\sum_{(i,j,k)\in\mathcal{T}}l_{i,j,k}.
\]
The triplets in $\mathcal{T}$ are the graph components. Also consider a simple triplet loss such as $l_{i,j,k}=1$ if $j$ is further away than $k$, and 0 otherwise (we call this the 0-1 triplet loss). Now, what would happen if every triplet in $\mathcal{T}$ contained at least one point very close to $i$, to serve as $j$ in the triplet? In other words, $\mathcal{T}$ omits all triplets where $j$ and $k$ are both far from $i$. In this case, we argue that the algorithm would completely lose its ability to preserve global structure. To see this, note that the 0-1 triplet loss is zero whenever all further points are further away than all neighbors. But this loss can be zero without preserving global structure. While maintaining zero loss, even simple topological structures may not be preserved. Figure \ref{fig:2D_curve_line} (right) shows an example of a curve generated from the data on Figure \ref{fig:2D_curve_line} (left) with an algorithm that focuses only on local structure by choosing triplets as we discussed (with at least two points near each other per triplet). Here the triplet loss is approximately zero, but global structure would not be preserved. Thus, although the loss is optimized to make all neighbors close and all further points far, this can be accomplished amidst a complete sacrifice of global structure.

Similarly, consider any algorithm that uses attractive forces and/or repulsive forces (e.g., t-SNE, UMAP), and attracts only neighbors and repulses only further points, as follows: 
\[
\sum_{(i,j)\in\mathcal{T}_{\neighbors}}l^{\textrm{attract}}(i,j)+\sum_{(i,k)\in\mathcal{T}_{\textrm{further}}}l^{\textrm{repulse}}(i,k),
\]
where $\mathcal{T}_{\neighbors}$ consists of neighbor pairs where attractive forces are applied and $\mathcal{T}_{\textrm{further}}$ consists of further-point pairs where repulsive forces are applied.
Again, consider a simple 0-1 loss where $l^{\textrm{attract}}(i,j)$ is 0 if $i$ is close to $j$ in the low-dimensional space, and that $l^{\textrm{repulse}}(i,k)$ is 0 if $i$ and $k$ are far. Again, even when the total loss is zero, global structure need not be preserved, by the same logic as in the above example.

From these two examples, we see that in order to preserve global structure, we must have \textit{forces on non-neighbors}. 
If this requirement is not met, the relative distances between further points do not impact the loss, which, in turn, sacrifices global structure. Given this insight, let us consider what DR algorithms actually do.


\textbf{Graph structure and repulsive forces for t-SNE}:  t-SNE attracts or repulses points by varying amounts, depending on the perplexity. The calculation below, however, shows that t-SNE's repulsive force has the issues we pinpointed: \textit{further points have little attractive or repulsive forces on them unless they are too close}, meaning that t-SNE is ``near-sighted.'' We start by looking at the gradient of t-SNE's objective with respect to movement in the low-dimensional space for point $i$:  
\begin{equation*}
\frac{\partial \textrm{Loss}^{\textrm{t-SNE}}}{\partial \mathbf{y}_i}=4\sum_j\left(p_{ij}-q_{ij}\right)\left(\mathbf{y}_i-\mathbf{y}_j\right)\left(1+\| \mathbf{y}_i-\mathbf{y}_j \|^2\right)^{-1}
\end{equation*}
where $j$ is one of the other points. 
For simplicity, we denote $\|\mathbf{y}_i - \mathbf{y}_j\|$ as $d_{ij}$, and the unit vector in the direction of $i$ to $j$ as $\mathbf{e}_{ij}$.
Recall that \[p_{ij}=
\frac{p_{j|i}+p_{i|j}}{2N} \textrm{ where } p_{j|i} = 
\frac{
e^{-\|\mathbf{x}_i-\mathbf{x}_j\|^2/2\sigma_i^2}
}
{
\sum_{k\neq i}
e^{-\|\mathbf{x}_i-\mathbf{x}_k\|^2/2\sigma_i^2}
},
\] 
and
\[q_{ij}=\frac{(1+d_{ij}^2)^{-1}}{\sum_{k \neq l}(1+d_{kl}^2)^{-1}},\] 
and $N$ is the number of points.

For a point $j$, we denote its force on $i$ (the $j$th term in the gradient above) as $\textrm{Force}_{ij}$. Following the separation of gradient given by \citet{Maaten2014AcceleratingTU},
\begin{eqnarray}
\textrm{Force}_{ij}&
=& 4\left(p_{ij}-q_{ij} \nonumber
\right)d_{ij}\left(1+d_{ij}^2\right)^{-1}{\mathbf{e}}_{ij}\\\label{twoterms}
&=&
4p_{ij}d_{ij}\left(1+d_{ij}^2\right)^{-1}{\mathbf{e}}_{ij}
- 4q_{ij}d_{ij}\left(1+d_{ij}^2\right)^{-1}{\mathbf{e}}_{ij}\label{eq:tsne_decomposition}\\
&=&F_{\textrm{attraction}} - F_{\textrm{repulsion}}.\nonumber
\end{eqnarray}
Since $p_{ij}$, $q_{ij}$ and $d_{ij}$ are always non-negative, the two forces $F_{\textrm{attraction}}$ and $F_{\textrm{repulsion}}$ are always non-negative, pointing in directions opposite to each other along the unit vector $\mathbf{e}_{ij}$.

\textit{Both of these forces decay rapidly to 0 when $i$ and $j$ are non-neighbors in the original space, and when $d_{ij}$ is sufficiently large.} Let us show this.

\textit{Attractive forces decay}: The attractive term depends on  $p_{ij}$, which is inversely proportional to the squared distance in the original space. Since we have assumed this distance to be large (i.e., $i$ and $j$ were assumed to be non-neighbors in the original space), its inverse is small. Thus the attractive force on non-neighbors is small.


For most of the modern t-SNE implementations \citep[for example,][]{Maaten2014AcceleratingTU, Linderman17FItSNE, Belkina19OptSNE}, $p_{ij}$ for most pairs is set to 0 within the code. In particular, if the perplexity is set to $\mathrm{Perplexity}$, the algorithm considers 3$\cdot\mathrm{Perplexity}$ nearest neighbors of each point. For points beyond these 3$\cdot\mathrm{Perplexity}$, the attractive force will be set to 0. This operation directly implies that there are no attractive forces on non-neighbors.

\textit{Repulsive forces decay}:
The repulsive force does not depend on distances in the original space, yet also decays rapidly to 0. To see this, temporarily define $a_{ij}:=\frac{1}{1+d_{ij}^2}$, and define $B_{ij}$ as $\sum_{k\neq l, kl \neq ij} (1+d_{kl}^2)^{-1}$. Simplifying the second term of (\ref{twoterms}) (substituting the definition of $q_{ij}$ as $a_{ij}/(a_{ij}+B_{ij})$), it is:
\[
4\frac{a_{ij}}{B_{ij}+a_{ij}} d_{ij}a_{ij}. 
\]
The fraction $\frac{a_{ij}}{B_{ij}+a_{ij}}$ is non-negative and at most 1, and generally much smaller for large $d_{ij}$ (i.e., small $a_{ij}$, where $a_{ij} = 1/(1+d_{ij}^2)$). The product $d_{ij}a_{ij}$ equals $d_{ij}/{(1+d_{ij}^2)}$. As $d_{ij}$ grows, the denominator grows as the square of the numerator, and thus the force decays rapidly to 0 at rate of $\Theta(1/d_{ij})$.

Further, the derivative of the forces smoothly decays to 0 as $d_{ij}$ grows.
Consider the partial derivative of the repulsion force with respect to the distance of a further pair in the low-dimensional embedding, which is:
$$\frac{\partial F_{\textrm{repulsion}}}{\partial d_{ij}} = -4\frac{B_{ij}d_{ij}^4 +(2B_{ij}+1)d_{ij}^2 - B_{ij} - 1}{(d_{ij}^2+1)^2  (B_{ij}d_{ij}^2 + B_{ij} + 1)^2}.$$
When $d_{ij}$ is large, the quartic term $B_{ij}d_{ij}^4$  will dominate the numerator, and the term $B_{ij}^2d_{ij}^8$ will dominate the denominator. Hence, when $d_{ij}$ is large, the force stays near 0.

\begin{figure}[hbt]
\begin{center}
\centerline{\includegraphics[width=0.95\columnwidth]{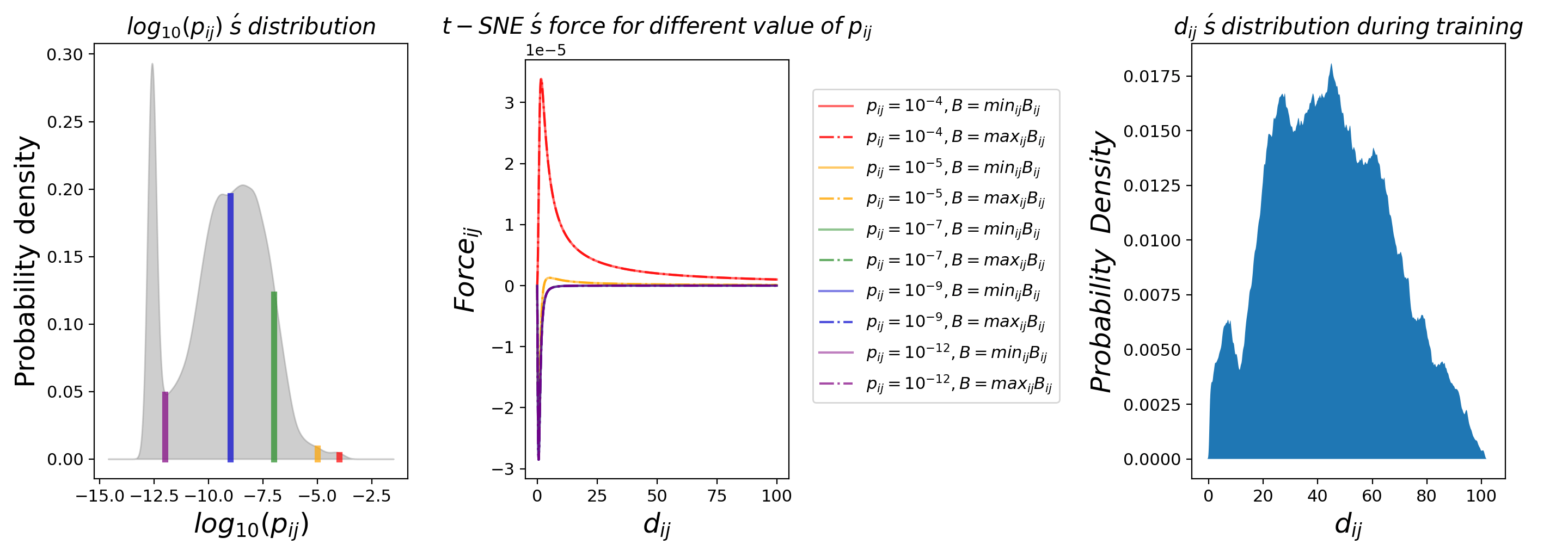}}
\caption{Forces on non-neighbors rapidly vanish for large $d_{ij}$. \textit{Left:} Distribution of $p_{ij}$ values. Since $p_{ij}$ is inversely proportional to the distances $\|\mathbf{x}_i-\mathbf{x}_j
\|$ in the original space, the neighbors are on the right of the plot. Neighbors are attracted and all other points are repulsed. Several $p_{ij}$ values were chosen (each represented by a different color), to represent the distribution of $p_{ij}$ in the data. \textit{Middle:} The largest $p_{ij}$ (in red) experiences an overall attractive (positive) force, the other $p_{ij}$ values  mainly experience an overall repulsive (negative) forces. Note that the forces visualized here consist of both attractive and repulsive components, and typically one dominates the other (depending on the pair's distance in the original space). \textit{The plot shows that the forces (both attractive and repulsive) decay quickly for non-neighbors as $d_{ij}$ increases.} \textit{Right:} The distribution of $d_{ij}$ at the 300th iteration of t-SNE is shown, illustrating that the horizontal axis in the center plot is representative of $d_{ij}$ values of the actual data.}
\label{fig:tsne_force}
\end{center}
\end{figure}

Now that we have asymptotically shown that t-SNE has minimal forces on non-neighbors, we also want to check what happens empirically. Figure~\ref{fig:tsne_force} provides a visualization of t-SNE's pairwise forces on the COIL-20 \citep{coil20} dataset. We can see that for different further pairs, the force applied on them is almost the same, and is vanishingly small. This means, in a similar way as discussed above, t-SNE cannot distinguish between further points, which explains why t-SNE loses its ability to preserve global structure.

Despite t-SNE favoring local structure, in its original implementation, it still calculates and uses distances between \textit{all} pairs of points, most of which are far from each other and exert little force between them. The use of all pairwise distances causes t-SNE to run slowly.
This limitation was improved in subsequent papers \citep[see, for example, ][as discussed earlier, using only 3$\cdot$Perplexity attraction points]{Amir13, Maaten2014AcceleratingTU, Linderman17FItSNE}. Overall, it seems that we are better off selecting a subset of edges to work with, rather than aiming to work with all of them simultaneously. 

\begin{figure}[ht]
\begin{center}
\centerline{\includegraphics[width=0.65\columnwidth]{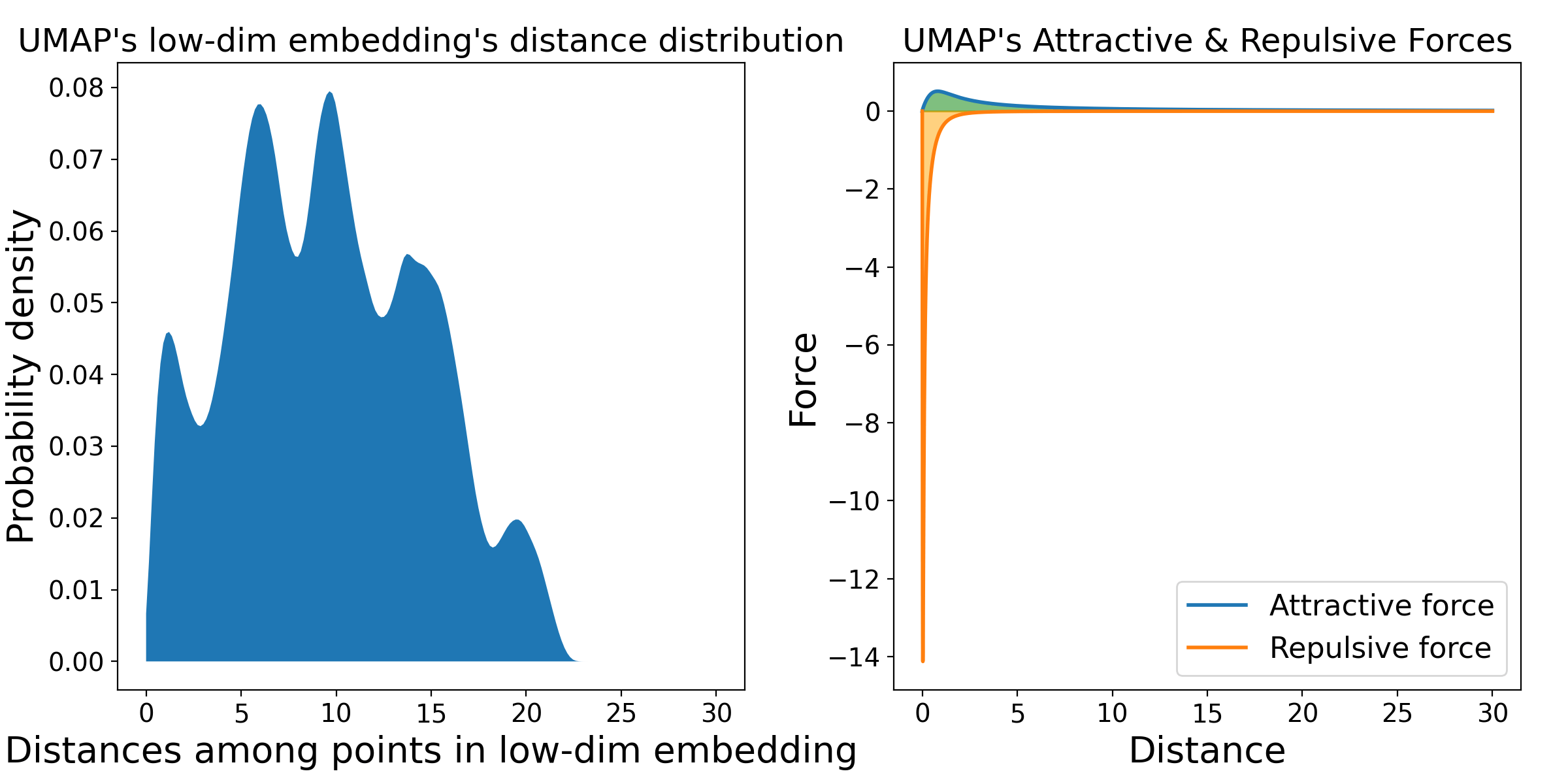}}
\caption{UMAP's gradients (forces) for neighbor edges and repulsion edges (right), and the distribution of distances in the low-dimensional embedding for visualizing the MNIST dataset (left). The right plot repulsion curve shows that the repulsion and attraction forces vanish as long as $j$ is sufficiently far from $i$; the repulsion force is mainly active if the repulsion point is too close. The left figure shows the distribution of pairwise distances after embedding, illustrating that many points are far away from each other. For instance, any further points that are 3 units or more from each other exert little repulsive or attractive forces; most of the pairs are at least this distance away. Although the embedding's distance distribution varies over different datasets, generally the distance range where repulsive forces are effective is narrow compared to the distance distribution of the embedding.} 
\label{fig:umap_loss_force_viz}
\end{center}
\end{figure}

\textbf{Graph structure and repulsive forces for UMAP}: UMAP's forces exhibit tendencies similar to those of t-SNE's. Its attractive forces on neighbors and its repulsive forces decay rapidly with distance. As long as a further point is sufficiently far from point $i$ in the low-dimensional space, the force is minimal.  
Figure \ref{fig:umap_loss_force_viz} (left) shows that most points are far enough away from each other that little pairwise force is exerted between them, either attractive or repulsive. UMAP has weights that scale these curves, however, the scaling does not change the shapes of these curves. 
Thus again, we can conclude that since UMAP does not aim to distinguish between further points, it will not necessarily preserve global structure.

\textbf{Graph structure and repulsive forces for TriMap}:  In the thought experiment considered above, if no triplets are considered such that $j$ and $k$ are both far from $i$, the algorithm fails to distinguish the relative distances of further points and loses global structure.

\begin{figure}[t]
\begin{center}
\centerline{\includegraphics[width=0.75\columnwidth]{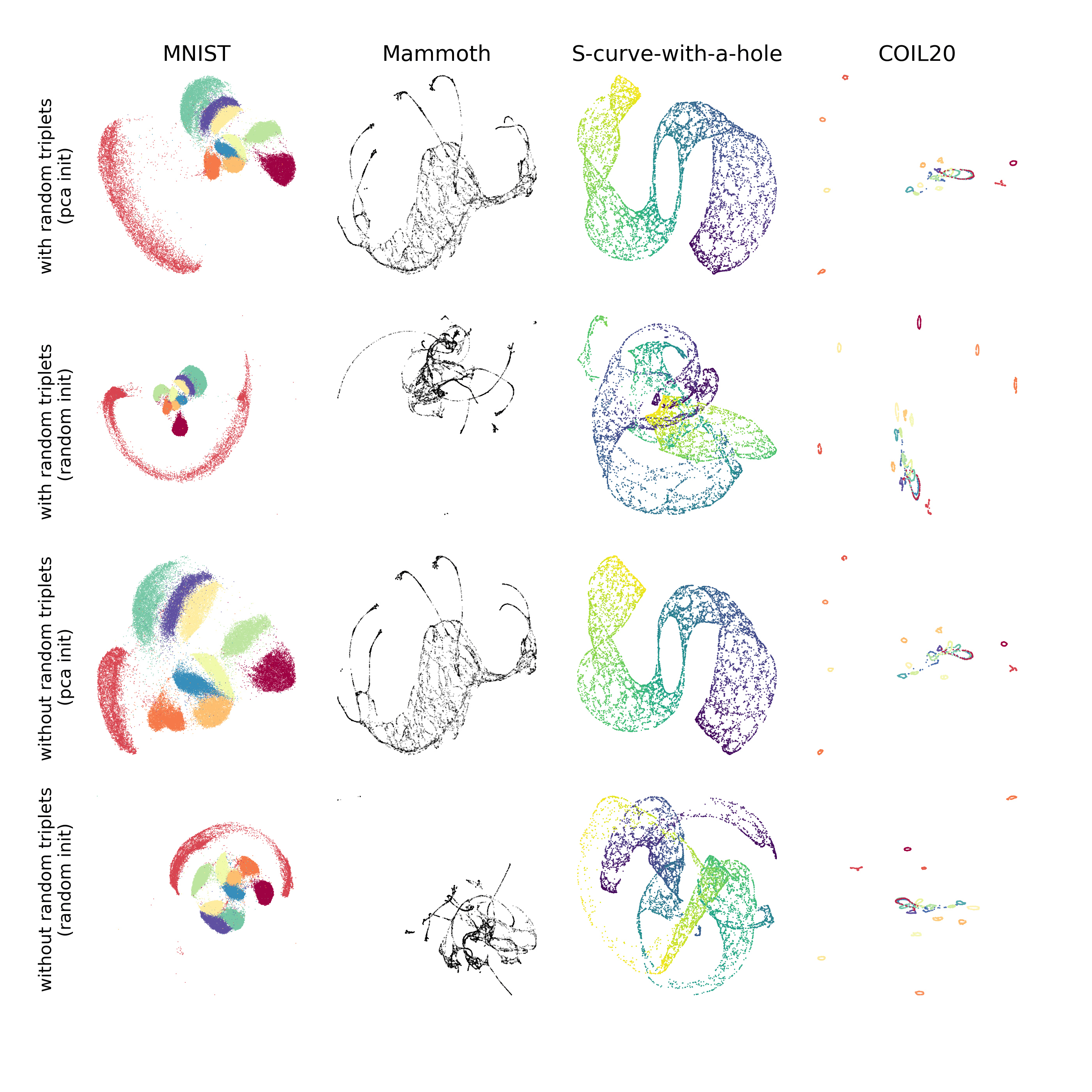}}
\caption{TriMap's performance with or without random triplets and PCA initialization. The second and fourth rows show that without PCA initialization, TriMap loses global structure.}
\label{fig:trimap_with_or_without_random_triplets}
\end{center}
\end{figure}

Approximately 90\% (55 triplets for each point, 50 of which involve a neighbor) of TriMap's triplets have $j$ as a neighbor of $i$. One would thus think its global structure preservation would arise from the 10\% of random triplets--which are likely comprised of points that are all far from each other--but its global structure preservation does not arise from these random triplets. Instead, its global structure preservation seems to arise from its use of PCA initialization. Without PCA initialization, TriMap's global structure is ruined, as shown in Figure \ref{fig:trimap_with_or_without_random_triplets}, where TriMap is run with and without the random triplets and PCA initialization. (More details are discussed in Section \ref{sec:initialization}). Given that the small number of TriMap's random triplets have little effect on the final layouts, in the discussion below, we consider triplets that contain a neighbor for analyzing how TriMap optimizes on neighbors and further points. 

Even though TriMap's rainbow figure does not look like that of the other loss functions, it still follows our technical principles, which are based on asymptotic properties. However, as discussed above, in spirit, Principle 5 should be concerned with pushing farther points away from each other that are very close, which corresponds to triplets that lie along the horizontal axis of the rainbow plot. Interestingly, TriMap's initialization causes Principle 5 to be somewhat unnecessary. After the PCA initialization (discussed above), there typically are no triplets on the bottom of the rainbow plot. This is because the PCA initialization tends to keep farther points separated from each other. In that way, TriMap manages to obey all of the principles, but only when the initialization allows it to avoid dealing with Principle 5.

\begin{figure}[ht]
\begin{center}
\centerline{\includegraphics[width=0.9\columnwidth]{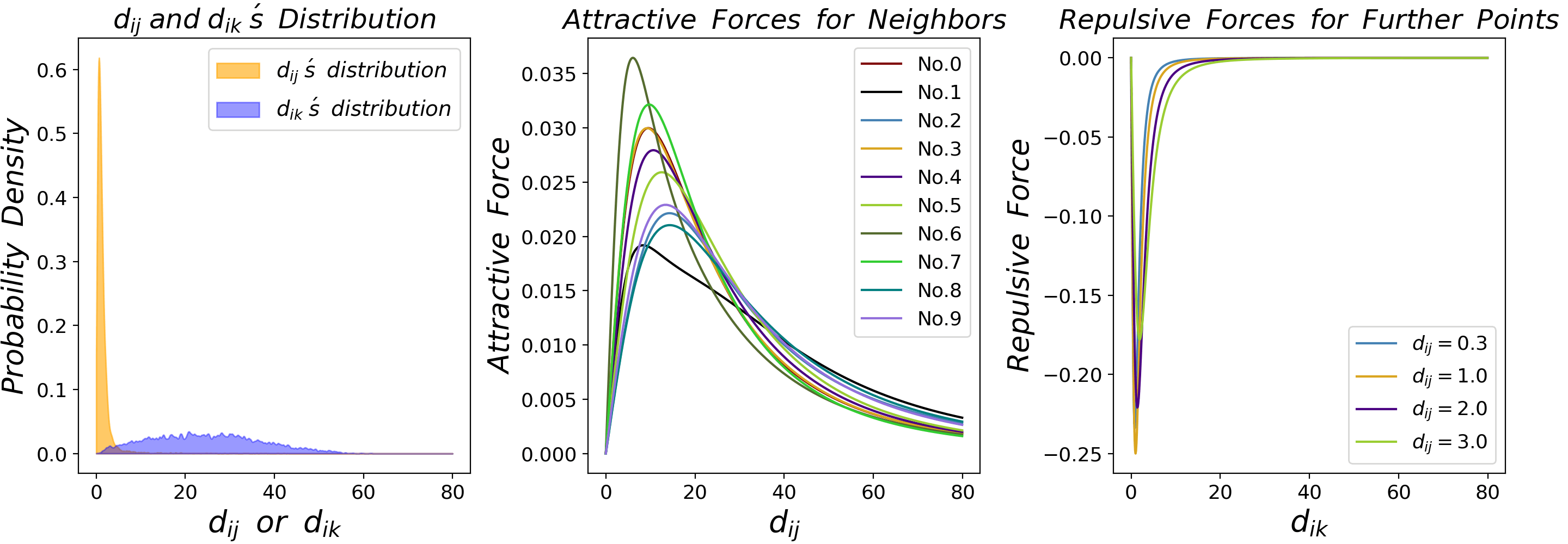}}
\caption{(a) Distribution of $d_{ij}$ and $d_{ik}$ during TriMap training (MNIST dataset; iteration 150 out of 400 total); (b) Attractive force for neighbor $j$ in a triplet; (c) Repulsive force for further point $k$ in a triplet.}
\label{fig:trimap_loss_separation}
\end{center}
\end{figure}

We can determine whether our principles are obeyed empirically by examining the forces on each neighbor $j$ and each further point $k$. 
We considered a moment during training of MNIST, at iteration 150 out of (a default of) 400 iterations.
We first visualize the distribution of $d_{ij}$ and $d_{ik}$, shown in Figure~\ref{fig:trimap_loss_separation}a. From this figure, we observe that if we randomly draw a neighbor $j$ then $d_{ij}$ is small. Conversely, if we randomly draw a further point, $d_{ik}$ has a wide variance, but is not often small.
The implication of Figure~\ref{fig:trimap_loss_separation}a is that most of TriMap's triplets will consist of a neighbor $j$ and a further point (that is far from both $i$ and $j$). This has implications for the forces on each neighbor $j$ and each further point $k$ as we will show in the next two paragraphs.

Let us calculate the forces on $i$'s neighbor $j$ that arise from the triplets associated with it.
For neighbor $j$, TriMap's graph structure includes 5 triplets for it, each with a further point, denoted $(k_1, k_2, k_3, k_4, k_5)$. The contribution to TriMap's loss from the 5 triplets including $j$ is:
\[
Loss_{ij}=\sum_{l=1}^5\frac{\tilde{d}_{ij}}{\tilde{d}_{ij}+\tilde{d}_{ik_l}}=\sum_{l=1}^5\frac{d^2_{ij}+1}{d^2_{ij}+d^2_{ik_l}+2}.
\]
We now consider the loss $Loss_{ij}$ as a function of $d_{ij}$, viewing all other distances as fixed
for this particular moment during training (when we calculate the gradients, we do not update the embedding). 
To estimate these forces during our single chosen iteration of MNIST, for each neighbor $j$, we randomly selected 5 $d_{ik}$ values among $d_{ik}$'s distribution and visualized their resulting attractive forces for $i$'s neighbor $j$. In Figure~\ref{fig:trimap_loss_separation}b, we visualize the results of 10 draws of the $d_{ik}$'s. This figure directly shows that Principles 4 and 6 seem to be obeyed.

Let us move on to the forces on $k$. For each further point $k$ in a triplet, its contribution to TriMap's loss depends on $d_{ij}$ where $j$ is the neighbor in that triplet:
\[
Loss_{ik}=\frac{\tilde{d}_{ij}}{\tilde{d}_{ij}+\tilde{d}_{ik}}=\frac{d^2_{ij}+1}{d^2_{ij}+d^2_{ik}+2}.
\]
Again let us view $d_{ij}$ as a fixed number for this moment during training. Note from Figure~\ref{fig:trimap_loss_separation}a that $d_{ij}$ takes values mainly from $0$ to $3$, which means the set of values $0.3$, $1$, $2$, and $3$ are representative of the typical $d_{ij}$ values we would see in practice. We visualize the repulsive forces received by $k$ corresponding to these 4 typical values of $d_{ij}$, shown in Figure~\ref{fig:trimap_loss_separation}c. From this figure we see directly that violations of Principle 5 are avoided (as discussed above) since there are almost no points along the horizontal axis. 

The other important principles, namely Principles 2 and 3, require consideration of the forces relative to each other, which can be achieved by a balance between attractive and repulsive forces that cannot be shown in Figure \ref{fig:trimap_loss_separation}. 

What we have shown is that TriMap's loss approximately obeys our principles in practice, at least midway through its convergence, after the (very helpful) PCA initialization. However, the principles are necessary for local structure preservation but not sufficient for global structure preservation -- because TriMap 
attracts mainly its neighbors in the high-dimensional space (that is, because the attractive forces in Figure~\ref{fig:trimap_loss_separation}b apply only to neighbors), and because repulsion is again restricted to points that are very close, global structure need not be preserved.

\subsection{PaCMAP's graph structure using mid-near edges and dynamic choice of graph elements} \label{sec:global_structure_and_midnear_edges}

As discussed above, if no forces are exerted on further points, global structure may not be preserved. This suggests that to preserve global structure, an algorithm should generally be able to distinguish between further points at different distances; it should not fail to distinguish between a point that is moderately distant and a point that is really far away. That is, if $j$ is much further than $k$, we should consider repulsing $k$ more than $j$, or attracting $j$ more than $k$.

Let us discuss two defining elements of PaCMAP: its use of \textit{mid-near pairs}, and its \textit{dynamic choice of graph elements} for preservation of global structure.
 \textit{In early iterations, PaCMAP exerts strong attractive forces on mid-near pairs to create global structure}, whereas in later iterations, it resorts to local neighborhood preservation through attraction of neighbors.

\begin{figure}[htb]
\begin{center}
\centerline{\includegraphics[width=0.8\columnwidth]{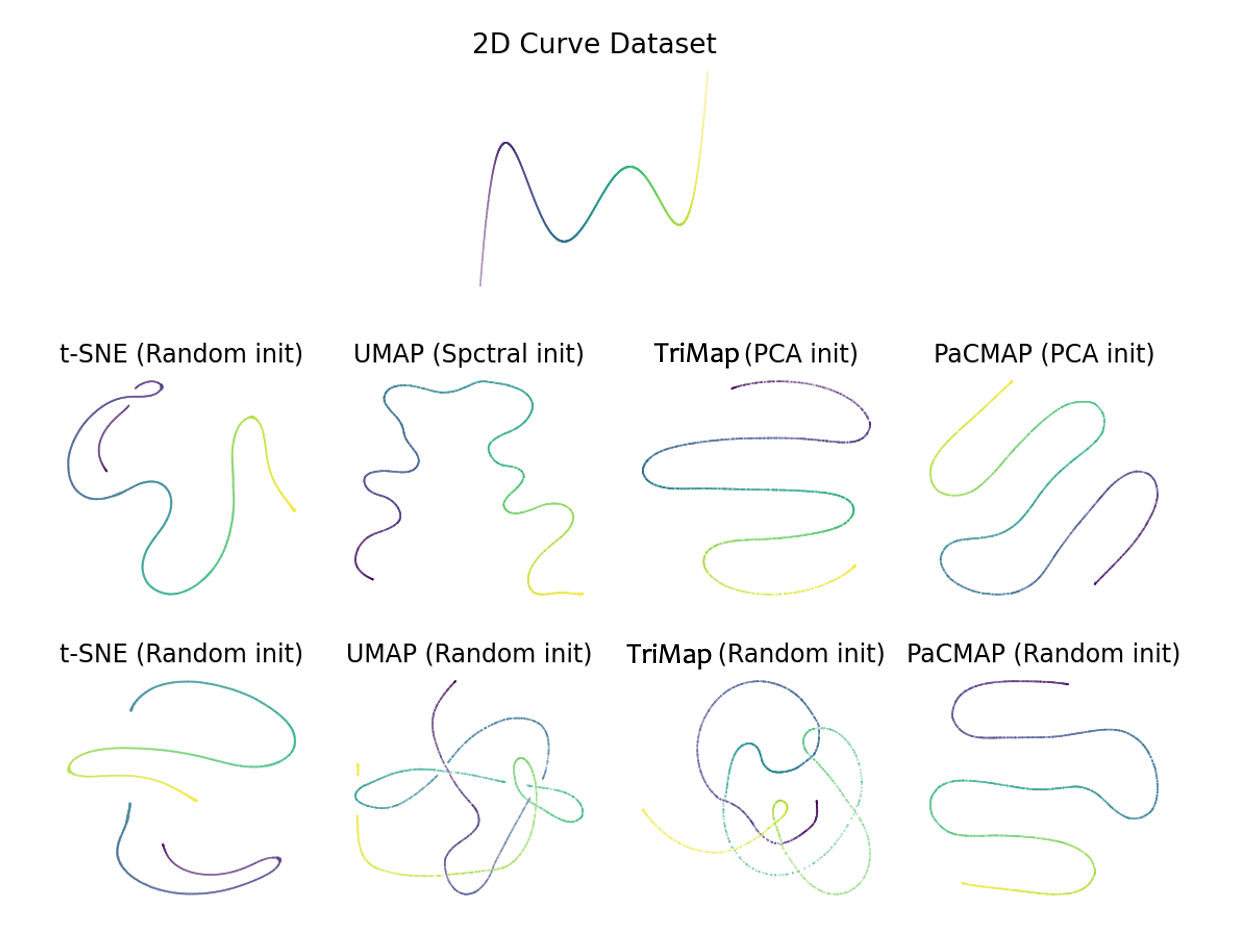}}
\caption{ 
2D line dataset (top), and projection also onto 2D using different initialization procedures and different algorithms. The two left t-SNE figures are both from random initialization; random is the default initialization for t-SNE. Initialization using spectral embedding is the default for UMAP, and PCA initialization is default for TriMap and PaCMAP.
}
\label{fig:2D_curve_line2}
\end{center}
\end{figure}

\begin{figure}[htb]

\begin{center}
\centerline{\includegraphics[width=.45\columnwidth]{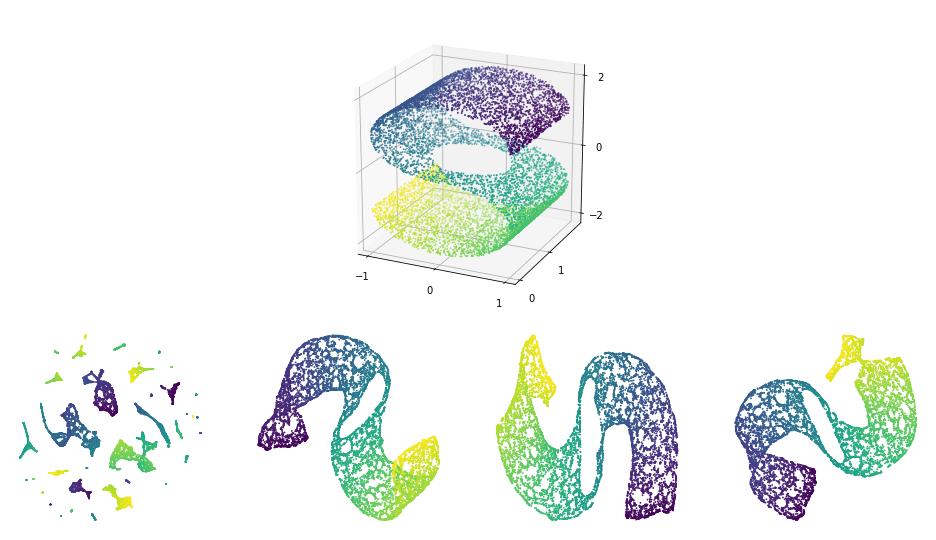}\;\;\;\;\;\;\; \includegraphics[width=.45\columnwidth]{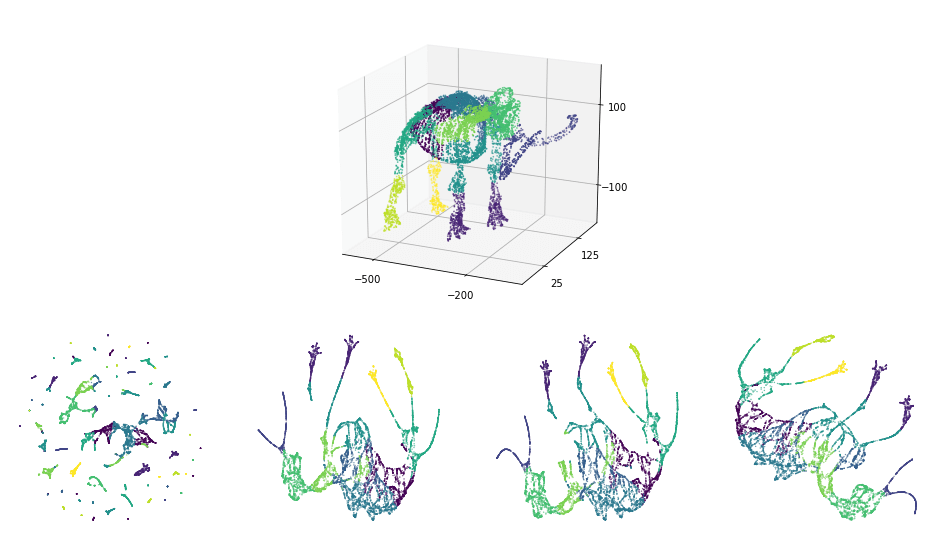}}
\caption{Effect of mid-near pairs on the S-curve with a hole and Mammoth datasets. All figures use PaCMAP. The number of mid-near pairs for the four figures at the bottom are 0, 2, 5, and 8, respectively.}


\label{fig:ablation}
\end{center}
\end{figure}

Figure \ref{fig:2D_curve_line2} illustrates the simple curve dataset. t-SNE and UMAP struggle with this dataset, regardless of how they are initialized. TriMap depends heavily on the initialization. PaCMAP's use of mid-near pairs allow it to robustly recover global aspects of the shape. Figure \ref{fig:ablation} shows this in more detail, where we gradually add more mid-near points to PaCMAP, recovering more of the global structure as we go from zero mid-near points to eight per point.  Let us explain the important elements of PaCMAP in more detail.

\textbf{Mid-near pair construction:} Mid-near points are weakly attracted. To construct a mid-near point for point $i$, PaCMAP randomly samples six other points (uniformly), and chooses the second nearest of these points to be a mid-near point. Here, the random sampling approach allows us to avoid computing a full ranked list of all points, which would be computationally expensive, but sampling still permits an approximate representation of the distribution of pairwise distances that suffices for choosing mid-near pairs.


\textbf{Dynamic choice of graph elements:}
As we will discuss in more depth later, PaCMAP gradually reduces the attractive force on the mid-near pairs. Thus, in early iterations, the algorithm focuses on global structure: both neighbors and mid-near pairs are attracted, and the further points are repulsed. Over time, once the global structure is in place, the attractive force on the mid-near pairs decreases, then stabilizes and eventually disappears, leaving the algorithm to refine details of the local structure.


Recall that the rainbow figure conveys information only about local structure since it considers only neighbors and further points, and not mid-near points. From the rainbow figures, we note that strong attractive forces on neighbors and strong repulsive forces on further points operate in narrow ranges of the distance;
if $i$ and $j$ are far from each other, there is little force placed on them. Because of this, the global structure created in the first stages of the algorithm tends to stay stable as the local structure is tuned later on. In other words, the global structure is constructed early from the relatively strong attraction of mid-near pairs and gentle repulsion of further points; when those forces are relaxed, the algorithm concentrates on local attraction and repulsion.

\subsection{Assigning weights for graph components is not always helpful}

Besides the choice of graph components, some DR algorithms also assign weights for graph components based on the distance information in high-dimensional space. For example, a higher weight for a neighbor pair $(i,j)$ implies that neighbor $j$ is closer to $i$ in the high-dimensional space. The weights can be viewed as a softer choice of graph component selection (the weight is 0 if the component is not selected). 

\begin{figure}[htb]
\begin{center}
\centerline{\includegraphics[width=0.7\columnwidth]{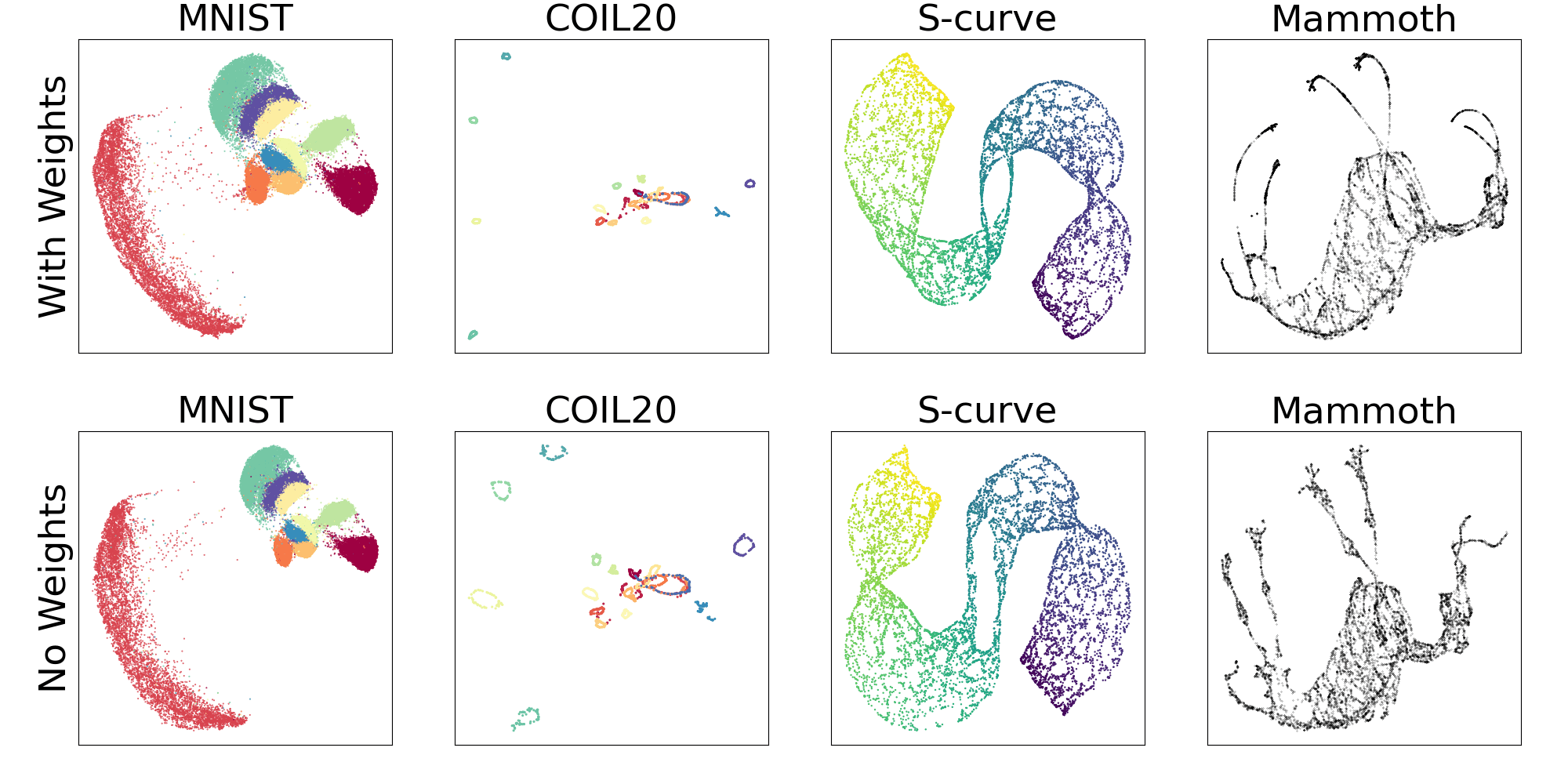}}
\caption{TriMap with and without weights}
\label{fig:trimap_with_without_weights}
\end{center}
\end{figure}

For TriMap, the weight computation is complicated, and yet the weights do not seem to have much of an effect on the final output; other factors (such as initialization and choice of graph components) seem to play a much larger role. 
In Figure~\ref{fig:trimap_with_without_weights}, we showed what happens when we remove the weights (i.e., we assign all graph components to uniform weights) and the resulting performance turns out to be very similar on several datasets. 

In contrast to TriMap, we have elected not to use complicated formulas for weights in PaCMAP. All neighbors, mid-near points, and further points receive the same weights. These three weights change in a simple way over iterations. 

\section{Initialization Can Really Matter}\label{sec:initialization}

Though issues with robustness of some algorithms are well known, in what follows, we present some insight into why they are not robust.


\subsection{Some algorithms are not robust to random initialization}



\begin{figure}[t]
\begin{center}
\centerline{\includegraphics[width=0.8\columnwidth]{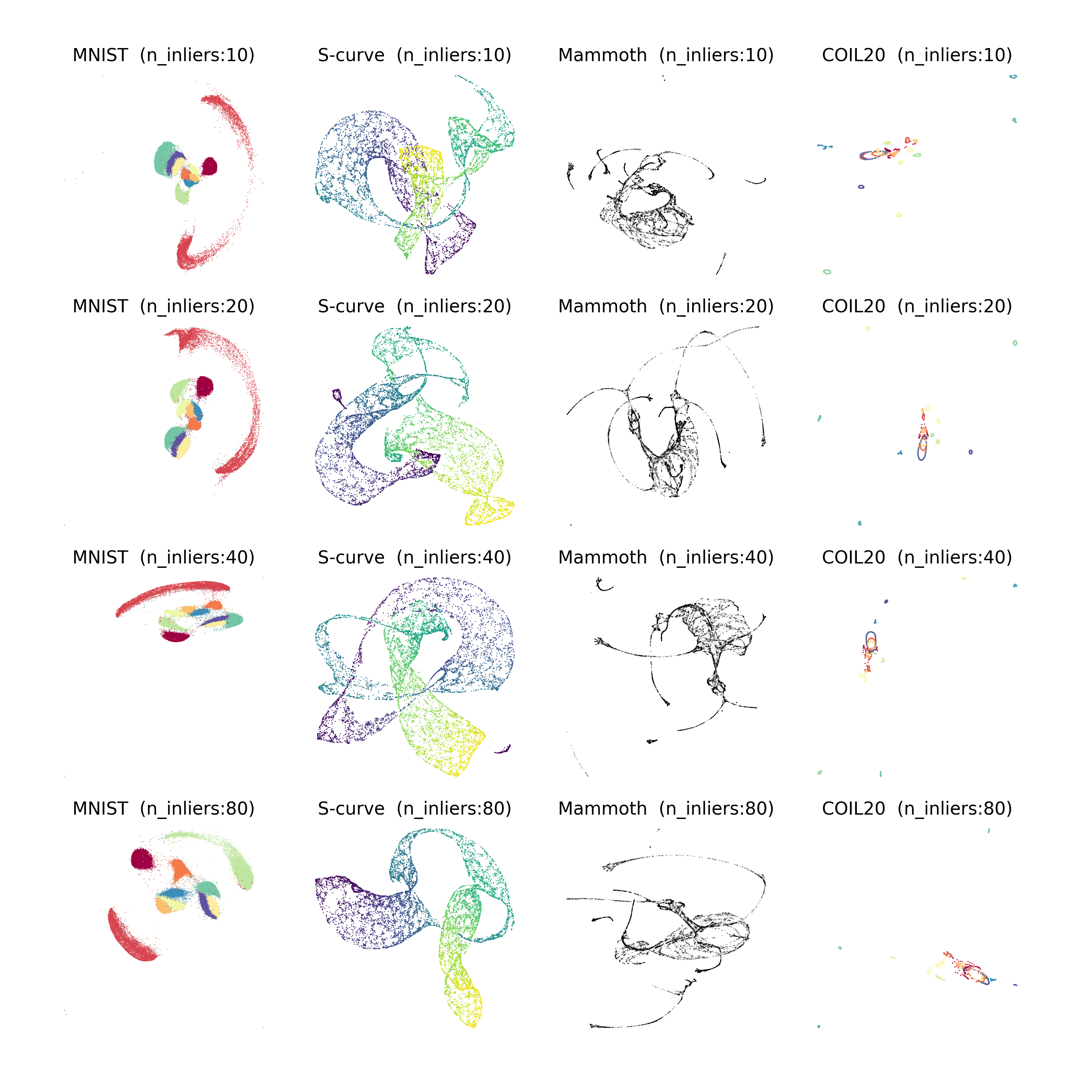}}
\caption{TriMap with random initialization. The ratio among ``n\_inliers,'' ``n\_outliers'' and ``n\_random'' is kept fixed to 2:1:1 (see Section~\ref{sec:trimap} for additional details).}
\label{random_init_trimap}
\end{center}
\end{figure}

\begin{figure}[htb]
\begin{center}
\centerline{\includegraphics[width=0.8\columnwidth]{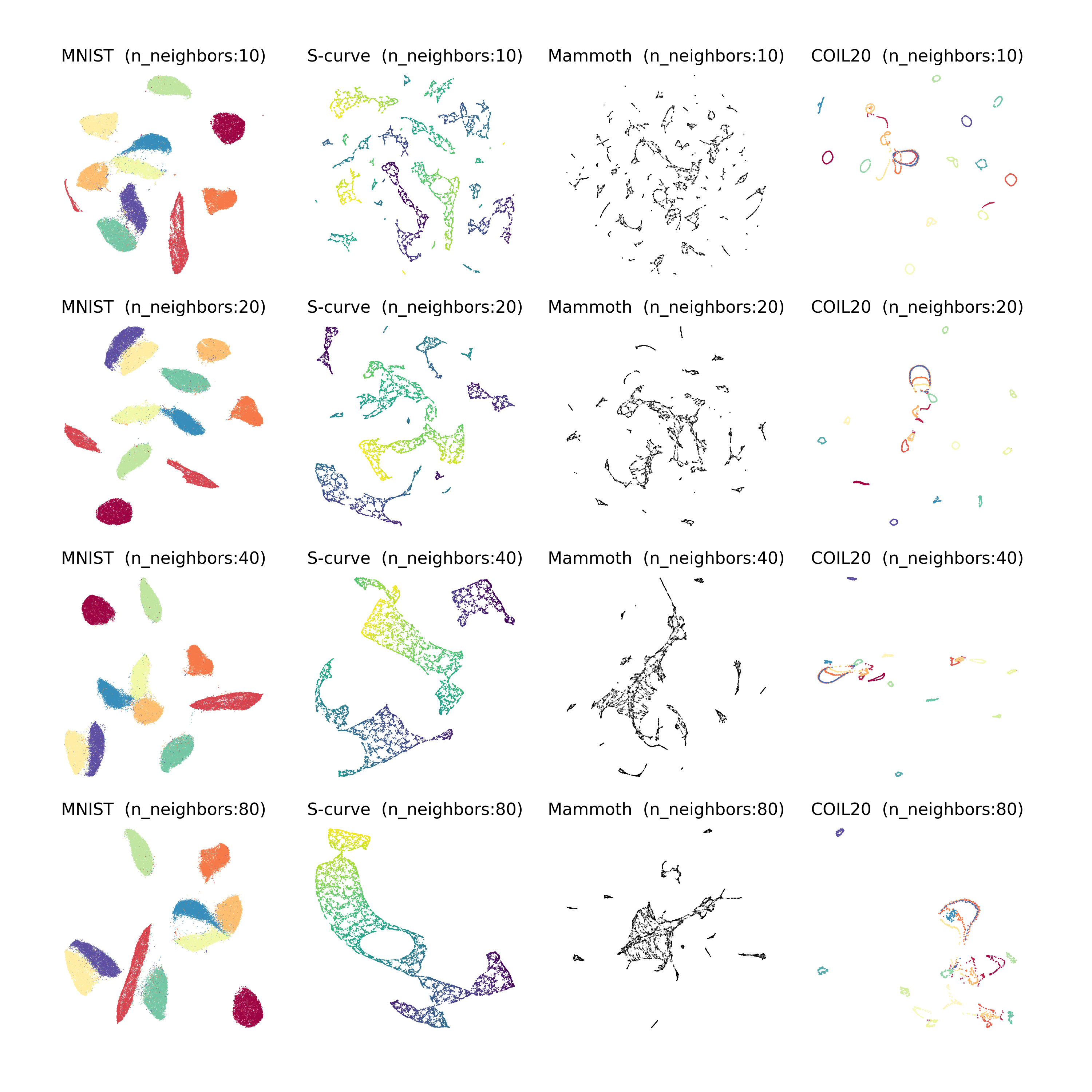}}
\caption{UMAP with random initialization}
\label{random_init_umap}
\end{center}
\end{figure}

The various methods approach initialization differently: t-SNE uses random values, UMAP applies spectral embedding \citep{spectralemb}, and TriMap first runs PCA. According to the original papers \citep{UMAP,TriMap}, both UMAP and TriMap can be initialized randomly, but the specific initializations provided were argued to provide faster convergence and improve stability. However, just as much recent research has discovered \citep{umapglobalstruct, artoftsne}, 
we found that the initialization has a much larger influence on the success rate than simply faster convergence and stability. When we used UMAP and TriMap with random initialization, which is shown in Figures~\ref{random_init_trimap} and \ref{random_init_umap}, the results were substantially worse, even after convergence, and these poor results were robust across runs (that is, robustness was not impacted by initialization, the results were consistently poor). Even after tuning hyper-parameters, we were still unable to observe similar performance with random initialization as for careful initialization. In other words, \textit{the initialization seems to be a key driver of performance for both UMAP and TriMap}. We briefly note that this point is further illustrated in Figures~\ref{fig:mds_init} and \ref{fig:spectral_init} where we apply MDS and spectral embedding.

An important reason why initialization matters so much for those algorithms is the limited ``working zone'' of attractive and repulsive forces induced by the loss, which was discussed in relation to
Figure~\ref{fig:pacmap_loss_force_distri_viz}.
Once a point is pushed outside of its neighbors' attractive forces, it is almost impossible to regain that neighbor, as there is little force (either attractive or repulsive) placed on it. This tends to cause false clusters, since clusters that should be attracted to each other are not. In other words, the fact that these algorithms are not robust to random initialization is a side-effect of the ``near-sightedness'' we discussed at length in the previous sections.

As discussed earlier, PaCMAP is fairly robust (but not completely robust) to the choice of initialization due to its mid-near points and dynamic graph structure.

\subsection{DR algorithms are often sensitive to the scale of initialization}

\begin{figure}[htb]
\begin{center}
\centerline{\includegraphics[width=0.8\columnwidth]{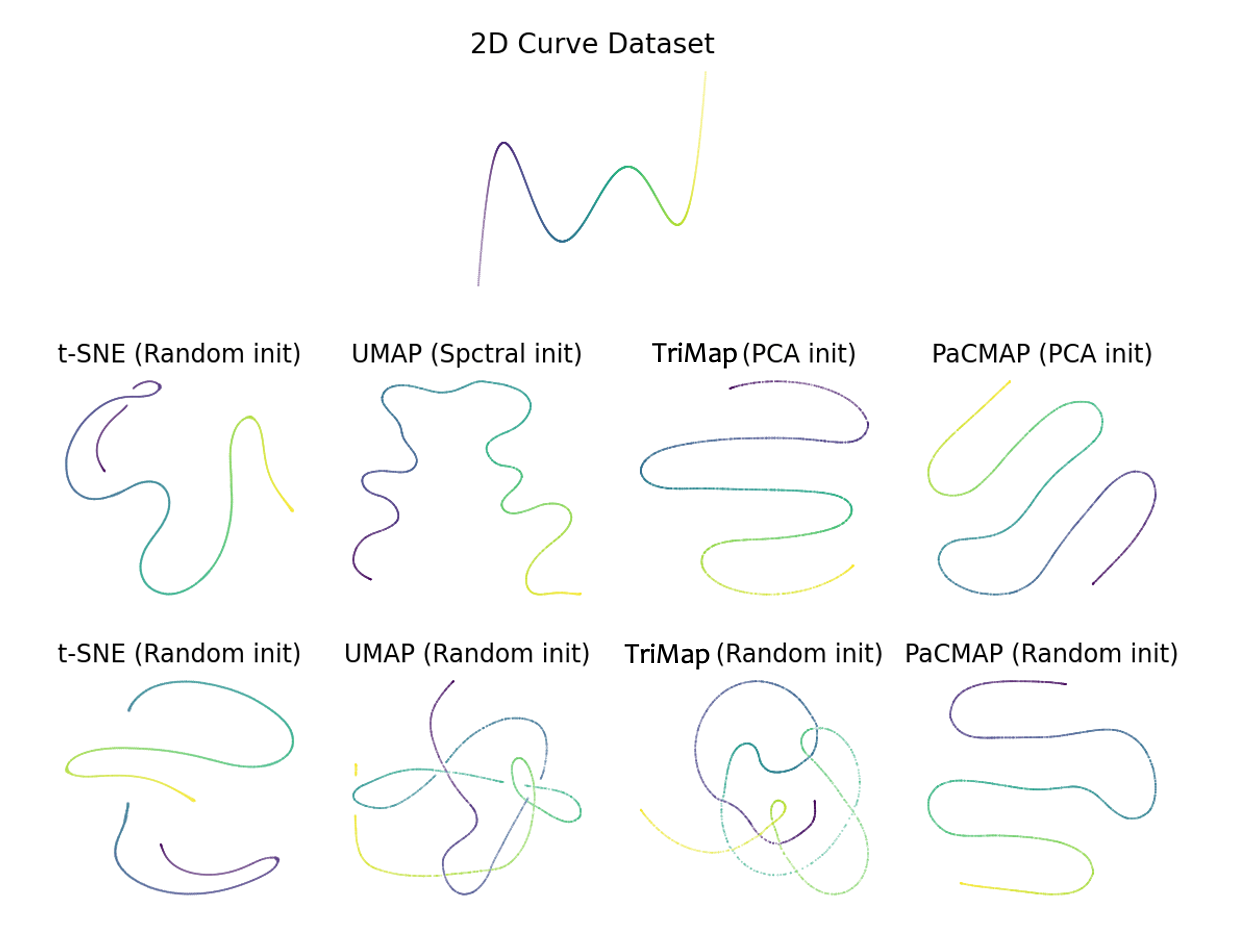}}
\caption{Algorithms t-SNE, UMAP, TriMap and PaCMAP's performance with different scales of initialization.}
\label{fig:initialization_of_different_scales}
\end{center}
\end{figure}

Interestingly, DR algorithms are not scale invariant. Figure \ref{fig:initialization_of_different_scales} shows this for the 2D curve dataset, where both the original and embedded space are 2D. Here we took the distances from the original data, and scaled them by a constant to initialize the algorithms, only to find that even this seemingly-innocuous transformation had a large impact on the result. 

The middle row of Figure \ref{fig:initialization_of_different_scales} is particularly interesting--the dataset is a 2D dataset, so one might presume that using the correct answer as the starting point would give the algorithms an advantage, since the algorithm simply needs to converge at the first iteration to achieve the correct result. However, even in this case, attractive and repulsive forces can exist, disintegrating global structure. 

In the third row, all points start out far from each other because of the scaling, which multiplies all dimensions by 1000. As we know from our earlier analysis, repulsion forces fade as distances fade, explaining why the algorithms all achieved perfect results; however, scaling by 1000 will not work in general, instead it will generally lead the algorithm to stop at the first iteration. The warning here is always to initialize the low-dimensional embedding so that the forces are not all zero.

Hence, because DR methods are not generally scale-invariant, if the scale of the initialization is too large or too small relative to the distance range where attractive and repulsive forces are effective, this could lead to poor DR outcomes.

\section{The PaCMAP Algorithm}\label{sec:pacmap}


We now formally introduce the Pairwise Controlled Manifold Approximation Projection (PaCMAP) method. Algorithm~\ref{alg:pacmap} outlines the implementation of PaCMAP. Its key steps are graph construction, initialization of the solution, and iterative optimization using a custom gradient descent algorithm. In what follows we discuss its finer details and the reasoning behind the design choices.




\begin{algorithm}
    \caption{Implementation of PaCMAP}
    \label{alg:pacmap}
    \begin{algorithmic}

    \REQUIRE
    \qquad\\ 

            $\bullet$ $\mathbf{X}$ - high-dimensional data matrix.
            
            $\bullet$ $\nneighbors$ - the number of neighbor pairs (default values: $\nneighbors=10$).
            
            $\bullet$ $MN\_{\ratio}, FP\_{\ratio}$ - the ratio between the number of mid-near pairs and further pairs with to the number of neighbor pairs (default values: $MN\_{\ratio}=0.5$, $FP\_{\ratio}=2$).
            
            $\bullet$ $n_{\iterations}$ - the number of gradient steps (default value: $n_{\iterations}=450$).
            
            $\bullet$ $init$ - initialization procedure for the lower dim$.$ embedding (default $init =$ PCA, alternatively, $init = random$,  which initializes $\mathbf{Y}$ using the multivariate Normal distribution $\mathcal{N}(0,10^{-4}I)$, with $I$ denoting the two-dimensional identity matrix; alternatively, ).
            
            $\bullet$ $\tau_1,\tau_2,\tau_3$ - beginning of the three optimization phases, satisfying $\tau_1=1 \leq \tau_2 \leq \tau_3 \leq n_{\iterations}$ (default values: $\tau_1=1,\tau_2=101,\tau_3=201$).
            
            $\bullet$ $w_{\neighbors},w_{MN},w_{FP}$ -- the weights associated with neighbor, mid-near, and further pairs at iteration $t$. The default values are: 

            \;\;\; - for $t\in [\tau_1,\tau_2)$: $w_{\neighbors}=2, w_{MN}(t)=1000\cdot\left( 1-\frac{t-1}{\tau_2-1} \right) + 3\cdot \frac{t-1}{\tau_2-1}, w_{FP}=1$;
            
            \;\;\; - for $t\in [\tau_2,\tau_3):$ $w_{\neighbors}=3, w_{MN}=3, w_{FP}=1$;
            
            \;\;\; - for $t\in [\tau_3,n_{\iterations}]:$ $w_{\neighbors}=1, w_{MN}=0, w_{FP}=1$.

    \ENSURE\quad
        
        $\bullet$ $\mathbf{Y}$ - low-dimensional data matrix.
        
    
    
    \FOR{$i=1$ \textbf{to} $N$}
        \STATE{
            $\bullet$ construct $\nneighbors$ neighbor edges by computing the $\nneighbors$ nearest neighbors of $\mathbf{x}_i$ using scaled distances $d^{2,\textrm{select}}_{ij}$.
            To take advantage of existing implementations of k-NN algorithms, for each sample we first select the $\min(n_{NB}+50,N)$ nearest neighbors according to the Euclidean distance and from this subset we pick the $n_{NB}$ nearest neighbors according to the scaled distance $d^{2,\textrm{select}}_{ij}$ (recall that $N$ is the total number of observations).

            
            $\bullet$ construct $n_{MN}=\lfloor \nneighbors \times MN\_{\ratio} \rfloor$ mid-near pairs. For each pair, construct it by sampling 6 observations, using $\mathbf{x}_i$ and the 2nd nearest observation to $\mathbf{x}_i$ as the mid-near pair.
            
            $\bullet$ construct $n_{FP}=\lfloor \nneighbors \times FP\_{\ratio} \rfloor$ further pairs by sampling non-neighbor points.
        }

    \ENDFOR
    \STATE $\bullet$ apply the initialization procedure $init$ to set the initial values of $\mathbf{Y}$.

    \STATE $\bullet$ run AdamOptimizer $n_{\iterations}$ iterations to  optimize the loss function $ \textrm{Loss}^{\textrm{PaCMAP}}$ while simultaneously adjusting  the weights according to the above scheme. 
    \begin{eqnarray*}
    \textrm{Loss}^{\textrm{PaCMAP}}&=&
    w_{\neighbors}\cdot\sum_{i,j \text{ are neighbors}}\frac{\tilde{d}_{ij}}{10 + \tilde{d}_{ij}} + w_{MN}\cdot\sum_{i,k \text{ are mid-near pairs}}\frac{\tilde{d}_{ik}}{10000 + \tilde{d}_{ik}} \\
    &&  + w_{FP}\cdot\sum_{i,l \text{ are further points}}\frac{1}{1 + \tilde{d}_{il}}.    \end{eqnarray*}
    
    
    
    \STATE \textbf{return} $\mathbf{Y}$
    \end{algorithmic}
\end{algorithm}

\paragraph{Graph construction.}
PaCMAP uses edges as graph components. As discussed earlier, PaCMAP distinguishes between three types of edges: neighbor pairs, mid-near pairs, and further pairs. The first group consists of the $\nneighbors$ nearest neighbors from each observation in the high-dimensional space. Similarly to TriMap, the following scaled distance metric is used: 
\begin{equation}\label{eq:d_select}
d^{2,\textrm{select}}_{ij}=\frac{\|\mathbf{x}_{i}-\mathbf{x}_{j}\|^2}{\sigma_{ij}} \text{ and }  \sigma_{ij}=\sigma_{i}\sigma_{j},
\end{equation}
where $\sigma_i$ is the average distance between $i$ and its Euclidean nearest fourth to sixth neighbors. These are used to construct the neighbor pairs $(i, j_t), t=1,2,...,\nneighbors$. The scaling is performed to account for the fact that neighborhoods in different parts of the feature space could be of significantly different magnitudes. Here, the scaled distances $d^{2,\textrm{select}}_{ij}$ are used only for selecting neighbors; they are not used during optimization.

As discussed in Section~\ref{sec:global_structure_and_midnear_edges}, the second group consists of $N\cdot n_{MN}$ mid-near pairs selected by randomly sampling from each observation 6 additional observations and using the second smallest of them for the mid-near pair. Finally, the third group consists of a random selection of $n_{FP}$ further points from each observation. 
For convenience, the number of mid-near and further point pairs is determined by the parameters $MN\_{\ratio}$ and $FP\_{\ratio}$ that specify the ratio of these quantities to the number of nearest neighbors, that is, $n_{MN}=MN\_{\ratio}\cdot \nneighbors$ and $n_{FP}=FP\_{\ratio}\cdot \nneighbors$.
Since the number of neighbors $\nneighbors$ is typically an order of magnitude smaller than the total number of observations, random sampling effectively chooses non-nearest neighbors as mid-near and further pairs. We note that the decision to choose pairs randomly rather than deterministically (e.g., certain fixed quantiles) is aimed at reducing the computational burden.

\paragraph{The loss function.}

As discussed in Section~\ref{sec:apply_principle_to_pacmap_loss}, PaCMAP uses three distinct loss functions for each type of pair:
$$
Loss_{\neighbors} = \frac{\tilde{d}_{ij}}{10 + \tilde{d}_{ij}},\quad
Loss_{MN} = \frac{\tilde{d}_{ik}}{10000 + \tilde{d}_{ik}},\quad
Loss_{FP} = \frac{1}{1 + \tilde{d}_{il}}.
$$
Where $\tilde{d}_{ab}=\|\mathbf{y}_{a}-\mathbf{y}_{b}\|^2 + 1$.
The pairs are further weighted by the coefficients $w_{\neighbors}$, $w_{MN}$, and $w_{FP}$, which altogether account for the total loss. The weights are updated dynamically during the algorithm according to a scheme we describe as part of the optimization process. 
The particular choice of using the transformed distance $\tilde{d}$ is motivated by the Student's t-distribution used in the similarity functions of t-SNE and TriMap. The choice of loss terms could be thought of as related to TriMap's triplet loss, but where the triplet has been decoupled and the third point is fixed. This loss was also motivated by the principles for good loss functions, discussed further in Section~\ref{sec:apply_principle_to_pacmap_loss}.



\paragraph{Initialization of PaCMAP.}
While PaCMAP's outcomes are fairly insensitive to the initialization method, we can still use PCA to improve the running time. See Section~\ref{sec:initialization} for additional details.

\paragraph{Dynamic Optimization.}

The optimization process consists of three phases, which are designed to avoid local optima.
In the first phase (iterations $\tau_1=1$ to $\tau_2-1$), the goal is to improve the initial placement of embedded points to one that preserves, to a certain degree, both the global and local structures, but mainly the global structure. This is achieved by heavily weighing the mid-near pairs. Over the course of the first phase, we gradually decrease the weights on the mid-near pairs, which allows the algorithm to gradually refocus from global structure to local structure.
In the second phase (iterations $\tau_2$ to $\tau_3-1$), the goal is to improve the local structure while maintaining the global structure captured during the first phase by assigning a small (but not zero) weight for mid-near pairs. 

\begin{figure}[ht]
\begin{center}
\centerline{\includegraphics[width=0.9\columnwidth]{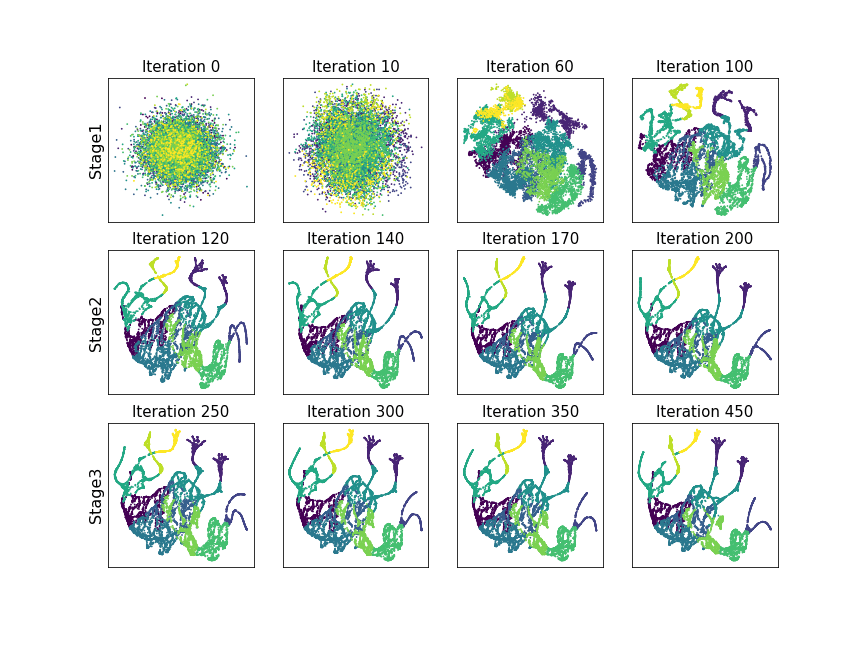}}
\vskip -0.3in
\caption{Visualizations of the low-dimensional embedding over the three-phase optimization process of PaCMAP on the mammoth dataset. Here we can see how the mid-near pairs preserve the global structure of the original embedding effectively, starting from a random embedding.}
\label{fig:change_itr_mammoth}
\end{center}
\end{figure}

\begin{figure}[hbt]
\begin{center}
\centerline{\includegraphics[width=0.9\columnwidth]{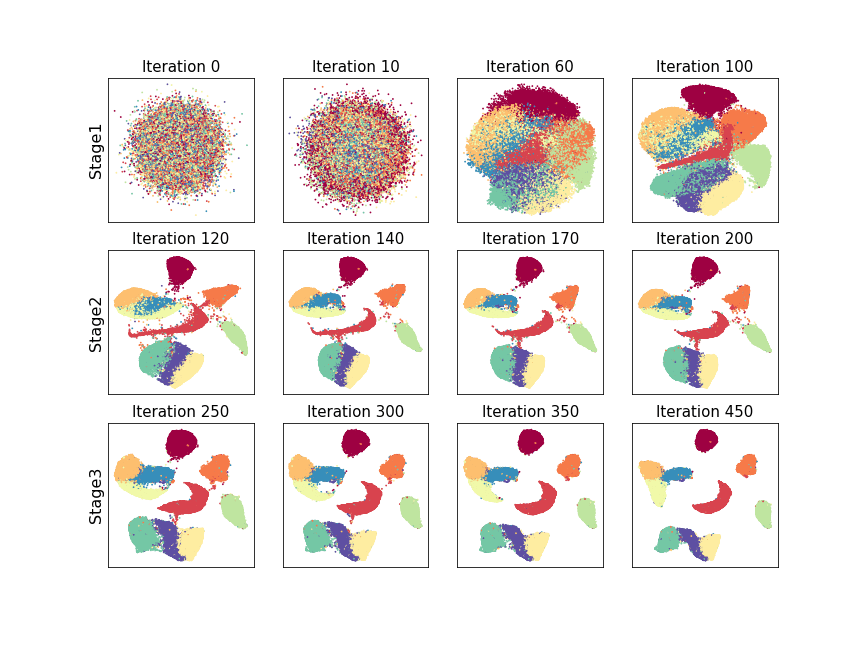}}
\vskip -0.3in
\caption{Visualizations of the low-dimensional embedding over the three-phase optimization process of PaCMAP on the MNIST dataset. This figure demonstrates how phases 2 and 3 refine the local structure of the embedding.}
\label{fig:change_itr_mnist}
\end{center}
\end{figure}

Together, the first two phases try to avoid local optima using a process that bares similarities with simulated annealing and the ``early exaggeration'' technique used by t-SNE. However, early exaggeration places more emphasis on neighbors, rather than mid-near points, whereas PaCMAP 
focuses on mid-near pairs first and neighbors later.
The key hurdles that these phases try to avoid are: first, neglecting to place forces on non-near further points, which ignores global structure; and second, placing neighbors in the low-dimensional space too far away from each other in early iterations, causing the derivatives to saturate, which makes it difficult for these neighbors to become close again. This would lead to false clusters in the low-dimensional embedding. The effect of the three-stage dynamic optimization is demonstrated in Figures~\ref{fig:change_itr_mammoth} and \ref{fig:change_itr_mnist}. In these two figures, PaCMAP is implemented with random initialization rather than PCA initialization which is the default choice. (This figure shows that, although PCA initialization can help PaCMAP achieve more stable global structure, PaCMAP does not rely on PCA initialization to capture global structure.)
With random initialization, these two figures can better demonstrate how PaCMAP actually works.

Finally, in the third phase (iterations $\tau_3$ to $n_{\iterations}$), the focus is on improving the local structure by 
reducing the weight of mid-near pairs to zero and that of neighbors to a smaller value, emphasizing the role of the repulsive force to help separate possible clusters and make their boundary clearer. The third stage seems to have a larger effect on datasets with primarily local structure, like MNIST (Figure~\ref{fig:change_itr_mnist}) as compared with datasets with global structure, such as Mammoth (Figure~\ref{fig:change_itr_mammoth}).

The algorithm uses AdamOptimizer \citep{kingma2014adam}, a modern stochastic gradient descent algorithm. 
Section~\ref{sec:review} reviews the specific algorithms used by the other DR methods.

As we will see in the experiments, PaCMAP has several characteristics worth noting: it tends to favor global structure when such structure exists, and also preserves local structure when such structure exists (the fact that its loss function obeys the principles above helps to ensure this). Its run time results give it a unique advantage, as well as the fact that its results are not nearly as sensitive to the choice of initialization procedure as other methods.

\section{Numerical experiments}\label{sec:numericals}

In this section, we report on the results of a numerical study conducted to assess the quality of PaCMAP on a wide range of datasets, and compare it against leading DR methods. Although PaCMAP is able to reduce a dataset to an arbitrary number of dimensions, we only consider the two dimensional case for visualization purpose in this paper.

Overall, we find that PaCMAP achieves a good balance of local and global structure: for datasets that mainly consist of local structure, it performs as well as algorithms that specifically preserve local structure (e.g., UMAP), while for datasets that consist of mainly global structure, it performs comparably with global structure-preservation algorithms (e.g., TriMap). PaCMAP also seems to be fairly robust in its hyperparameter choices (within non-extreme ranges for those parameters). We find that PaCMAP is significantly faster than other algorithms in runtime experiments.

\subsection{Experimental design}

\paragraph{Algorithms.} We compare PaCMAP against t-SNE, UMAP, TriMap (Section~\ref{sec:review}), as well as LargeVis \citep{Tang16}. 
%
%
DR methods are known to be sensitive to the selection of hyperparameters \citep{wattenberg2016use,UmapMammoth} and there is no single hyperparameter setting would be expected to fit all the datasets. Thus, we selected three different hyperparameters that fall into the suggested tuning ranges for each algorithm. We report the best of these three hyperparameter results on each metric. The specific hyperparameter values we chose for each algorithm are:  
t-SNE, perplexity $\in$ \{10, 20, 40\};
UMAP,  $\nneighbors$ $\in$ \{10, 20, 40\};  
TriMap, $n_{inlier}$ $\in$ \{8, 10, 15\}; 
LargeVis, perplexity $\in$ \{30, 50, 80\}; and
PaCMAP, $\nneighbors \in$ \{5, 10, 20\}.

\paragraph{Datasets.} 
Similarly to other studies of DR methods \citep{Tang16, UMAP, TriMap, Belkina19OptSNE}, we use the following datasets to evaluate and compare DR methods: Olivetti Faces \citep{olivetti}, COIL-20 \citep{coil20}, COIL-100 \citep{coil100}, S-curve with hole dataset (synthesized by the authors), Mammoth \citep{SmithsonianMammoth,UmapMammoth}, USPS \citep{USPS},  MNIST \citep{lecun2010mnist}, FMNIST \citep{fmnist}, Mouse scRNA-seq \citep{scRNAseq}, 20 NewsGroups \citep{20NG}, Flow cytometry \citep{flowcyto}, and KDD Cup 99 \citep{kddcup99}. Note that the S-curve with hole dataset, the Mouse scRNA-seq dataset and the Flow cytometry dataset do not possess class labels. 

\paragraph{Metrics and methodology.} Similarly to other works, we use labeled datasets to assess the quality of the various DR methods, where labels are only used for evaluation once the DR algorithms complete their execution. We consider various types of measures to capture the preservation of local and global structure. 


The quality of the preservation of \textbf{\textit{local structure}} is measured in two ways. First by applying leave-one-out cross validation using the K-Nearest Neighbors (KNN) classifier. This has become a standard evaluation method for DR. The intuition behind this method is that labels tend to be similar in small neighborhoods and therefore the classification accuracy based on neighborhoods would remain close to that in the high-dimensional space. Moreover, one would expect the prediction accuracy of KNN to deteriorate if the DR method does not preserve neighborhoods. 
For each dataset and algorithm, we performed hyper-parameter tuning for the number of neighbors $k$.
We denote this metric as \textbf{\textit{KNN Accuracy}}. 

In addition, we measure the accuracy of nonlinear support vector machine (SVM) models with a radial basis function (RBF) kernel using 5-fold cross validation. Similarly to KNN, the SVM accuracy measures the cohesiveness of neighborhoods, but this is done in a potentially more flexible manner that is less impacted by the density of the data.
Specifically, for each DR method we partition the embedding into 5 folds, each time using 4 folds as the training data for the SVM model and using the remaining fold for the evaluation of accuracy. 
To further reduce the running time, we used the Nystr\"om method, which approximates the kernel matrix by a low rank matrix, using \textrm{sklearn.kernel\_approximation.Nystroem} \citep{scikit-learn}. Thereafter, we trained linear SVM models (using sklearn.svm.LinearSVC) on the non-linearly transformed features. We denote this metric as \textbf{\textit{SVM Accuracy}}.

To measure the preservation of \textbf{\textit{global structure}}, that is the relative positioning of neighborhoods, we sample observations and compute the \textbf{\textit{Random Triplet Accuracy}}, which is the percentage of triplets whose relative distance in the high- and low-dimensional spaces maintain their relative order. For numerical tractability, we use a sample of triplets rather than considering all triplets. Due to randomness, we apply this metric for five times and report the mean value and standard deviation. Note that random triplet accuracy does not require labels, so we can evaluate it on unlabeled datasets.
In the same spirit, but at a lower resolution, for labeled datasets, we also measure global structure preservation by computing the centroids of each class in the high- and low-dimensional spaces, and construct triplets using the relative distances between centroids in the high-dimensional space. We report the percentage of preserved centroid triplets, denoted as \textbf{\textit{Centroid Triplet Accuracy}}.




\paragraph{Computational environment.} 
We implemented PaCMAP in Python by adapting code from TriMap \citep{TriMap} and using the packages ANNOY \citep{Annoy} and Numba \citep{numba}. We used the following implementations for the other DR methods: t-SNE \citep{Belkina19OptSNE}, LargeVis \citep{Tang16}, UMAP \citep{UMAP}, and TriMap \citep{TriMap}. The algorithms were executed on a Dell R730 Server with 2 Intel Xeon E5-2640 v4 2.4GHz CPU. We limited the memory usage to 64 GB and runtime to 24 hours. The code for the numerical experiment is available online at \url{https://github.com/YingfanWang/PaCMAP}. 

\subsection{Main Results}
\label{sec:main_results}
 
\renewcommand{\arraystretch}{0.8}
\begin{figure}[t]
\begin{center}
\centerline{\includegraphics[width=0.85\columnwidth]{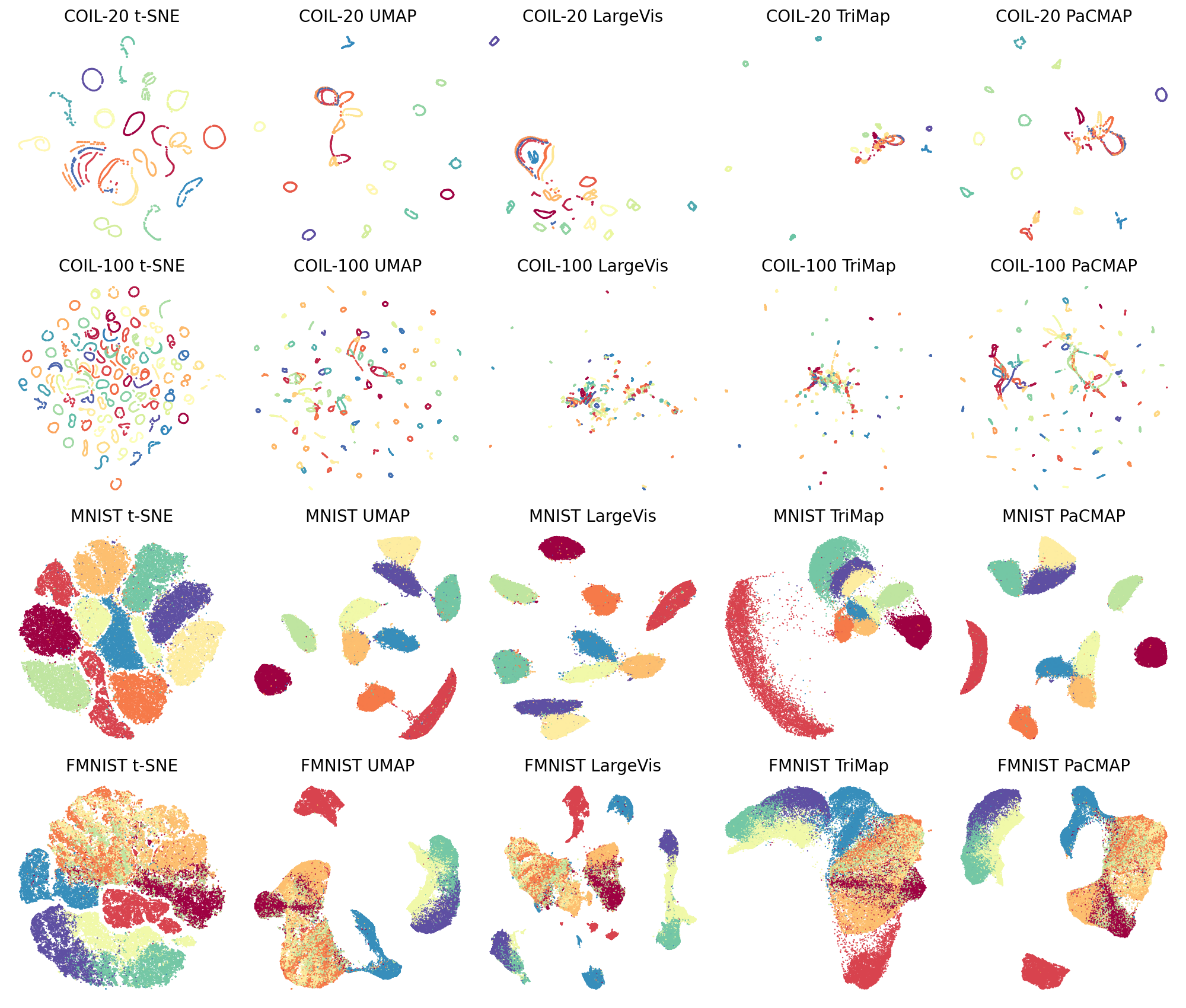}}
\caption{Visualizations of COIL-20, COIL-100, MNIST and FMNIST datasets using t-SNE, UMAP, LargeVis, TriMap, and PaCMAP. Some of the typical behavior of various algorithms is present: t-SNE tends to fill up the space. PaCMAP tends qualitatively to behave like the algorithms that favor local structure (UMAP, LargeVis) on datasets with local structure (MNIST, FMNIST), while behaving like TriMap for datasets with more global structure characteristics (see Figure \ref{fig:Summary_2}). Note on the MNIST t-SNE result, there are two classes that have each been separated into two distinct clusters that do not touch each other; this does not happen with other algorithms on MNIST.}
\label{fig:Summary}
\end{center}
\end{figure}

\begin{figure}[t]
\begin{center}
\centerline{\includegraphics[width=0.85\columnwidth]{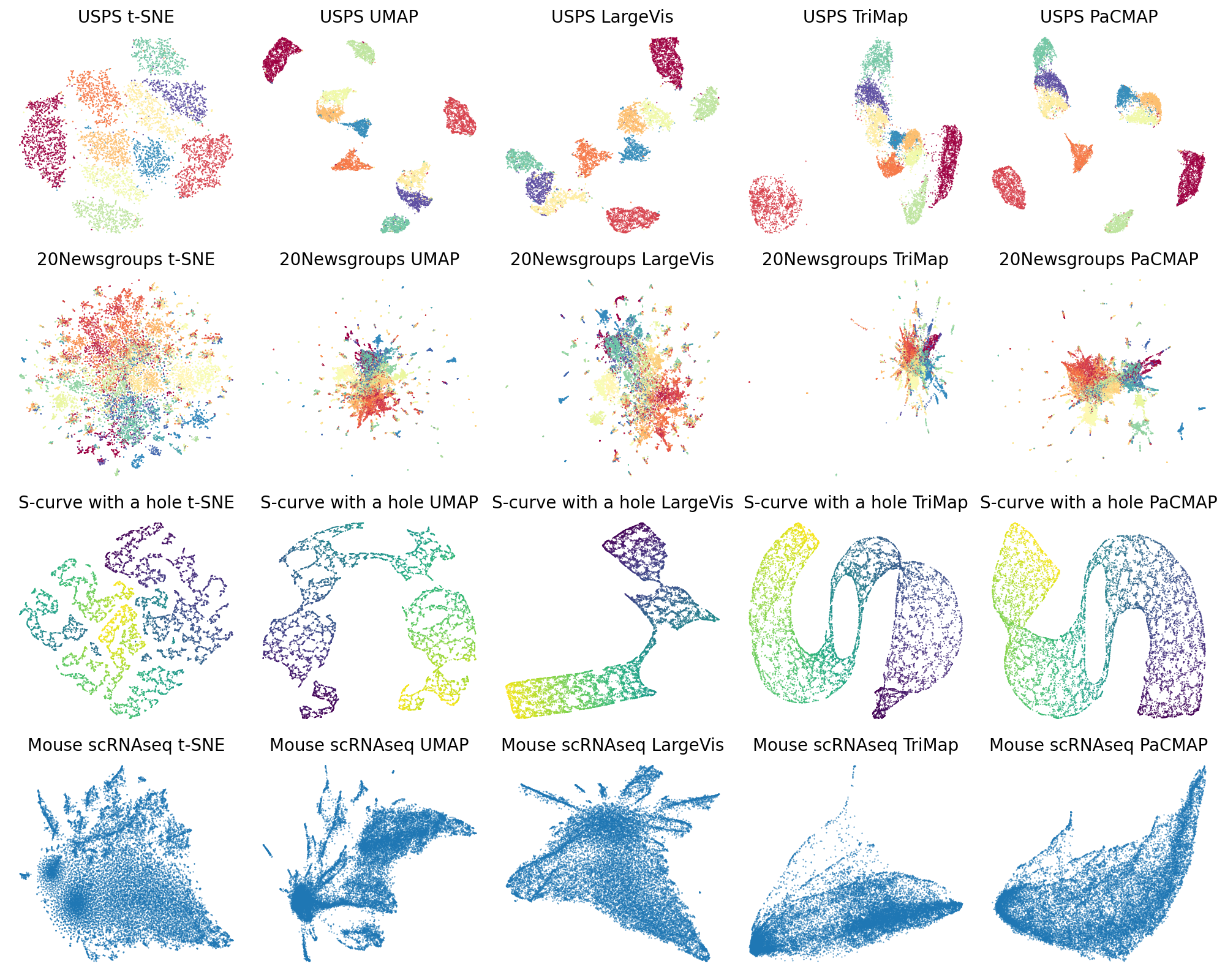}}
\caption{Visualizations of USPS, 20Newsgroups, S-curve with a hole and Mouse scRNAseq datasets using t-SNE, UMAP, LargeVis, TriMap, and PaCMAP. Here PaCMAP and TriMap are able to preserve more of the global structure than other algorithms, particularly for the S-curve data.}
\label{fig:Summary_2}
\end{center}
\end{figure}

\begin{figure}[ht]
\begin{center}
\centerline{\includegraphics[width=.95\columnwidth]{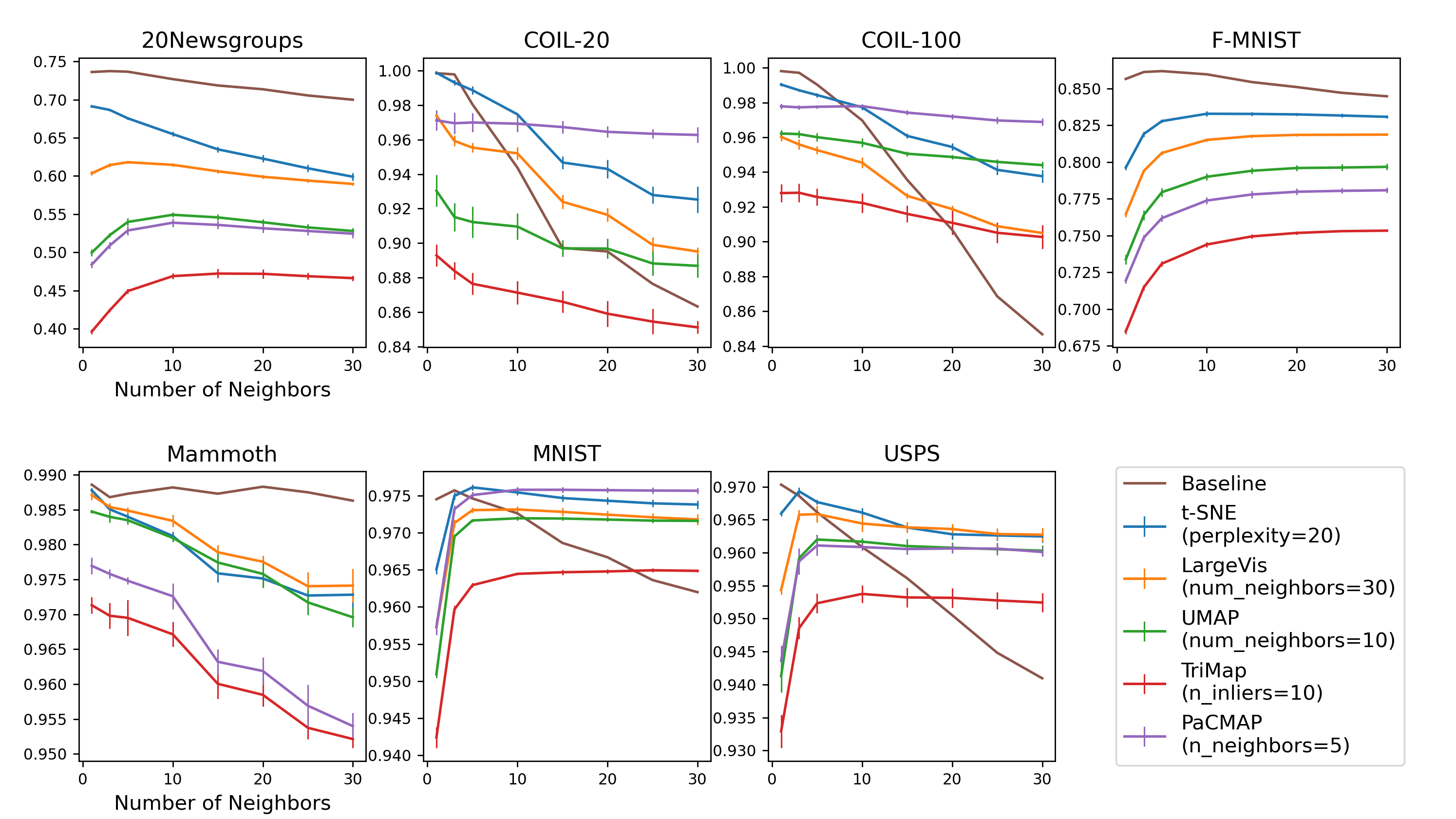}}

\caption{KNN accuracy on different datasets. Algorithms that favor preservation of local structure (t-SNE, UMAP, LargeVis) tend to perform better for KNN accuracy, which measures only local structure preservation. The baseline is calculated in the original high-dimensional space.}
\label{fig:knn_acc}
\end{center}
\end{figure}

\begin{table}[ht]
\caption{SVM Accuracy for datasets that possess class labels. Here local methods such as t-SNE perform well. Some local structure is lost on methods that are able to preserve global structure. For the KDD cup 99 dataset, values are missing because the particular DR method cannot complete the task in the given time or has run out of memory.}
\begin{scriptsize}
\label{table-svmacc}
\begin{center}
\hspace*{-25pt}
\begin{sc}
\begin{tabular}{lcccccc}
\toprule
Dataset (Size) & Baseline & t-SNE & LargeVis & UMAP & TriMap & PaCMAP \\
\midrule
Olivetti Faces (0.4K) & 0.965 & \textit{0.590 $\pm$ 0.039} & 0.048 $\pm$ 0.004 & 0.562 $\pm$ 0.021 & 0.572 $\pm$ 0.014 & \textbf{\textit{0.614 $\pm$ 0.013}}\\

COIL-20 (1.4K) &  0.972 & {0.909 $\pm$ 0.015} & 0.799 $\pm$ 0.020 & 0.844 $\pm$ 0.004 & 0.778 $ \pm $ 0.010 & \textbf{0.942 $\pm$ 0.009} \\

COIL-100 (7.2K) & 0.989 & {0.911 $\pm$ 0.004} & 0.707 $\pm$ 0.014 & 0.879 $\pm$ 0.007 & 0.737 $\pm$ 0.019 & \textbf{0.933 $\pm$ 0.009}\\

USPS (9k) & 0.949 & \textbf{\textit{0.959 $\pm$ 0.002}}  &  \textit{0.957 $\pm$ 0.001} &  \textit{0.956 $\pm$ 0.002} &  0.946 $\pm$ 0.001 & \textit{0.958 $\pm$ 0.001}\\

Mammoth (10K) & 0.961 & 0.927 $\pm$ 0.009 & 0.923 $\pm$ 0.011 & \textit{\textbf{0.941 $\pm$ 0.003}} & 0.900 $\pm$ 0.004 & \textit{0.933 $\pm$ 0.004}\\

20Newsgroups (18K)\hspace*{-15pt} & 0.792 & 0.435 $\pm$ 0.014  &  \textit{0.444 $\pm$ 0.012}  & 0.431 $\pm$ 0.013 & 0.410 $\pm$ 0.007 & \textbf{\textit{0.447 $\pm$ 0.006}}\\

MNIST (70K) & 0.926 & 0.967 $\pm$ 0.002 & 0.965 $\pm$ 0.004 & 0.970 $\pm$ 0.001 & 0.960 $\pm$ 0.001 & \textbf{0.974 $\pm$ 0.001}\\

F-MNIST (70K) & 0.854  & \textbf{\textit{0.754 $\pm$ 0.003}} & \textit{0.748 $\pm$ 0.003}  & 0.742 $\pm$ 0.003 & 0.729 $\pm$ 0.001 & \textit{0.752 $\pm$ 0.004} \\

KDD Cup 99 (4M)  & 0.956 & -- & -- & --  & 0.767 $\pm$ 0.034 & \textbf{0.909 $\pm$ 0.035} \\

\bottomrule
\end{tabular}
\end{sc}
\end{center}
\end{scriptsize}
\end{table}

\begin{table}[t]
\caption{Random triplet accuracy on all datasets. For the Flow cytometry and KDD cup 99 dataset, values are missing because the particular DR method cannot complete the task in the given time or has run out of memory. On this metric, algorithms that favor global structure preservation tend to be more successful.} 
\label{table-rteacc}
\begin{center}
\begin{scriptsize}
\begin{sc}
\hspace*{-25pt}
\begin{tabular}{lccccc}
\toprule
Dataset (Size) & t-SNE & LargeVis & UMAP & TriMap & PaCMAP  \\
\midrule
Olivetti Faces (0.4K) & 0.739 $\pm$ 0.012 & 0.532 $\pm$ 0.013 & \textit{0.754 $\pm$ 0.010} & 0.739 $\pm$ 0.012 & \textbf{\textit{0.761 $\pm$ 0.003}}\\

COIL-20 (1.4K)  & 0.698 $\pm$ 0.016 & \textbf{0.735 $\pm$ 0.011} & 0.649 $\pm$ 0.014 & 0.659 $ \pm $ 0.006 & 0.699 $\pm$ 0.007 \\

COIL-100 (7.2K)   & 0.577 $\pm$ 0.012 & 0.630 $\pm$ 0.021 & 0.568 $\pm$ 0.011 & 0.633 $\pm$ 0.002 & \textbf{0.718 $\pm$ 0.005}\\

USPS (9K) & 0.654 $\pm$ 0.013 & \textit{0.668 $\pm$ 0.011} & \textbf{\textit{0.669 $\pm$ 0.002}} & 0.640 $\pm$ 0.002 & \textit{0.665 $\pm$ 0.002} \\

S-Curve with Hole (9.5K) & 0.722 $\pm$ 0.045 & 0.834 $\pm$ 0.041 & 0.800 $\pm$ 0.013 & 0.838 $\pm$ 0.004 & \textbf{0.866 $\pm$ 0.010} \\

Mammoth (10K) & 0.701 $\pm$ 0.038 & 0.766 $\pm$ 0.024  & 0.816 $\pm$ 0.001 & \textit{\textbf{0.874 $\pm$ 0.001}} & \textit{0.872 $\pm$ 0.003}\\

20Newsgroups (18K) & 0.645 $\pm$ 0.002 & 0.632 $\pm$ 0.001 & 0.664 $\pm$ 0.002 & \textbf{0.704 $\pm$ 0.002} & 0.666 $\pm$ 0.003 \\

Mouse scRNA-seq (20K)   & 0.715 $\pm$ 0.002 & 0.719 $\pm$ 0.003 & \textit{0.727 $\pm$ 0.002} & \textbf{\textit{0.728 $\pm$ 0.001}} & {\textit{0.727 $\pm$ 0.001}}\\

MNIST (70K)  & 0.600 $\pm$ 0.007 & 0.601 $\pm$ 0.007 & 0.614 $\pm$ 0.001 & 0.600 $\pm$ 0.001 & \textbf{0.619 $\pm$ 0.001}\\

F-MNIST (70K)   & 0.679 $\pm$ 0.019 & 0.657 $\pm$ 0.011 & 0.740 $\pm $0.001 & \textbf{0.777 $\pm$ 0.001} & 0.741 $\pm$ 0.002 \\

Flow Cytometry (3M)  & -- & -- & --  & 0.857 $\pm$ 0.001 & \textbf{0.894 $\pm$ 0.005} \\

KDD Cup 99 (4M)  & -- & -- & --  & 0.660 $\pm$ 0.007 & \textbf{0.752 $\pm$ 0.002} \\

\bottomrule
\end{tabular}
\end{sc}
\end{scriptsize}
\end{center}
\end{table}

\begin{table}[t]
\caption{Centroid Triplet Accuracy. This is again a measure of global structure preservation. Missing entries in the table represent running out of time or memory.}
\label{table-cteacc}
\begin{center}
\hspace*{-20pt}
\begin{footnotesize}
\begin{sc}
\begin{tabular}{lccccc}
\toprule
Dataset (Size) & t-SNE & LargeVis & UMAP & TriMap & PaCMAP  \\
\midrule
Olivetti Faces (0.4K) & 0.788 $\pm$ 0.008 & 0.534 $\pm$ 0.006 & \textbf{0.804 $\pm$ 0.002} & 0.786 $\pm$ 0.007 & 0.797 $\pm$ 0.003\\

COIL-20 (1.4K)   & {0.751 $\pm$ 0.017} & \textbf{{0.799 $\pm$ 0.015}} & {0.705 $\pm$ 0.017} & 0.715 $ \pm $ 0.004 & {0.758 $\pm$ 0.012} \\

COIL-100 (7.2K)   & 0.625 $\pm$ 0.014 & 0.680 $\pm$ 0.015 & 0.630 $\pm$ 0.001 & 0.704 $\pm$ 0.001 & \textbf{0.756 $\pm$ 0.005}\\

USPS (9K)  & 0.756 $\pm$ 0.029 & 0.811 $\pm$ 0.025 & 0.814 $\pm$ 0.018 & \textbf{0.874 $\pm$ 0.002} & 0.852 $\pm$ 0.002\\

Mammoth (10K)  & 0.654 $\pm$ 0.067 & 0.714 $\pm$ 0.036 & 0.818 $\pm$ 0.009 & 0.837 $\pm$ 0.003 & \textbf{0.877 $\pm$ 0.002}\\

20Newsgroups (18K)  & \textit{0.787 $\pm$ 0.004} & \textit{0.779 $\pm$ 0.026} & \textit{0.774 $\pm$ 0.016} & \textbf{\textit{0.794 $\pm$ 0.007}} & {0.773 $\pm$ 0.008}\\

MNIST (70K)  & 0.650 $\pm$ 0.026 & 0.668 $\pm$ 0.030 & 0.793 $\pm$ 0.009 & \textbf{0.806 $\pm$ 0.001} & 0.772 $\pm$ 0.008\\

F-MNIST (70K)   & 0.726 $\pm$ 0.052 & 0.749 $\pm$ 0.070 & 0.869 $\pm $ 0.002 & \textbf{0.895 $\pm$ 0.001} & 0.858 $\pm$ 0.002 \\

KDD Cup99 (4M)   & -- & -- & -- & {0.536 $\pm$ 0.010} & \textbf{0.572 $\pm$ 0.004} \\

\bottomrule
\end{tabular}
\end{sc}
\end{footnotesize}
\end{center}
\end{table}

    \paragraph{ \textbf{Qualitative assessment.}}
    Figures~\ref{fig:Summary} and~\ref{fig:Summary_2} show the output of the various DR methods (rows correspond to datasets and columns to DR methods; additional results are presented in Appendix~\ref{sec:Supplemental}).
    We observe that t-SNE tends to distribute data fairly uniformly around the space, which may potentially contribute to the preservation of local structure and hinder the preservation of the global structure. In the particular example of the MNIST dataset, we see that this actually results in creating false clusters (the yellow and red clusters are split into two parts). 
    We also observe that UMAP and LargeVis tend to nicely preserve local structure (observed through cleaner separation of classes), while TriMap tends to focus on the global structure (as seen by its application to the S-curve dataset, and by better preservation of relative distances between clusters in the COIL-20, COIL-100 and MNIST datasets) at the cost of local structure. 
    The figures suggest that PaCMAP balances between local and global, where for datasets that have mainly local structure, it tends to behave like UMAP (preserving local structure), whereas for datasets that have mainly global structure, it tends to behave more like TriMap (preserving global structure).
    
    \paragraph{ \textbf{Preservation of local structure.}}
    Figure~\ref{fig:knn_acc} and Table~\ref{table-svmacc} show the average performance of the KNN and SVM classification accuracies on multiple datasets; these comparisons measure different qualities of the embeddings as discussed above.
    We observe that the baseline nearest neighbor classifier often performs poorly in high dimensions for larger values of $k$ because it ignores the manifold structure on which the data reside. 
    In terms of KNN accuracy, unsurprisingly, TriMap performed the worst in our experiments, while t-SNE arguably attains the best results across datasets, with the other algorithms being competitive with each other with some variability. 
    
    In terms of SVM accuracy, PaCMAP achieved the best performance for more of the datasets than other algorithms, demonstrating its ability to preserve manifold structures on complicated datasets. Again, not favoring local structure, TriMap (and LargeVis, which also favors local structure) performed poorly on this particular experiment.
    
    We also observe that there is a gap between the baseline prediction and the predictions made in lower dimensions. This indicates that the neighborhoods are not perfectly preserved, which could result from the limited capacity of the lower dimension space.
    
    To summarize, \textit{methods that favor preservation of local structure (UMAP, t-SNE) tend to perform better on local-preservation metrics, along with PaCMAP, which also preserves local structure and performs well.}

    \paragraph{ \textbf{Preservation of global structure.}} Tables~\ref{table-rteacc} and \ref{table-cteacc} show the average and standard deviation of the random triplet accuracy and centroid triplet accuracy, respectively. These global structure metrics are where algorithms like PaCMAP and TriMap really shine.
    It is worth noting that unlike UMAP and TriMap, which gain their global structure through initialization (see Section~\ref{sec:NNpair}), PaCMAP creates the global structure completely through its graph component selection and dynamic changes in the component selection, as we have shown in Sections~\ref{sec:global_structure_and_midnear_edges} and~\ref{sec:pacmap}. An additional comparison with random initialization is provided in Section~\ref{sec:sensitivity_analysis}. 
    
    \textit{In short, PaCMAP preserves global structure, without sacrificing local structure or depending on initialization.}
    
      We note that t-SNE, which performed well on the local structure metrics for F-MNIST, has one of the worst performance results with respect to global structure on F-MNIST.
    
    \paragraph{ \textbf{Run time.}} 
    Table~\ref{table-runtime} compares the running time of the different algorithms. \textit{PaCMAP is significantly faster than other algorithms, attaining a speedup of more than 1.5 times faster than other methods for most datasets.} It can also run efficiently on large-scale datasets, such as the Flow Cytometry and KDD Cup99 datasets, whereas multiple DR algorithm failed to converge under the time limit or ran out of memory.
    PaCMAP's speed can be attributed to the design of its loss function which reduces the number of pairs that need to be considered at each iteration, which reduces the usage of computing and memory resources.

\subsection{PaCMAP-specific considerations and sensitivity analysis}
\label{sec:sensitivity_analysis}
 

The most important parameters for controlling PaCMAP behavior are $\nneighbors$, $MN\_{\ratio}$, $FP\_{\ratio}$, and $init$. The meaning of each parameter can be found in Section~\ref{sec:pacmap}. PaCMAP uses the following default parameters: $\nneighbors=10$, $MN\_{\ratio}=0.5$, $FP\_{\ratio}=2$, and $init$ = PCA.
In this subsection, we will perform sensitivity analyses for each of them and assess performance qualitatively. 

\paragraph{Robustness to initialization.} 
Figure~\ref{fig:PaCMAP_10_init} illustrates robustness of performance to initialization choices. \textit{PaCMAP's qualitative performance seems to be consistent regardless of whether random or PCA initialization is used.} In contrast, the same cannot be said of UMAP, particularly for the S-Curve dataset (see Figure~\ref{fig:UMAP_10_init}). MNIST is a dataset that favors local structure, and all algorithms tend to perform well. The S-curve dataset possesses global structure that UMAP reliably does not capture, no matter which method is used for initialization.

We discussed TriMap's heavy dependence on initialization in Section \ref{sec:initialization} and Figure~\ref{random_init_trimap}.
 
\paragraph{Robustness to the number of graph components.}
Figure~\ref{fig:pacmap_robust_nn} demonstrates the robustness of PaCMAP on the Mammoth and MNIST datasets to the number of neighbors (while preserving the ratios of neighbors to mid-near and further point pairs).
We see that increasing the number of graph components in this way can change the structure significantly, as the algorithm focuses less on local structure, and loses detail, e.g., the mammoth's feet.
Increasing the number of neighbors and other graph components makes clusters more compact.

\paragraph{Robustness to ratio of further points.} 
Figure~\ref{fig:pacmap_robust_fp} demonstrates the robustness of PaCMAP on the Mammoth and MNIST datasets to the ratio of further points, while preserving the ratio of mid-near pairs and keeping the number of neighbors fixed at its default value.
We see that changing the ratio of further points also changes the low-dimensional structure, though we cannot draw a strong conclusion on the implications of varying it. Adding more further points potentially improves local structure by pushing neighboring structures apart for MNIST.


\paragraph{Robustness to ratio of mid-near points.} 
Figure~\ref{fig:pacmap_robust_mid} demonstrates the robustness of PaCMAP on the Mammoth and MNIST datasets to the ratio of mid-near points (while preserving the number of neighbors and the number of further pairs).
Figure \ref{fig:ablation} shows that PaCMAP is not robust to the omission of mid-near points; it must use at least a few. The result is fairly robust for a wide range of mid-near values. For extreme values of mid-near pairs, we can see from Figure \ref{fig:pacmap_robust_mid} that the mid-near pairs can overwhelm other attractive and repulsive forces and cause problems with both global and local structure. We would generally not use a large number of mid-near pairs anyway for computational reasons; it becomes computationally expensive to use more than a few, so we suggest the $MN\_{\ratio}$ to be fixed at 0.5, so that the number of mid-near pairs is half the number of neighbors.

\begin{table}[htb]
\caption{Running time comparison (execution was limited to 24 hours). For the flow cytometry and KDD Cup99 dataset, t-SNE cannot complete the task in 24 hours. LargeVis and UMAP ran out of memory.}
\label{table-runtime}
\vskip 0.15in
\begin{center}
\begin{footnotesize}
\begin{sc}
\begin{tabular}{lcccccr}
\toprule
Dataset (Size) & t-SNE & LargeVis & UMAP & TriMap & PaCMAP & Speedup \\
\midrule
Olivetti Faces (0.4K)  & 00:00:04 & 00:08:13 & 00:00:02 & \textbf{00:00:01} & \textbf{00:00:01}\\
COIL-20 (1.4K)    & 00:00:08 & 00:10:18 & 00:00:05 & 00:00:02 & \textbf{00:00:01} & 2.00\\
COIL-100 (7.2K)    & 00:00:49 & 00:09:53 & 00:00:10 & 00:00:06 & \textbf{00:00:03} & 2.00\\
S-Curve with Hole (9.5K) & 00:01:17 & 00:10:09 & 00:00:15 & 00:00:08 & \textbf{00:00:05} & 1.60\\
USPS (9.5K) & 00:01:14 & 00:10:15 & 00:00:15 & 00:00:07 & \textbf{00:00:05} & 1.40\\

Mammoth (10K)    & 00:00:58 & 00:10:36 & 00:00:16 & 00:00:08 & \textbf{00:00:05} & 1.60 \\

20Newsgroups (18K) & 00:03:29 & 00:11:40 & 00:00:19 & 00:00:18 & \textbf{00:00:12} & 1.50\\

Mouse scRNA-seq (20K)    & 00:04:43 & 00:12:52 & 00:00:24 & 00:00:20 & \textbf{00:00:13} & 1.54\\
MNIST (70K)  & 00:14:02 & 00:20:19 & 00:01:09 & 00:01:14 & \textbf{00:00:52} & 1.33\\
F-MNIST (70K)   & 00:12:43 & 00:17:11 & 00:00:59 & 00:01:13 & \textbf{00:00:47} & 1.26\\
Flow Cytometry (3M)  & - & - & -  & 02:10:27 & \textbf{00:58:28} & 2.23\\
KDD Cup99 (4M)  & - & - & -  & 03:34:57 & \textbf{02:05:19} & 1.72\\

\bottomrule
\end{tabular}
\end{sc}
\end{footnotesize}
\end{center}
\end{table}


\section{Discussion and Conclusion}
\label{sec:discussion}

In this study, our goal was to \textit{empirically} dissect 
what approaches to dimension reduction work and what do not, using a selection of datasets for which local- and global-structure preservation needs are different, and where the algorithms could be visually evaluated after projecting to two dimensions.

The observations from our studies turned out to yield valuable and unexpected results, particularly in showing what is \textit{not} necessarily important for DR: while some of the loss functions and forces are motivated through probabilistic modeling (e.g., t-SNE) or mathematical constructs such as simplices (UMAP), these extra layers of statistics and mathematics may obscure a simple explanation of the how these methods are similar or different from each other, or what is important in general. We showed that the derivations for these other algorithms are modeling choices, rather than choices inherent to the problem: there is no inherent reason why neighbors should be probabilistically related to each other, nor that an understanding of simplices are necessary for deriving DR algorithms; TriMap's and PaCMAP’s purely loss-based approaches are also equally feasible modeling choices. 

By seeing the results of these choices empirically, we showed that several seemingly important aspects of several DR algorithms \textit{do not} drive performance, and can actually hurt it. In particular, we showed several reasons that t-SNE and UMAP's weighting choices and graph component selections do not preserve global structure. In particular, they leave out forces on further points, making these algorithms ``near-sighted'' in preserving only local structure. Even TriMap has this problem; we showed that a key driver of TriMap's excellent performance on global structure is a seemingly-innocent initialization choice.

In striving to compare algorithms, we found that the rainbow figures and their accompanying force plots provide insight into what principles a good loss function for DR should possess; straying from these principles destroys the DR result. These principles, in turn, led to the development of a simple loss, comprised of three simple fractional losses that, when combined with other insights, yield the powerful and robust PaCMAP algorithm. 

PaCMAP has several carefully-chosen elements that allow it to uniquely preserve both local and global structure:
\begin{itemize}
    \item Its choice of graph components ensures that further points have non-zero forces, which remedies a fault we found in other algorithms. This observation is a key to preservation of global structure that is not found in other methods.
    \item Its dynamic graph component selection focuses on global structure in early iterations, fine-tuning local structure at later iterations, allowing the result to preserve both local and global structure. It behaves like UMAP for datasets that mainly have local structure characteristics, and behaves like TriMap for datasets whose important structure is global. Unlike TriMap, it does not rely on initialization to preserve global structure; its dynamic graph component selection ensures that it performs well even under random initialization.
    \item Its loss function does not require triplets and does not use too many unnecessary graph components, leading to faster computation and convergence times. 
    \item Its performance is relatively robust within reasonable parameter choices.
\end{itemize}

The qualitative performance results we examine for PaCMAP led directly to high-quality quantitative results measured on 12 datasets.

There are many avenues for future work.
First, one could consider the possibility of a continuum between local and global structure. While considering a simple dichotomy between local and global has been convenient, it is possible that the data contains multi-scale or hierarchical structure at many levels. We provide an analysis of a simple synthetic dataset with hierarchical structure in Appendix~\ref{sec:hierarchical}, where all methods have difficulty displaying structure on all levels of the hierarchy (although PaCMAP slightly outperforms other methods). A second direction for future work is to study very large scale datasets, particularly in high dimensions. Large datasets pose problems for DR methods, both in scaling DR methods to handle data of such sizes, but also, parameters that have been tuned on smaller datasets often do not extend to good performance on larger datasets. Determining what adjustments need to be made for larger datasets is an important direction for future work. Third, there are open questions about the combination of distance metric learning in high dimensions and how it might couple with DR methods. If we do not know the distance metric perfectly, learning it so that DR behaves well might be useful; on the other hand, combining distance metric learning with DR  risks luring the method towards distance metrics that may be unfaithful to the data just so that DR results look convincing. Fourth, the development of evaluation metrics is important for DR. We have introduced several evaluation metrics that we have found useful. It is possible that other evaluation metrics might also be useful, and it could be possible to develop specialized metrics for various application domains. Finally, one could use the general principles established in this work to design new DR methods. These new algorithms could have very different functional forms than PaCMAP while still obeying the principles we have established. The principles  open the door to a wide variety of other algorithms and new ways to design them.




\acks{This project developed from the Statistics and Applied Mathematical Sciences Institute (SAMSI) working group on interpretable deep neural networks. We would like to acknowledge funding from SAMSI as well as the Duke Department of Computer Science. Finally, we would like to thank the review team for the efficient handling of the paper and the insightful comments.}


\appendix
\section{Supplemental Figures}
\label{sec:Supplemental}

\setcounter{figure}{0}    
\renewcommand\thefigure{\thesection.\arabic{figure}}

\begin{figure}[hbt!]
\begin{center}
\centerline{\includegraphics[width=1.0\columnwidth]{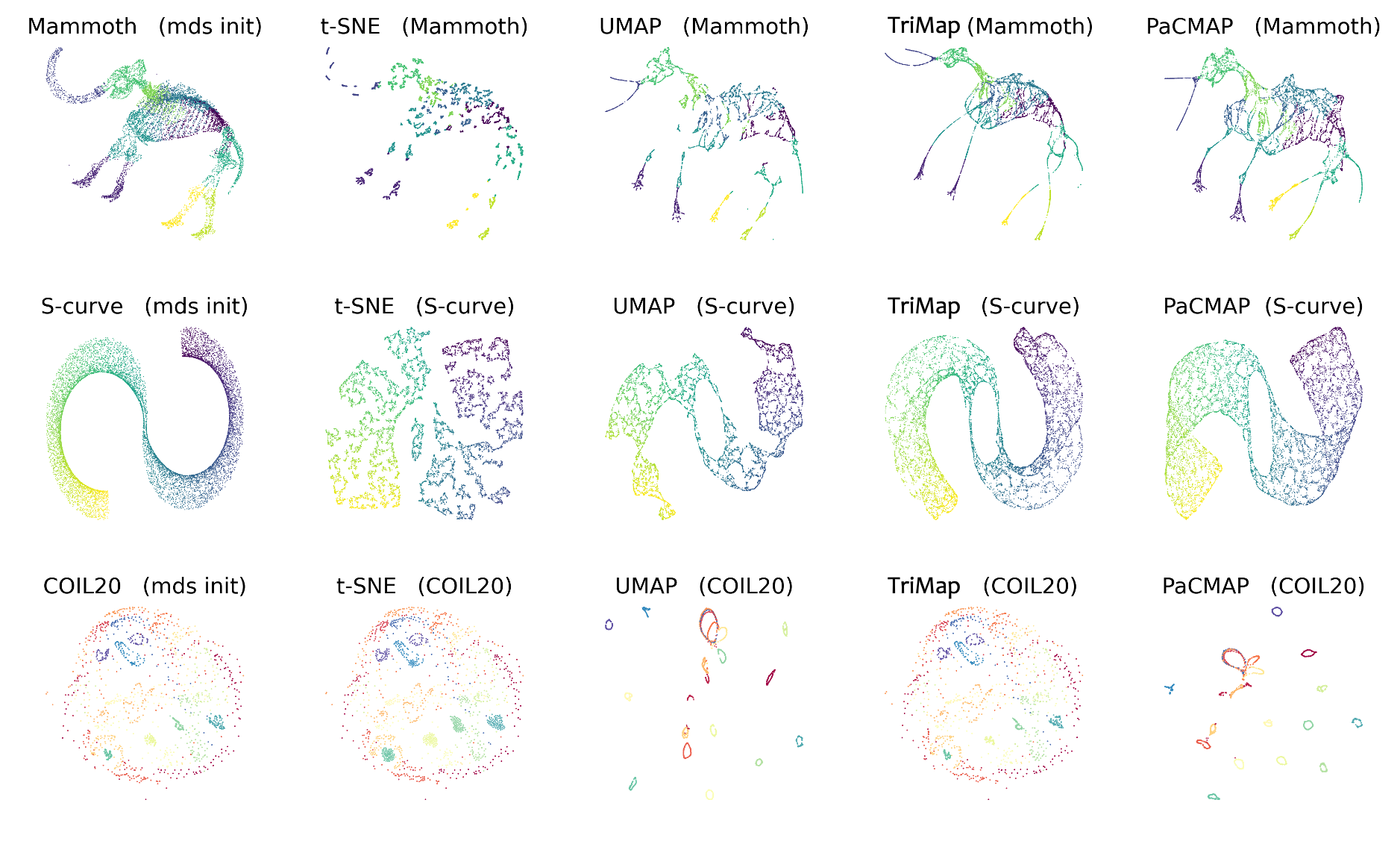}}
\caption{DR algorithms' results with MDS initialization.}
\label{fig:mds_init}
\end{center}
\end{figure}

\begin{figure}[hbt!]
\begin{center}
\centerline{\includegraphics[width=1.0\columnwidth]{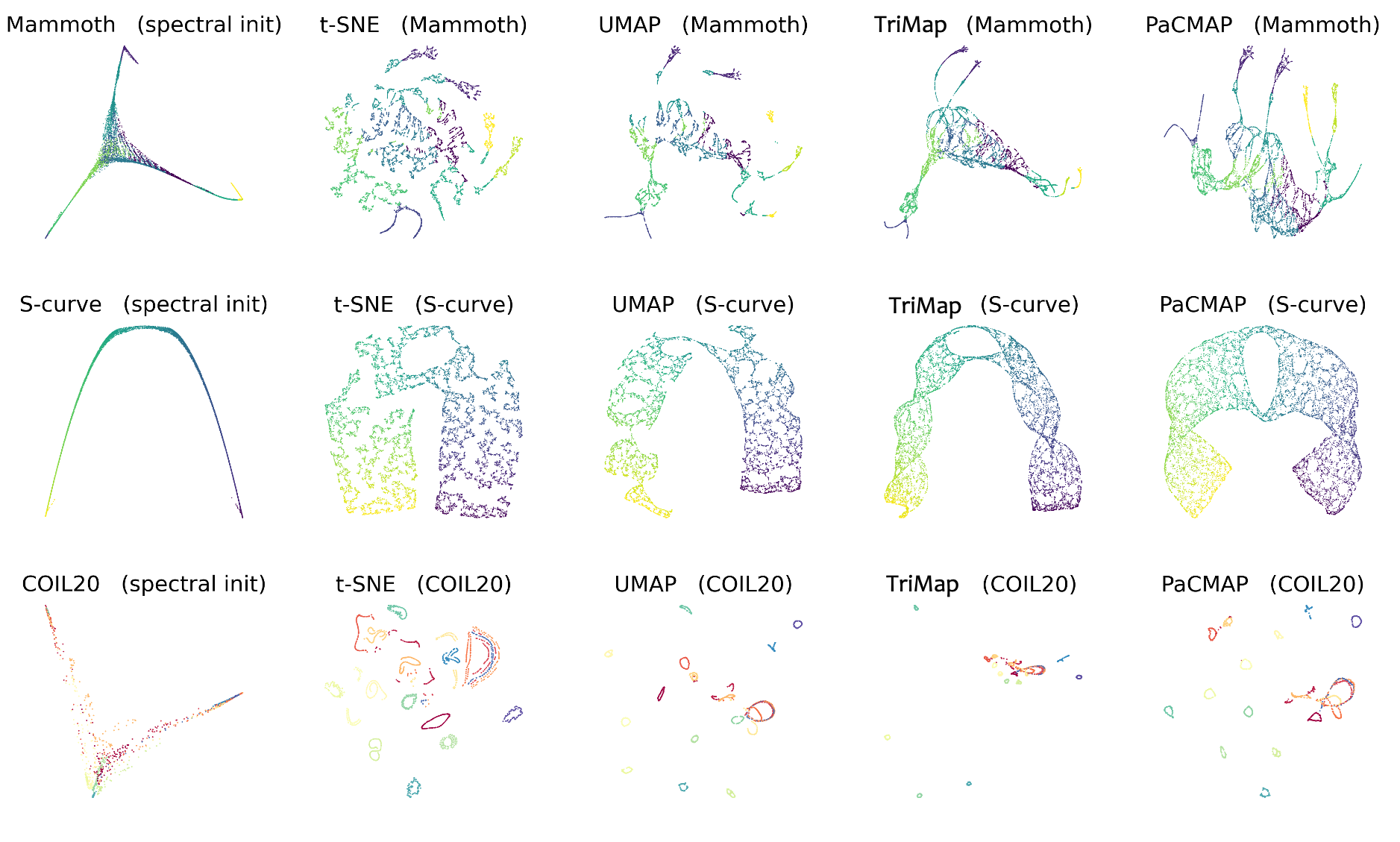} }
\caption{DR algorithms' results with spectral embedding initialization.}
\label{fig:spectral_init}
\end{center}
\end{figure}

Figures \ref{fig:mds_init} and \ref{fig:spectral_init} show the output of the various DR algorithms with MDS and spectral embedding initialization, respectively.

\begin{figure}[hbt!]
\begin{center}
\centerline{\includegraphics[width=1.3\columnwidth]{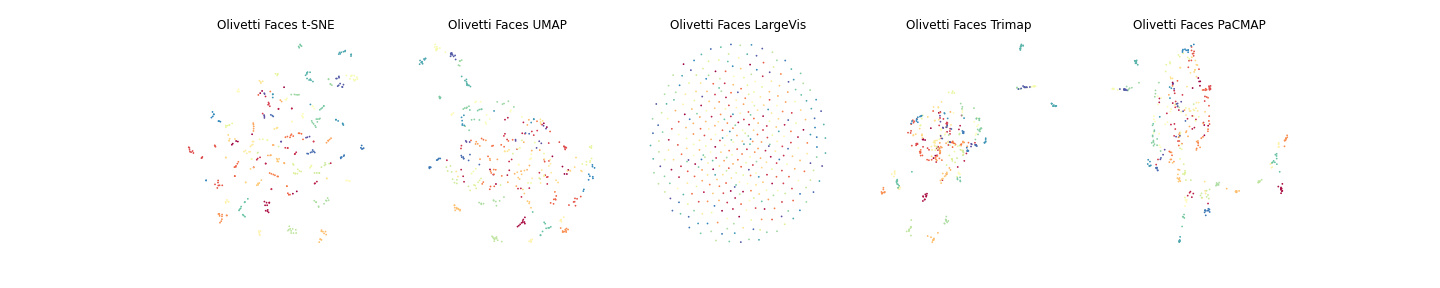} }
\caption{Visualization of the Olivetti Faces dataset using t-SNE, UMAP, LargeVis, TriMap and PaCMAP.}
\label{fig:olivetti_faces}
\end{center}
\end{figure}

\begin{figure}[hbt]
\begin{center}
\centerline{\includegraphics[width=0.5\columnwidth]{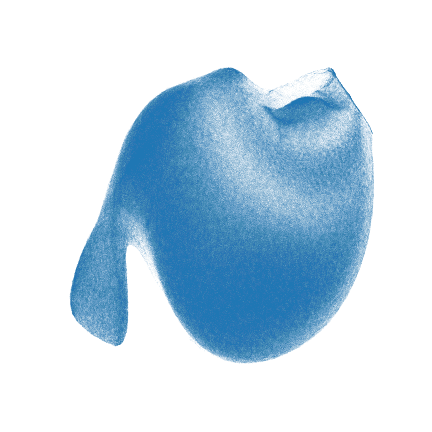}}
\caption{Visualization of the Flow Cytometry dataset by PaCMAP}
\label{fig:pacmap_flow}
\end{center}
\end{figure}

\begin{figure}[hbt]
\begin{center}
\centerline{\includegraphics[width=0.5\columnwidth]{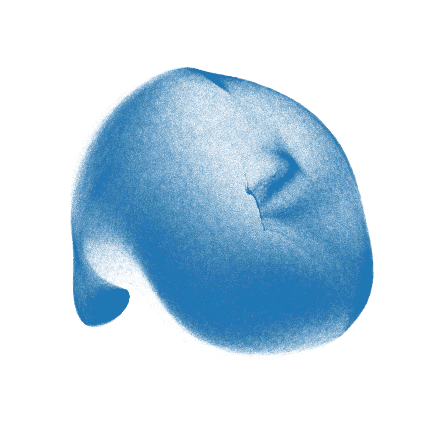}}
\caption{Visualization of the Flow Cytometry dataset by TriMap}
\label{fig:trimap_flow}
\end{center}
\end{figure}

\begin{figure}[hbt]
\begin{center}
\centerline{\includegraphics[width=0.4\columnwidth]{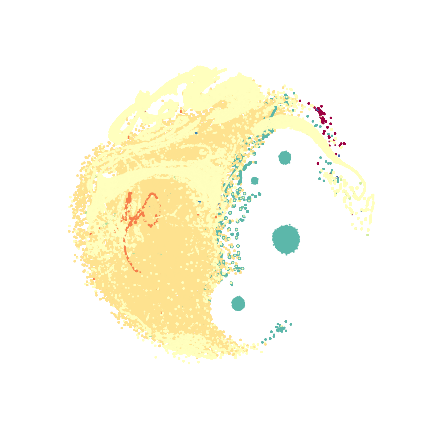}}
\caption{Visualization of the KDD Cup99 dataset by PaCMAP}
\label{fig:pacmap_kdd}
\end{center}
\end{figure}

\begin{figure}[hbt]
\begin{center}
\subfloat[KDD Cup99 Visualization of TriMap]{\includegraphics[width = 0.4\columnwidth]{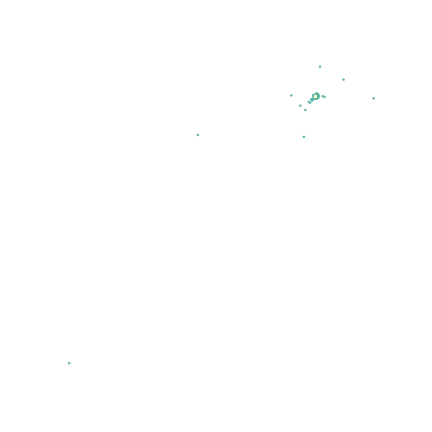}} 
\label{subfig:trimapkdd_a}
\subfloat[Zoomed-in version of the top right corner ]{\includegraphics[width = 0.4\columnwidth]{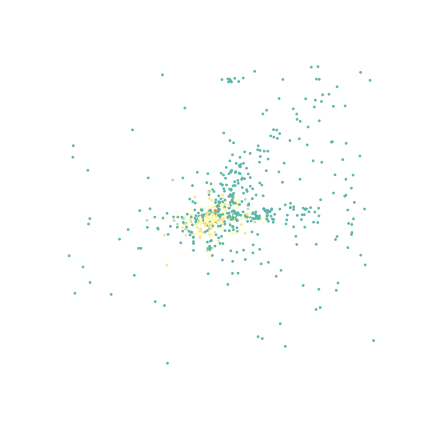}} 
\label{subfig:trimapkdd_b}
\caption{Visualization of the KDD Cup99 dataset by TriMap. Due to TriMap's tendency to preserve global structure, most of the points are projected together, whereas a few outliers are separated from them. We zoom into the top right corner (right) to see the outliers.}
\label{fig:trimap_kdd}
\end{center}
\end{figure}

Figures \ref{fig:olivetti_faces}, \ref{fig:pacmap_flow}, \ref{fig:trimap_flow}, \ref{fig:pacmap_kdd}, and \ref{fig:trimap_kdd} include additional outcomes of DR discussed in Section~\ref{sec:main_results}.

\begin{figure}[t]
\begin{center}
\begin{tabular}{c}
\subfloat[PaCMAP with random initialization on MNIST dataset]{\includegraphics[width = 1\columnwidth]{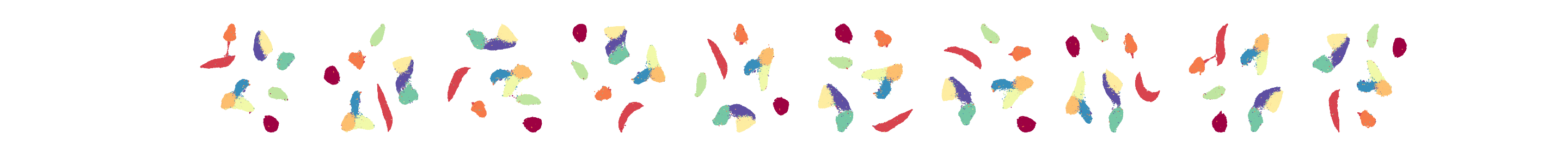}} 
\label{subfig:M_PMAP_R}\\
\subfloat[PaCMAP with PCA initialization on MNIST dataset]{\includegraphics[width = 1\columnwidth]{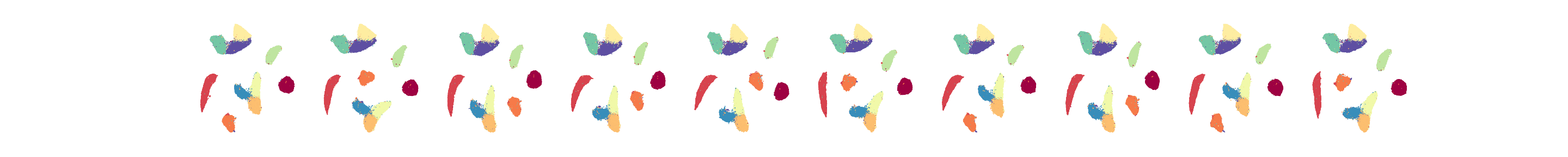}} 
\label{subfig:M_PMAP_I}\\
\subfloat[PaCMAP with random initialization on S curve with a hole dataset]{\includegraphics[width = 1\columnwidth]{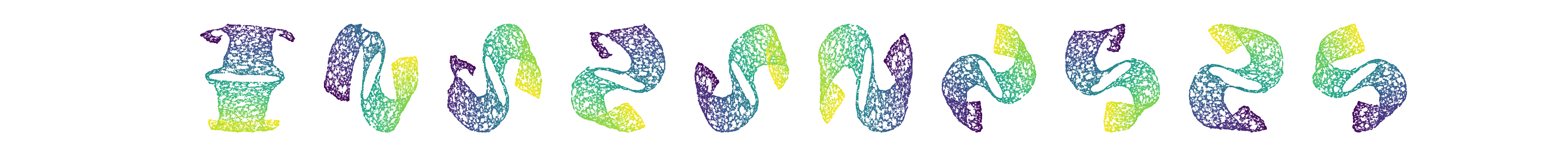}}
\label{subfig:S_PMAP_R}\\
\subfloat[PaCMAP with PCA initialization on S curve with a hole dataset]{\includegraphics[width = 1\columnwidth]{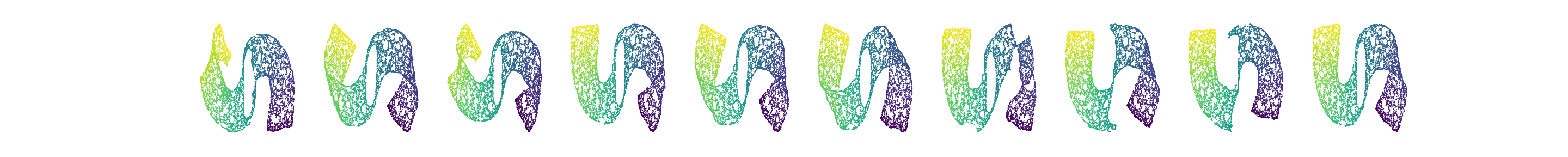}}
\label{subfig:S_PMAP_I}\\

\end{tabular}
\caption{PaCMAP's performance with different initialization choices. We ran PaCMAP ten times for each choice to show performance consistency.} 
\label{fig:PaCMAP_10_init}

\end{center}
\end{figure}

\begin{figure}[t]
\begin{center}
\begin{tabular}{c}
\subfloat[UMAP with random initialization on MNIST dataset]{\includegraphics[width = 1\columnwidth]{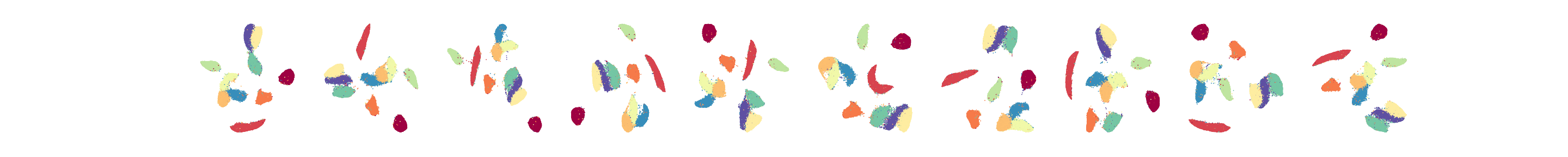}} 
\label{subfig:M_UMAP_R}\\
\subfloat[UMAP with Spectral Embedding (default) initialization on MNIST dataset]{\includegraphics[width = 1\columnwidth]{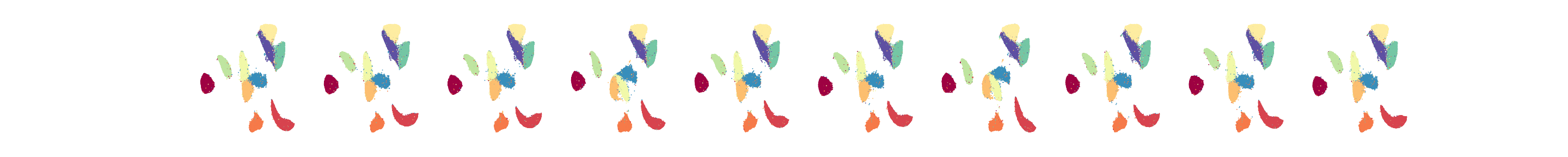}} 
\label{subfig: M_UMAP_I}\\
\subfloat[UMAP with random initialization on S curve with a hole dataset]{\includegraphics[width = 1\columnwidth]{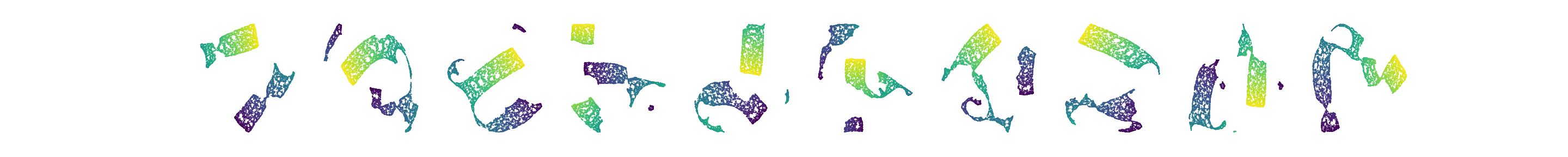}} 
\label{subfig:S_UMAP_R}\\
\subfloat[UMAP with Spectral Embedding (default) initialization on S curve with a hole dataset]{\includegraphics[width = 1\columnwidth]{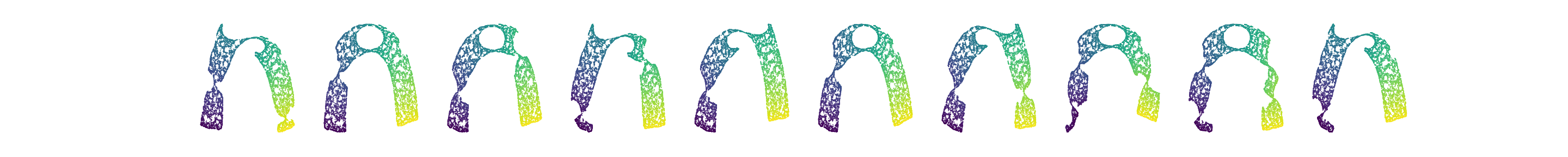}}
\label{subfig:S_UMAP_I}\\

\end{tabular}
\caption{UMAP's performance over ten runs to show performance consistency.}
\label{fig:UMAP_10_init}
\end{center}
\end{figure}

\begin{figure}[hbt]
\begin{center}
\centerline{\includegraphics[width=0.8\columnwidth]{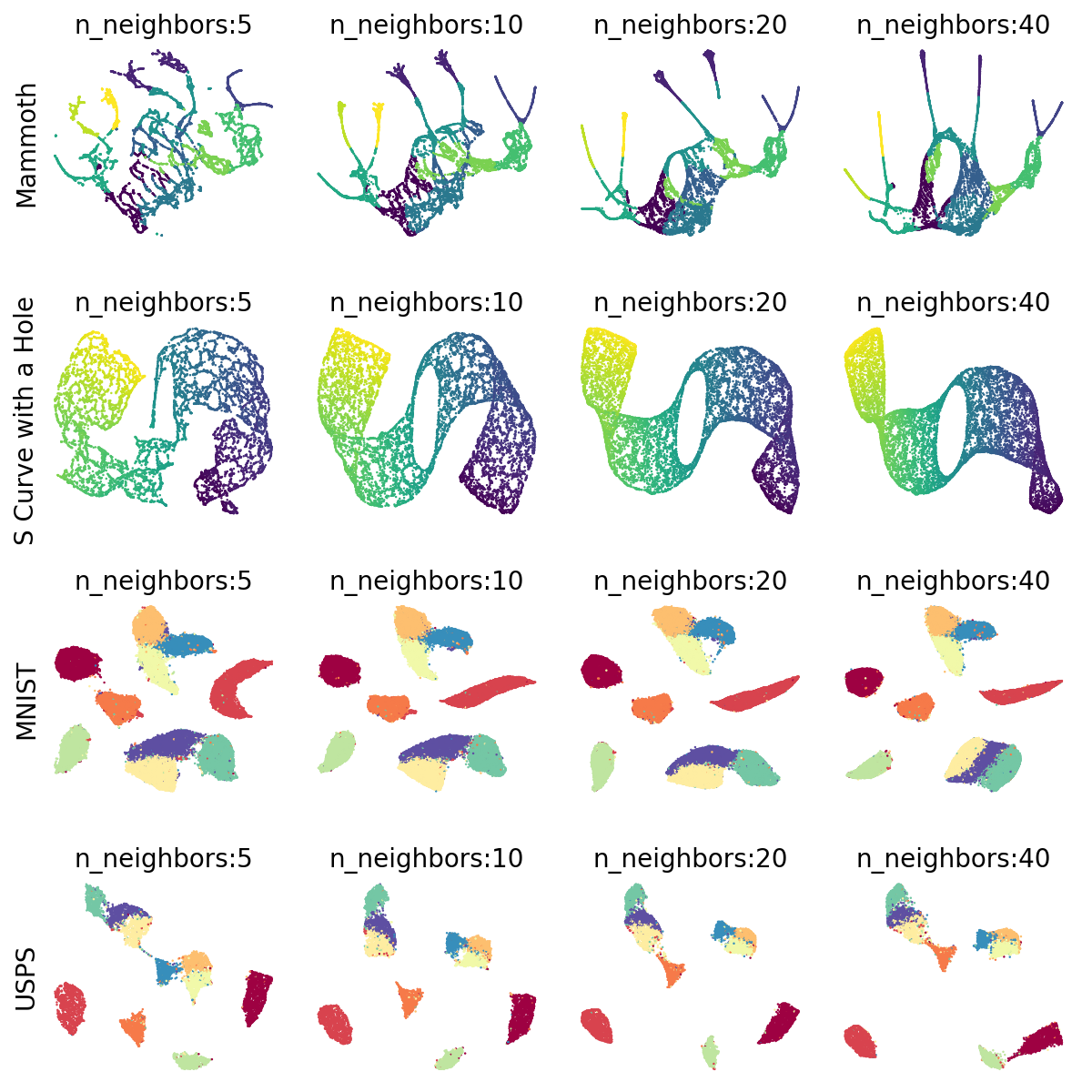}}
\caption{Robustness of PaCMAP to the number of neighbors (preserving the ratio of neighbors to mid-near and further point pairs). }
\label{fig:pacmap_robust_nn}
\end{center}
\end{figure}

\begin{figure}[hbt]
\begin{center}
\centerline{\includegraphics[width=0.8\columnwidth]{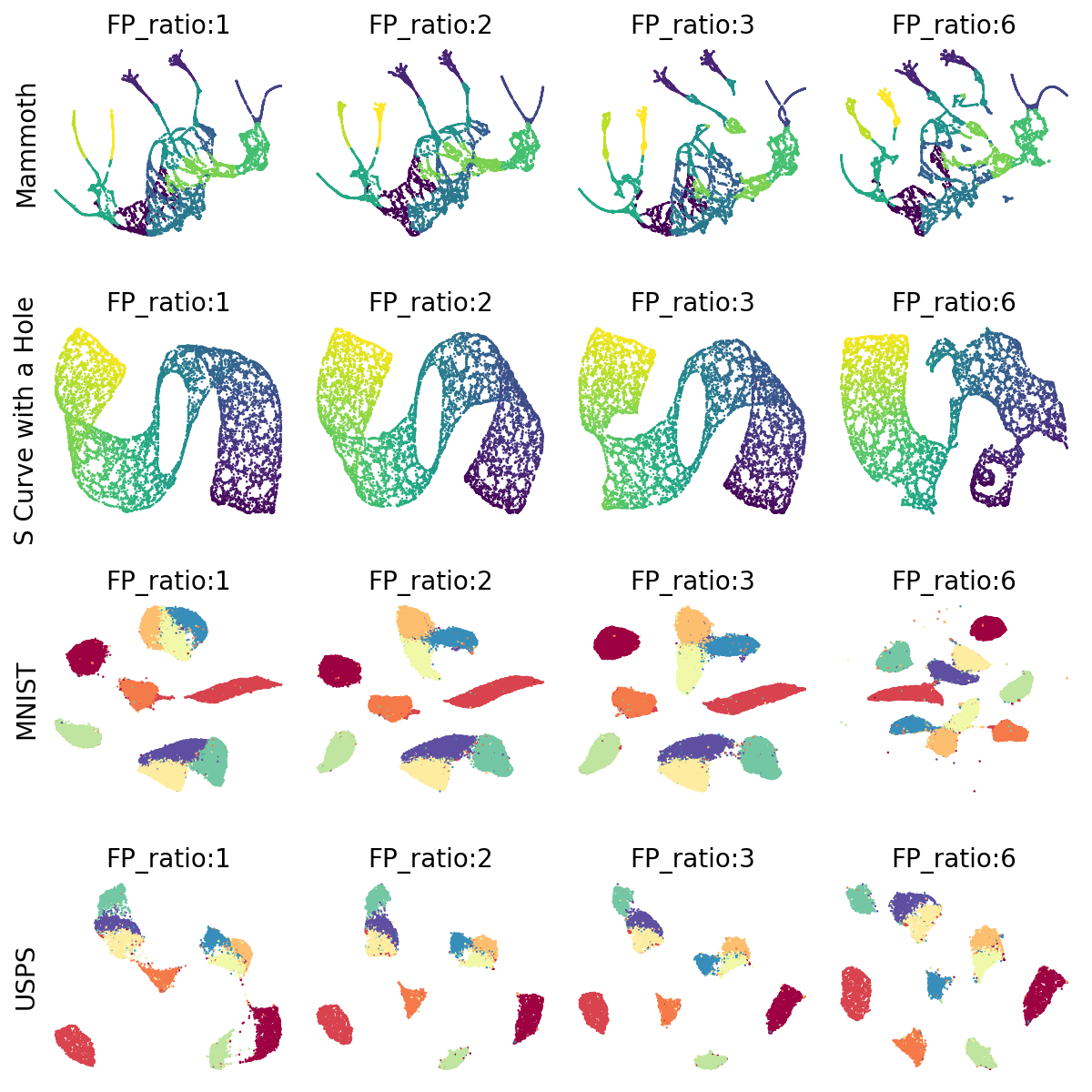}}
\caption{Robustness of PaCMAP to the ratio of further points.  }
\label{fig:pacmap_robust_fp}
\end{center}
\end{figure}

\begin{figure}[hbt!]
\begin{center}
\centerline{\includegraphics[width=0.8\columnwidth]{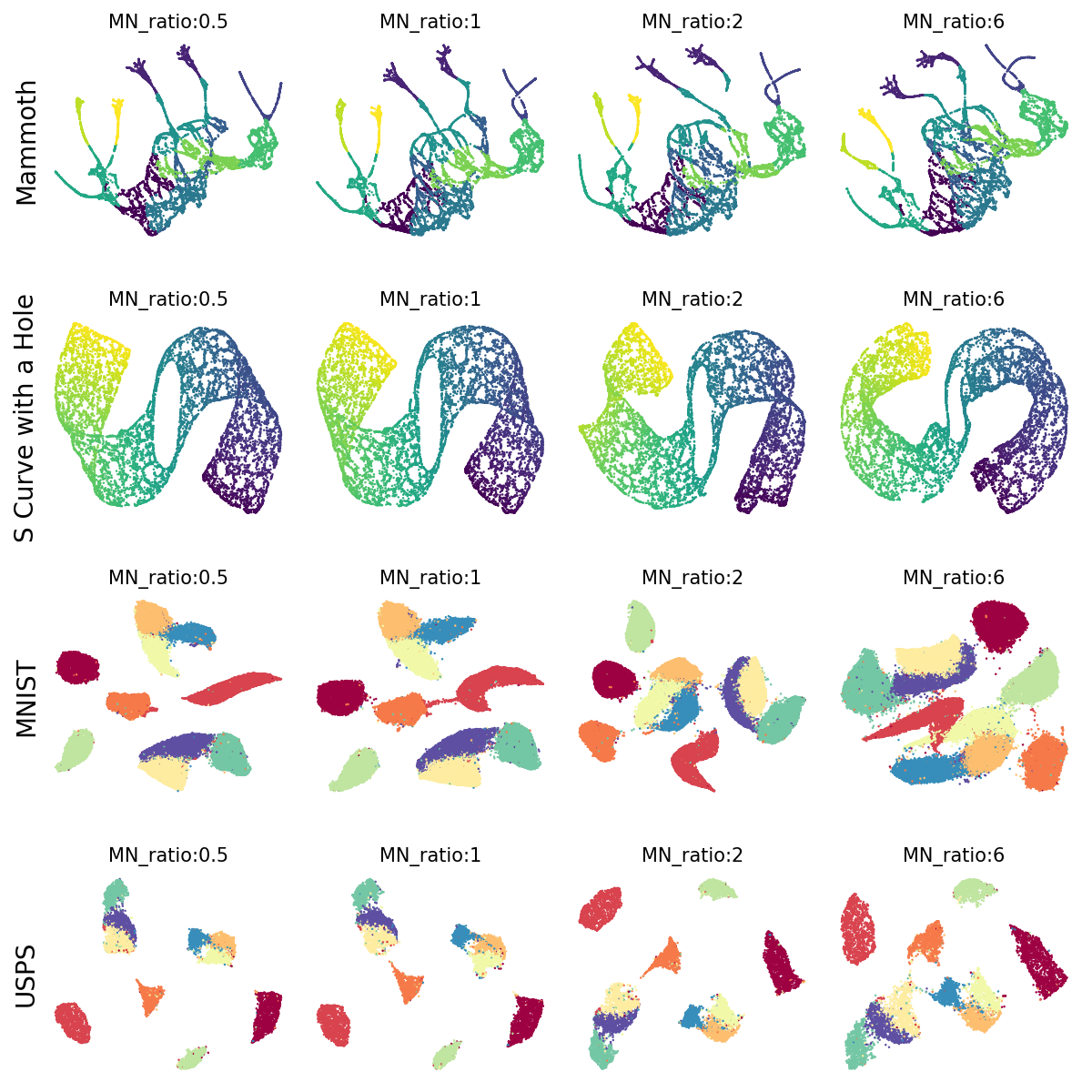}}
\caption{Robustness of PaCMAP to the ratio of mid-near points.  }
\label{fig:pacmap_robust_mid}
\end{center}
\end{figure}

Figures \ref{fig:PaCMAP_10_init}, \ref{fig:UMAP_10_init}, \ref{fig:pacmap_robust_nn}, \ref{fig:pacmap_robust_fp}, and \ref{fig:pacmap_robust_mid} include additional outcomes of DR discussed in Section~\ref{sec:sensitivity_analysis}.

\begin{figure}[hbt!]
\begin{center}
\centerline{\includegraphics[width=0.8\columnwidth]{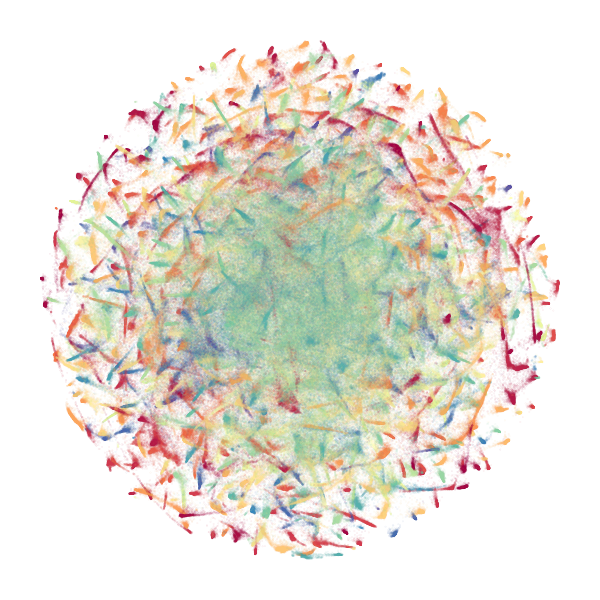}}
\caption{t-SNE visualization of the activations of all data from the ILSVRC 2012 dataset \citep{imagenet} arising from the penultimate layer of a ResNet-50 Convolutional Neural Network \citep{resnet}. Samples are colored by their labels. Similar labels share similar colors.}
\label{fig:imagenet_tsne}
\end{center}
\end{figure}

\begin{figure}[hbt!]
\begin{center}
\centerline{\includegraphics[width=0.8\columnwidth]{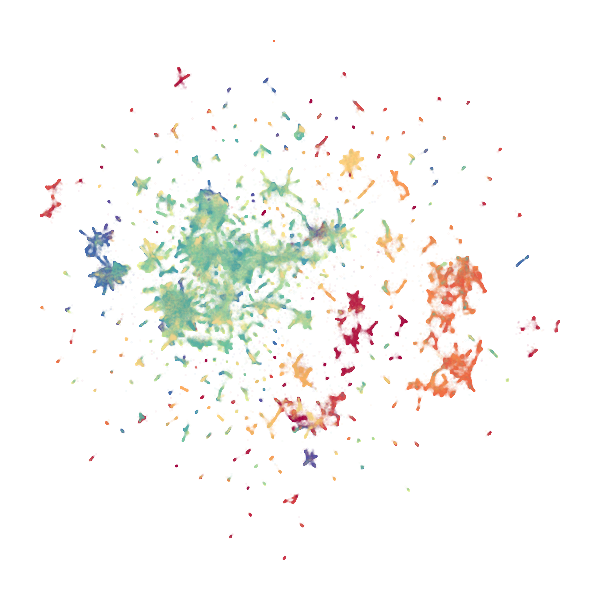}}
\caption{LargeVis visualization of the activations of all data from the ILSVRC 2012 dataset \citep{imagenet} arising from the penultimate layer of a ResNet-50 Convolutional Neural Network \citep{resnet}. Samples are colored by their labels. Similar labels share similar colors.}
\label{fig:imagenet_largevis}
\end{center}
\end{figure}

\begin{figure}[hbt!]

\begin{center}
\centerline{\includegraphics[width=0.8\columnwidth]{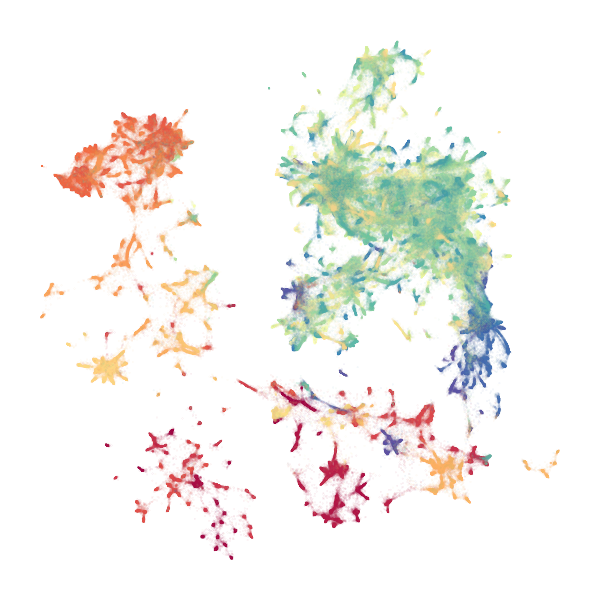}}
\caption{UMAP visualization of the activations of all data from the ILSVRC 2012 dataset \citep{imagenet} arising from the penultimate layer of a ResNet-50 Convolutional Neural Network \citep{resnet}. Samples are colored by their labels. Similar labels share similar colors.}
\label{fig:imagenet_umap}
\end{center}
\end{figure}

\begin{figure}[hbt!]

\begin{center}
\centerline{\includegraphics[width=0.8\columnwidth]{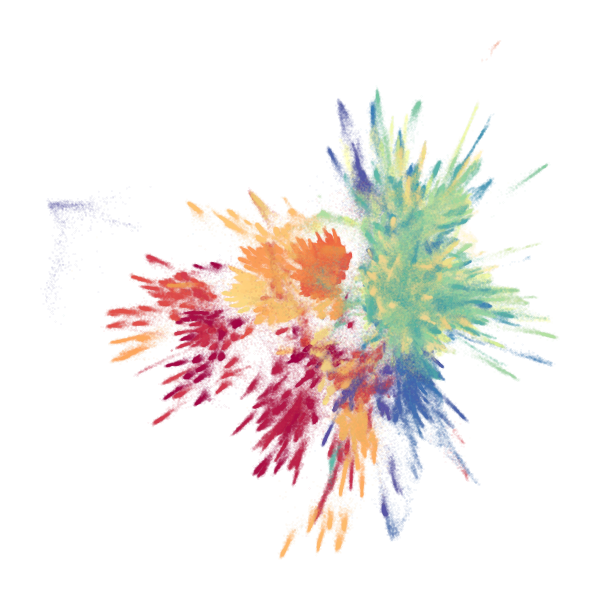}}
\caption{TriMap visualization of the activations of all data from the ILSVRC 2012 dataset \citep{imagenet} arising from the penultimate layer of a ResNet-50 Convolutional Neural Network \citep{resnet}. Samples are colored by their labels. Similar labels share similar colors.}
\label{fig:imagenet_trimap}
\end{center}
\end{figure}

\begin{figure}[hbt!]

\begin{center}
\centerline{\includegraphics[width=0.8\columnwidth]{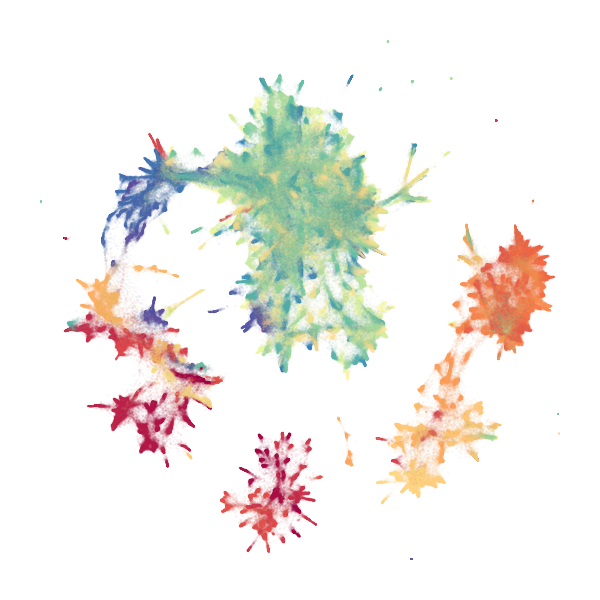}}
\caption{PaCMAP visualization of the activations of all data from the ILSVRC 2012 dataset \citep{imagenet} arising from the penultimate layer of a ResNet-50 Convolutional Neural Network \citep{resnet}. Samples are colored by their labels. Similar labels share similar colors.}
\label{fig:imagenet_PaCMAP}
\end{center}
\end{figure}

Figures \ref{fig:imagenet_tsne}, \ref{fig:imagenet_largevis}, \ref{fig:imagenet_umap}, \ref{fig:imagenet_trimap}, and \ref{fig:imagenet_PaCMAP} include visualization of the activations of all data from the ILSVRC 2012 dataset \citep{imagenet} arising from the penultimate layer of a ResNet-50 Convolutional Neural Network \citep{resnet}. 


\clearpage
\setcounter{figure}{0}    

\section{A Synthetic Hierarchical Dataset}
\label{sec:hierarchical}

In this section, we briefly describe a numerical experiment that was carried out to test the robustness of the various DR algorithms to hierarchical datasets. 
To this end, we created a synthetic hierarchical dataset that consists of one hundred and twenty five clusters, each including 500 observations. The data generation process was as follows: 

\begin{enumerate}
    \item We sampled five macro cluster centers from a 50-dimensional multivariate normal distribution with zero mean ($\textbf{0}_{50}$) and a covariance matrix that is equal to the identity matrix multiplied by 10000 (i.e., $10000 \times \textbf{1}_{50}$). 
    
    \item We then sampled five meso cluster centers for each of the macro clusters ($5\times 5 =$25 in total) using a 50-dimensional multivariate normal distribution whose mean values are the macro cluster centers and whose covariance matrices are equal to the identity matrix multiplied by 1000 (i.e., $1000 \times \textbf{1}_{50}$).
    
    \item We then sampled five micro cluster centers for each of the meso clusters ($25 \times 5 =$125 in total) using a 50-dimensional multivariate normal distribution whose mean values are the meso cluster centers and whose covariance matrices are equal to the identity matrix multiplied by 100 (i.e., $100 \times \textbf{1}_{50}$).

    \item Finally, for each of the one hundred and twenty five micro cluster centers, we sampled five hundred observations from a 50-dimensional multivariate normal distribution whose mean values are the micro cluster centers and whose covariance matrices are equal to $10 \times \textbf{1}_{50}$.
\end{enumerate}

\begin{figure}[ht]
\begin{center}
\centerline{\includegraphics[width=0.7\columnwidth]{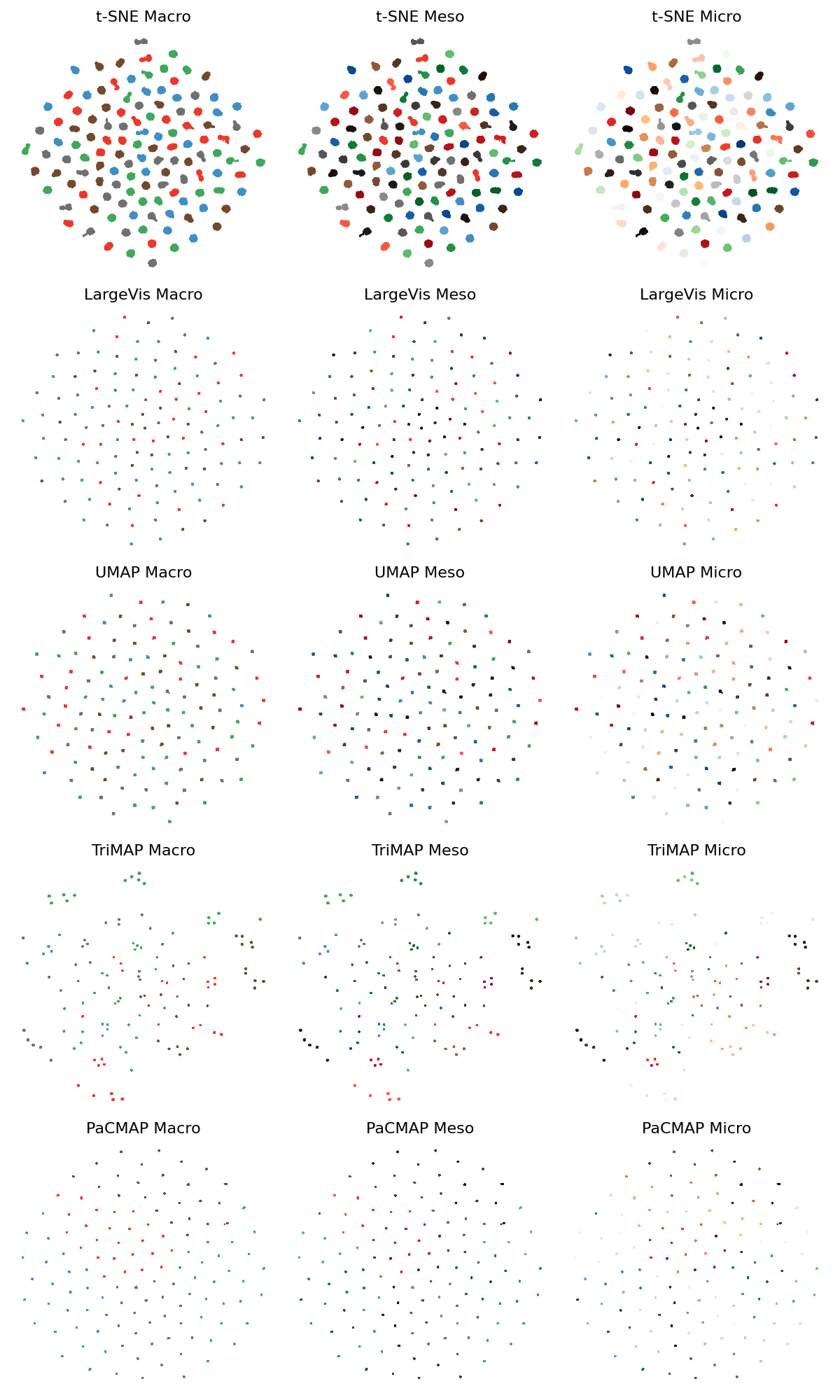} }  
\caption{Visualization of the three layer hierarchical synthetic dataset using t-SNE, UMAP, LargeVis, TriMap and PaCMAP.
In each column, the same embeddings are presented but are colored using different color schemes, according to the macro (first column), meso (second column), and micro (third column) levels.
In the middle and right figures, sub-clusters are visualized using colors similar to the higher level cluster.
}
\label{fig:hierarchical}
\end{center}
\end{figure}

We applied t-SNE, UMAP, LargeVis, TriMap and PaCMAP to the synthetic dataset. Figure~\ref{fig:hierarchical} shows the visualization of the dataset using different DR methods. We applied three different color schemes (in each column of the grid), according to the macro, meso, and micro cluster levels. That is, in the first column, we only use 5 colors based on the macro clusters, in the second column we use 25 colors based on the meso clusters, and in the third column we use 125 distinct colors. The spectrum of colors in each lower level cluster matches the color of its higher level cluster (e.g., if the higher level cluster is red, its lower level clusters would be colored in some shade of red).
Table~\ref{table-hierarchical} summarizes the quantitative evaluation metrics on the results of different DR methods. 

From the visualization, we can see that all of the DR methods preserve local structure (it appears that 125 micro clusters are identified and the metrics for the local structure preservation shown in Table~\ref{table-hierarchical} are perfect). On the other hand, not all preserve global structure. In the DR results of t-SNE, LargeVis, UMAP and TriMap, clusters with related colors (i.e., blues, grays, greens, reds and browns) are not always close together, showing that the global hierarchical structure is not preserved. In contrast, PaCMAP successfully preserved global structure by locating clusters within the same macro and meso group nearer to each other. This can also be observed by the metrics that quantify preservation of global structure, presented in  Table~\ref{table-hierarchical}. 
However, none of the algorithms were able to separate the macro and meso clusters from each other in a way that would be visible without the colored labels. 

\begin{table}[htb]
\caption{Evaluation metrics for the DR results on the 50 dimensional synthetic hierarchical dataset. The first two quantify the preservation of local structure, and the following two quantify the preservation of global structure (see Section~\ref{sec:numericals} for additional details). RT Accuracy stands for Random Triplet Accuracy, and CT Accuracy stands for Centroid Triplet Accuracy.}

\label{table-hierarchical}
\vskip 0.15in
\begin{center}
\hspace*{-20pt}
\begin{footnotesize}
\begin{sc}
\begin{tabular}{lccccc}
\toprule
Metric & t-SNE & LargeVis & UMAP & TriMap & PaCMAP  \\
\midrule
1NN accuracy & \textbf{1.000 $\pm$ 0.000} & \textbf{1.000 $\pm$ 0.000} & \textbf{1.000 $\pm$ 0.000} & \textbf{1.000 $\pm$ 0.000} & \textbf{1.000 $\pm$ 0.000} \\

SVM accuracy   & \textbf{1.000 $\pm$ 0.000} & \textbf{1.000 $\pm$ 0.000} & \textbf{1.000 $\pm$ 0.000} & \textbf{1.000 $\pm$ 0.000} & \textbf{1.000 $\pm$ 0.000}\\

RT accuracy & 0.508 $\pm$ 0.003 & 0.506 $\pm$ 0.004 & {0.506 $\pm$ 0.001} & 0.665 $\pm$ 0.003 & \textbf{{0.801 $\pm$ 0.002}}\\

CT accuracy& 0.499 $\pm$ 0.002 & 0.496 $\pm$ 0.006 & {0.497 $\pm$ 0.001} & 0.651 $\pm$ 0.003 & \textbf{0.794 $\pm$ 0.001}\\

Running Time  & 00:02:18 & 00:15:73 & 00:00:45 & {00:00:54} & \textbf{00:00:24}\\

\bottomrule
\end{tabular}
\end{sc}
\end{footnotesize}
\end{center}
\end{table}

\setcounter{figure}{0}    

\section{ForceAtlas2}

ForceAtlas2 \citep{ForceAtlas2} is the default graph layout algorithm for Gephi, a network visualization software. Similar to t-SNE and its variants, ForceAtlas2 maps a graph into a 2-dimensional space and uses attractive and repulsive forces to adjust the distance between points. ForceAtlas2 did not tend to perform well in practice, potentially owing to problems with its loss functions that we could identify using the rainbow figure. 


Nevertheless, researchers in the single-cell transcriptomic community have been applying force-directed graph visualization algorithms, such as ForceAtlas2, on k-Nearest Neighbor graphs constructed on transcriptomic data, because they have found that these algorithms are sometimes useful for discovering developmental patterns over time \citep{Unify}.

In this section, we aim to provide a similar analysis to that we performed on other algorithms in Section~\ref{sec:principles_good_objective} for ForceAtlas2. Here we consider the case where ForceAtlas2 is applied to a symmetric k-Nearest Neighbor graph, where the weight of the edge between two points, $i$ and $j$, equals 1 if one of them is a neighbor of the other, and 0 otherwise. This is the most common way these algorithms are implemented in single-cell transcriptomic data \citep[see, e.g.][]{Wagner981, Weinrebeaaw3381}.

\subsection{The attractive and repulsive force of ForceAtlas2}
Instead of defining a loss function and updating the layout with the gradient of the loss, the ForceAtlas2 algorithm directly defines its attractive and repulsive forces (akin to defining the derivatives of the loss, without defining the loss itself). It uses two simple functions for these forces. 

\paragraph{The attractive force.}
The attractive force between two points that are connected in the graph depends linearly on the Euclidean distance between two points in the low-dimensional space. Formally, the attractive force between two points, $i$ and $j$ is:

\[F_\textrm{attractive} = d_{ij}.\]

\paragraph{The repulsive force.}
ForceAtlas2 applies repulsive force between pairs of points regardless of whether an edge exists between them. The magnitude of the force is inversely proportional to the distance between two points in the low-dimensional space, and proportional to the degree (the number of edges that are connected to them) of the two points. Formally, the repulsive force between two points, $i$ and $j$ is:
\[F_\textrm{repulsive} = k_r\frac{(deg(i)+1)(deg(j)+1)}{d_{ij}}\]
where $deg(i)$ is the degree of i, and $k_r$ is a coefficient set by the algorithm that is capped at 10. 

If ForceAtlas2 is applied to a symmetric k-Nearest Neighbor graph, the degree of each point should be around $k$. Therefore, we can simply use $(k+1)^2$ to substitute for the numerator of the repulsive force.

\subsection{Visualizing and Analyzing the Loss function}
Following the definition in the last subsection, we can easily derive the loss associated with each type of pair, by integrating the forces:

$$
Loss_{\textrm{attractive}} = \frac{d_{ij}^2}{2},\quad
Loss_{\textrm{repulsive}} = k_r(k+1)^2\log(d_{ij}).
$$

\begin{figure}[hbt]
\begin{center}
\centerline{\includegraphics[width=.85\columnwidth]{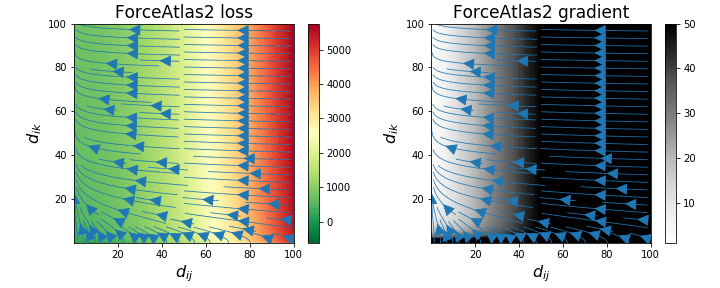}}

\caption{Rainbow Figure for ForceAtlas2.}
\label{fig:FA2Rainbow}
\end{center}
\end{figure}

\begin{figure}[hbt]
\begin{center}
\centerline{\includegraphics[width=.85\columnwidth]{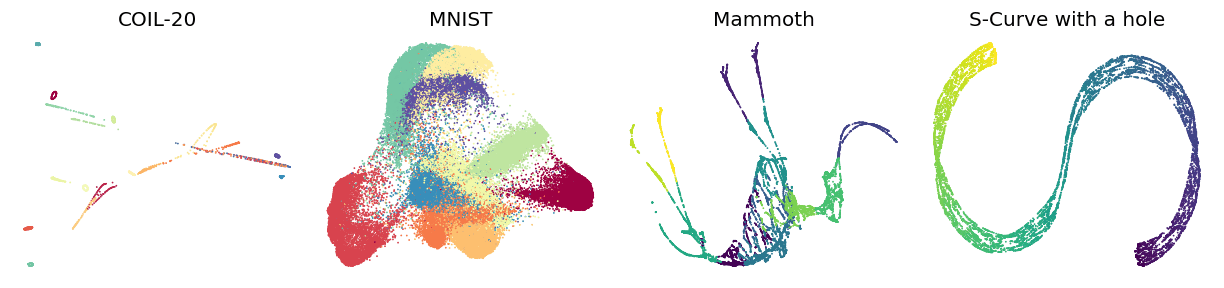}}

\caption{ForceAtlas2 Visualization of four datasets. We initialize the graph layout with the first two principal components of each dataset, with the standard deviation set to 10,000 to match the magnitude of the ForceAtlas2 algorithm.}
\label{fig:vizFA2}
\end{center}
\end{figure}

Similar to Section~\ref{sec:principles_good_objective}, we use a rainbow figure to visualize the loss and gradient of ForceAtlas2. As we can see in Figure~\ref{fig:FA2Rainbow}, ForceAtlas2 violates Principle 6 of Section~\ref{sec:principles_good_loss}, leading to a rainbow figure that looks similar to that of BadLoss 4. The difference between them is that ForceAtlas2 satisfies Principles 3 and 5, which leads to a stronger repulsive force. It is also worth mentioning that in ForceAtlas2, the repulsive force is applied to every pair of points in the low-dimensional embedding -- even the neighbors. As a result, as we can see in Figure~\ref{fig:vizFA2}, the local structure is better preserved, although there is still some overlapping between different clusters. The strong attractive force connects different clusters together through their outliers, making them harder to separate, but on the other hand, provides an opportunity to assess whether a continuous manifold exists in the dataset. However, just like many DR algorithm, ForceAtlas2 is not robust to hyperparameter choices and will require careful hyperparameter tuning to achieve optimal results. In our experiment, it fails to discover the hole on the continuous manifold in the S-curve with a hole dataset.

\bibliography{example_paper}

\end{document}